\InputIfFileExists{arXiv_update/project-paths.tex}{}{}
\documentclass[5p,10pt,a4paper,nopreprintline]{elsarticle}

\AtBeginDocument{\setlength{\mathindent}{0pt}}

\usepackage{amsmath,amssymb,amsfonts,amsthm,mathtools}
\usepackage{bm}
\usepackage{algorithm}
\usepackage{algorithmic}
\usepackage{enumitem}
\newlist{resultparts}{enumerate}{1}
\setlist[resultparts]{label=(\roman*),ref=\roman*,
  leftmargin=*,labelsep=.5em,topsep=3pt,itemsep=3pt,parsep=0pt,partopsep=0pt,
  before={\setlength{\abovedisplayskip}{6pt plus 1pt minus 1pt}%
          \setlength{\belowdisplayskip}{6pt plus 1pt minus 1pt}%
          \setlength{\abovedisplayshortskip}{3pt plus 1pt}%
          \setlength{\belowdisplayshortskip}{4pt plus 1pt minus 1pt}}}
\newcommand{\resultheading}[1]{#1}
\usepackage{graphicx}
\usepackage{tikz}
\usetikzlibrary{arrows.meta, decorations.pathmorphing}
\usepackage{subcaption}
\usepackage{booktabs}
\usepackage{threeparttable}
\usepackage{makecell}
\usepackage{multirow}
\usepackage{array}
\newsavebox{\PaperTableBox}
\newlength{\PaperTableWidth}
\NewDocumentEnvironment{paperTable}{m +b}{%
  \begingroup
  \setlength{\PaperTableWidth}{\linewidth}
  \sbox{\PaperTableBox}{%
    \begin{tabular}{@{}#1@{}}#2\end{tabular}
  }%
  \ifdim\wd\PaperTableBox>\PaperTableWidth
    \resizebox{\PaperTableWidth}{!}{\usebox{\PaperTableBox}}%
  \else
    \begin{tabular*}{\PaperTableWidth}{@{\extracolsep{\fill}}#1@{}}%
      #2%
    \end{tabular*}%
  \fi
  \endgroup
}{}

\usepackage{siunitx}
\usepackage{xcolor}
\definecolor{ReferenceBlue}{HTML}{00008B}

\usepackage{placeins}
\usepackage{needspace}
\usepackage{microtype}

\usepackage{aliascnt}
\usepackage{xurl} 
\usepackage[colorlinks=true,allcolors=ReferenceBlue]{hyperref}
\usepackage[nameinlink,noabbrev]{cleveref}
\AtBeginDocument{\hypersetup{allcolors=ReferenceBlue}}
\makeatletter
\let\ReferenceTagForm\tagform@
\DeclareRobustCommand{\eqref}[1]{%
  \textup{\begingroup\color{ReferenceBlue}%
    \ReferenceTagForm{\ref{#1}}\endgroup}}
\pretocmd{\NAT@citenum}{\color{ReferenceBlue}}{}{%
  \PackageError{Manuscript}{Unable to colour complete citations}{Check the natbib citation hook.}}
\makeatother

\hypersetup{bookmarksdepth=3}

\makeatletter
\renewcommand\subsection{\@startsection{subsection}{2}{\z@}%
  {12\p@ \@plus 6\p@ \@minus 3\p@}%
  {3\p@ \@plus 6\p@ \@minus 3\p@}%
  {\normalfont\normalsize}}
\makeatother

\newtheoremstyle{preprintstatement}
  {\topsep}{\topsep}
  {\itshape}{}{\bfseries\itshape}{.}
  {5pt plus 1pt minus 1pt}
  {\thmname{#1}\thmnumber{ #2}\thmnote{ (#3)}}
\theoremstyle{preprintstatement}
\newtheorem{theorem}{Theorem}[section]
\newaliascnt{proposition}{theorem}
\newtheorem{proposition}[proposition]{Proposition}
\aliascntresetthe{proposition}
\newtheorem{lemma}{Lemma}[section]
\newaliascnt{corollary}{theorem}
\newtheorem{corollary}[corollary]{Corollary}
\aliascntresetthe{corollary}
\newtheorem{definition}{Definition}[section]
\newtheorem{assumption}{Assumption}[section]
\newtheorem{remark}{Remark}[section]

\biboptions{numbers,sort&compress}
\newcommand{\EvalRes}{\textsc{EvalRes}}
\newcommand{\SO}{\mathrm{SO}}

\newcommand{\tr}{\operatorname{tr}}
\newcommand{\diag}{\operatorname{diag}}
\input{supplementary_refmap.tex}
\begin{document}
\hypersetup{
  pdftitle={Wrench-Based Bayesian Pose Estimation via Matrix Fisher–Gaussian Inference},
  pdfauthor={Jianyu Chen, Lin Yang, Yibang Li, Domenico Campolo, Cyrus Mostajeran}
}
\begin{frontmatter}

\title{{Wrench-Based} Bayesian Pose Estimation via \\ {Matrix Fisher--Gaussian} Inference}

\author[spms]{Jianyu Chen}
\ead{jianyu.chen@ntu.edu.sg}
\author[mae]{Lin Yang}
\ead{YANG0752@e.ntu.edu.sg}
\author[spms]{Yibang Li}
\ead{YIBANG001@e.ntu.edu.sg}
\author[mae,r3i]{Domenico Campolo}
\ead{d.campolo@ntu.edu.sg}
\author[spms]{Cyrus Mostajeran\corref{cor1}}
\ead{cyrussam.mostajeran@ntu.edu.sg}

\cortext[cor1]{Corresponding author.}
\affiliation[spms]{organization={School of Physical and Mathematical Sciences, Nanyang Technological University},
  city={Singapore},
  postcode={637371},
  country={Singapore}}
\affiliation[mae]{organization={School of Mechanical and Aerospace Engineering, Nanyang Technological University},
  city={Singapore},
  postcode={639798},
  country={Singapore}}
\affiliation[r3i]{organization={R3 Institute (R3I) Singapore, Nanyang Technological University},
  city={Singapore},
  postcode={639798},
  country={Singapore}}

\begin{abstract}
Estimating object position and orientation from contact forces and torques is important in robotics when vision is unavailable or unreliable.
The inverse problem involves an implicit mechanical equilibrium and a nonlinear pose space.
We develop a Bayesian framework for quasi-static contact measurements on ${\SO(3)}\times\mathbb{R}^3$, using a coupled Matrix Fisher--Gaussian (MFG) distribution to represent position and orientation uncertainty.
The observation likelihood is generally not conjugate to this family.
Implicit differentiation of the equilibrium equations yields a reduced wrench sensitivity for constructing a local quadratic model of the negative log posterior.
The Coupled MFG update computes parameters of a local MFG posterior approximation in closed form by matching a conditional-mean-linearized surrogate to this model.
To improve robustness to the linearization point, we introduce a residual-safeguarded refinement procedure that reuses the original prior and measurements.
We establish conditional bounds on posterior approximation error and analyse the computational trade-off with particle-based inference.
Numerical comparisons demonstrate computational advantages under the tested conditions.
Ellipsoid and smoothed Stanford bunny simulations illustrate concentration of the fitted MFG distributions as measurements accumulate, while robotic experiments demonstrate pose estimation from contact measurements on objects with known geometry.
\end{abstract}


\end{frontmatter}

\section{Introduction}
\label{sec:introduction}
Accurate pose estimation from contact-induced force and torque measurements is important in contact-rich robotic manipulation, assembly, and insertion. In industrial, medical, and confined environments, vision may be degraded by occlusion, reflection, poor illumination, or limited field of view, making haptic sensing{, here referring to contact-based force and torque sensing,} a valuable source of information about object geometry and interaction state. Foundational impedance and compliance models establish that physical interaction is mechanically coupled through forces and constraints \cite{hogan1985impedance,mason1981compliance}. Haptic feedback is likewise important when visual information is insufficient \cite{okamura2009haptic}. Recent quasi-static haptic manipulation work uses force-rich contact for planning, adaptation, and online pose or shape inference \cite{yang2025RAL,yang2026adaptive,lu2026haptic}. Nevertheless, estimating an unknown object pose from wrench observations alone remains difficult because the measurement map is nonlinear, configuration dependent, and shaped by contact geometry and quasi-static equilibrium.

A further challenge is geometric: the pose lies on
{the direct product} $\mathrm{SO}(3)\times\mathbb{R}^3$. 
Classical attitude filters maintain a global attitude representation while
using local error coordinates for filtering and covariance propagation
\cite{crassidis2007survey,markley2003attitude}.
{Earlier geometric complementary observers also exploit the structure
of $\mathrm{SO}(3)$ for multimodal attitude estimation
\cite{campolo2008multimodal}.} Invariant and equivariant
observer/filter constructions provide a related geometric route for
Lie-group estimation \cite{barrau2017invariant,vangoor2023eqf}.  From the perspective of posterior
uncertainty, directional-statistical models provide a complementary route:
the Bingham distribution treats antipodal quaternion uncertainty, and the
{matrix Fisher distribution} is defined intrinsically on $\mathrm{SO}(3)$
\cite{bingham1974antipodally,khatri1977von,lee2018bayesian}.  The {Matrix Fisher--Gaussian} (MFG)
distribution further combines {a matrix Fisher distribution as its rotational marginal} with a
conditional Gaussian Euclidean component, thereby representing
rotation--translation coupling on
$\mathrm{SO}(3)\times\mathbb{R}^{n}$ \cite{wang2022matrix}.
Beyond this established family, our contribution is a
closed-form local reconstruction for an implicit wrench likelihood through
second-order matching of a conditional-mean-linearized surrogate.

A closely related local alternative is the Laplace approximation, which
constructs a Gaussian from a posterior mode and the local curvature of
the negative log posterior \cite{tierney1986accurate}.
For pose estimation, this Gaussian can be expressed in tangent
coordinates and mapped to the pose manifold
\cite{barfoot2014associating}.
Both {tangent-space Gaussian} and coupled MFG representations can retain
rotation--translation correlations. The distinction is that MFG models
rotational uncertainty by {the density of a normalized matrix Fisher distribution} defined
directly on $\mathrm{SO}(3)$, rather than by a Gaussian in local
rotational coordinates. This motivates a coupled MFG representation for
a predominantly unimodal posterior with appreciable rotational
uncertainty, while the likelihood approximation remains local.
{Both approaches describe the posterior through a finite set of
distribution parameters. A nonparametric alternative represents it
directly by weighted samples.}
Particle methods, including resample--move methods and sequential
Monte Carlo samplers, approximate the posterior with weighted samples
and can retain separated pose hypotheses
\cite{gilks2001resamplemove,delmoral2006smc}.
For an implicit contact model, evaluating the likelihood at new
particle poses generally requires equilibrium solves under the
relevant controls. This cost motivates a compact parametric
approximation {when a single dominant pose region is sufficient to describe the posterior.} 
Koval et al.~\cite{koval2015manifold} address particle starvation during
planar contact manipulation by sampling on contact manifolds.
Zheng et al.~\cite{zheng2026bayesian} use a customized particle filter
for active joint object recognition and six-dimensional pose
estimation from force/torque, GelSight, and free-space evidence.

Existing filtering results based on the matrix Fisher distribution and MFG models mainly address attitude
dynamics, direct attitude information, or reference-vector observations
\cite{lee2018bayesian,wang2022matrix,wang2025nonunit}. In contrast, we derive a local update based on implicit contact likelihoods, 
where the observations are not measurements of 
pose but contact forces and torques, whose dependence on pose is predicted by an implicit model. In our quasi-static model, the predicted observation is a generalized force evaluated at equilibrium, obtained by differentiating a reduced elastic potential with respect to the chosen control perturbation 
after eliminating the end-effector equilibrium state. 
The residual map is differentiable, but the likelihood is not conjugate to the coupled MFG family and may provide little information about rotation even when translation is well constrained.
These features raise three control-theoretic 
questions: how to reduce the implicit contact model to a pose sensitivity map, 
how to perform a structure-preserving Bayesian update on $\mathrm{SO}(3)\times\mathbb{R}^3$ when global conjugacy is absent, and how to diagnose when the available contact batch is intrinsically rotation-informative.

This paper answers these questions through a local, model-based inference framework. First, we 
propose {the Coupled MFG update in closed form}.  We represent pose uncertainty by 
a coupled MFG distribution and model each force/torque observation through a differentiable 
quasi-static contact potential. Eliminating the end-effector equilibrium state yields a 
reduced {generalized-force} residual whose first-order sensitivity is obtained by a Schur-complement formula, which depends jointly on pose geometry, contact geometry, and the local equilibrium response. {Here, a \emph{nominal pose} is the reference pose at which the local model and its derivatives are evaluated~\cite{barfoot2017state}. It need not coincide with the prior centre or a posterior mode.} Around {this pose}, the nonlinear likelihood is then replaced by a Gauss--Newton quadratic information model in right-perturbation coordinates. {We show that this local posterior model admits {the Coupled MFG update, whose} conditional-mean-linearized (CL) surrogate energy matches the prescribed local gradient and curvature.} Thus, 
the proposed update is not claimed to be globally conjugate. {It is a local MFG reconstruction defined by second-order
matching of the CL surrogate to the quadratic
posterior model.} 
Building on equilibrium-based haptic estimation~\cite{yang2026adaptive}
and quasi-static mechanical modelling~\cite{campolo2025geometric}, we develop a
local framework for Bayesian pose inference from wrench observations, including joint posterior
reconstruction and a translation-compensated rotational information diagnostic. {Algorithm~\ref{alg:safeguarded_multipass_mfg_refinement} adds a residual
acceptance test to the single-pass reconstruction in
Algorithm~\ref{alg:single_pass_mfg_update}. We call this test the
\emph{residual safeguard}: a candidate nominal pose is accepted only if
its recomputed merit satisfies
$\rho_{\rm try}\leq\rho_t-\varepsilon_{\rm acc}$,
$\varepsilon_{\rm acc}\geq0$. Thus accepted residuals are nonincreasing,
and decrease strictly when $\varepsilon_{\rm acc}>0$
(\Cref{lem:residual_monotonicity}); this condition does not guarantee a
reduction in true pose error or posterior approximation error.} {To assess the informativeness of wrench measurements}, the same model also gives a compensated rotational information matrix obtained by eliminating translational perturbations through a Schur complement. Its smallest eigenvalue measures the weakest locally observable attitude direction after optimal translational compensation. This quantity serves as a local identifiability diagnostic and helps explain information-limited behaviour in sparse contact batches. {This interpretation connects the diagnostic to local rank conditions,
Fisher information and optimal experiment design
\cite{ljung1999system,pukelsheim2006optimal}.}

The main contributions of this work are summarized as follows:
\begin{enumerate}
    \item \textbf{Reduced Jacobian and rotational Schur complement.} We derive an equilibrium-consistent reduced {generalized-force} Jacobian
    by accounting for the implicit response of the quasi-static
    contact equilibrium. We provide a coupled rotational--translational
    measurement information matrix, which distinguishes likelihood
    information from prior contributions and quantifies effective
    rotational information through a Schur complement. 

    \item \textbf{{Coupled MFG update and optional freezing.}}
    We establish a closed-form local reconstruction theorem for the coupled MFG family.
    The reconstructed parameters yield a
    conditional-mean-linearized (CL) MFG surrogate that matches a
    prescribed quadratic data-and-prior energy through second
    order, with an \(O(\|\xi\|^3)\) local remainder. 
    We also provide an optional freezing variant
    (Corollary~\ref{cor:frozen_statistic_reconstruction}) that reuses the prior
    statistic and its Jacobian to reduce the computational cost of determining
    the MFG parameters. Proposition~\ref{prop:opt-freezing-certificate} gives
    sufficient conditions for using this variant.

    \item \textbf{Fixed-prior multi-pass recentring.} We develop single- and multi-pass {Coupled MFG
    update strategies}, with an iterative recentring procedure that
    preserves the original prior and measurement batch. Successive
    passes refine the local approximation without treating
    previously used measurements as new evidence.

    \item \textbf{Posterior discrepancy, MC error and computational efficiency.}
    {We derive conditional posterior-distance bounds in total variation
    and decompose mean-square integration error into squared bias and
    Monte Carlo (MC) sampling variance.}
    {{Proposition~\ref{prop:opt-freezing-certificate} also bounds} the
    discrepancy between the distributions produced by the freezing variant
    and the Coupled MFG update.}
\end{enumerate}

The main text develops the {potential-based model}, information structure, {Coupled MFG update rule} and safeguarded refinement algorithm.
{The appendices contain the derivations, optional-freezing conditions,
auxiliary algorithms and error analyses supporting
Sections~\ref{sec:wrench_sensitivity_information}--\ref{sec:evaluation}.
Supplementary Sections~\SMref{supp:sec:geometry-contact}--\SMref{supp:s4:physical-validation}
provide model details and additional experiments;
Sections~\SMref{supp:posterior-extensions} and~\SMref{app:opt-evidence}
provide further posterior-error analysis and paired cost--error comparisons,
respectively.  The
ellipsoid and bunny simulations in Section~\ref{subsec:shape_pose_clouds} are explained 
in detail in Supplementary Section~\SMref{supp:shape-clouds},
including geometry construction, posterior visualization and numerical checks.}

\section{Problem Formulation and Coupled MFG Prior}%
\label{sec:problem_formulation}

\subsection{Pose representation and local coordinates}
\label{subsec:pose_coordinates}

Let $X_B=(R_B,p_B)\in G:=\mathrm{SO}(3)\times\mathbb R^3$ be the unknown pose.
{We use the direct-product representation, with componentwise
composition $(R,p)\cdot(S,q)=(RS,p+q)$. This choice of group structure does
not impose statistical independence between rotation and translation in
the coupled MFG model.}
Around {the nominal pose} $\bar X_B=(\bar R_B,\bar p_B)$ use the product-space right
perturbation
$
 X_B=\bar X_B\oplus\xi
 :=(\bar R_B\exp(\phi^\wedge),\bar p_B+v)$,  where $ \xi=[\phi^\top\;v^\top]^\top\in\mathbb R^6.$
{Here $(\cdot)^\wedge:\mathbb R^3\to\mathfrak{so}(3)$ is the
standard linear isomorphism, with inverse $(\cdot)^\vee$. The rotation
increment $\phi$ uses right exponential coordinates, combined with additive
world-frame translation coordinates $v$.} Rotation increments are body-fixed at
$\bar R_B$, translation increments use world coordinates, and local
derivatives use $\xi$ unless stated otherwise. We use the following error measures on the product manifold $\mathrm{SO}(3)\times\mathbb R^3$.

\begin{definition}[Pose Error]
\label{def:pose_errors}
\leavevmode\par\nobreak\noindent
For $X_i=(R_i,p_i)$, define
\[
\begin{aligned}
d_{\mathrm{SO}(3)}(R_1,R_2)&=\|\operatorname{Log}(R_1^\top R_2)^\vee\|_2\\
\text{and}\quad d_{\mathbb R^3}(p_1,p_2)&=\|p_2-p_1\|_2.
\end{aligned}
\] The errors for rotation and translation are
\begin{equation}
 e_R=d_{\mathrm{SO}(3)}(R_B^{\rm true},\hat R_B),\qquad
 e_p=\|p_B^{\rm true}-\hat p_B\|_2.
 \label{eq:pose_errors}
\end{equation}

\end{definition}

\subsection{Reduced Observation Model}
\label{subsec:force_observation_model}

{Consider an end effector $A$ interacting with an object $B$.
Let $X_A=(R_A,p_A)$ denote the end-effector pose and
$X_U=(R_U,p_U)$ the commanded reference pose. For a settled, quasi-static
measurement, we denote the selected equilibrium pose by
$X_A^\star(X_U,X_B)$; the superscript marks the equilibrium value
of $X_A$. This pose generally differs from $X_U$, as illustrated in
Figure~\ref{fig:contact_scene}.} For measurement $k$, the input $X_{U,k}$ specifies this
commanded reference pose. The object pose remains fixed over the batch of
$K$ settled, quasi-static measurements, with known controls
$\mathcal U=\{X_{U,k}\}_{k=1}^K\subset G$.

Each $y_k=[\tau_k^\top,f_k^\top]^\top\in\mathbb R^6$ denotes an observation
in the model's generalized-force coordinates, conjugate to the local
control perturbation. The rotational and translational components have
units $\mathrm{N\,m}$ and $\mathrm N$, respectively. A physical load paired
with this perturbation has its moment about $p_{U,k}$ expressed in the
body axes of $U$, and its force expressed in world axes.
In a physical experiment, a six-axis force/torque sensor reports a wrench
in its own sensor frame about the sensor origin; the setup in
Section~\ref{subsec:original_physical} uses an ATI Mini40 sensor.
The relation between such readings and the model coordinates is specified
below, after eliminating the equilibrium state.

\begin{figure}[!htbp]
\centering
\resizebox{\linewidth}{!}{%
\begin{tikzpicture}[
  x=1cm,y=1cm,
  font=\small,
  every node/.style={text=black},
  line cap=round,line join=round,
  axis/.style={-{Latex[length=1.7mm]},draw=black,line width=.65pt},
  reference/.style={axis,densely dashed},
  physical/.style={draw=black,line width=.8pt},
  guide/.style={draw=black!65,line width=.6pt}
]
\path (0,0) rectangle (11.5,4.7);

\draw[physical] (1.05,1.72) rectangle (1.55,2.57);
\draw[physical,line width=2.3pt] (1.55,2.15) -- (2.55,2.15);
\draw[physical]
  (2.55,1.77) rectangle (3.15,2.53);
\draw[physical,line width=2.3pt] (3.15,2.15) -- (3.85,2.15);
\draw[physical] (3.85,1.92) rectangle (4.24,2.38);
\draw[physical,line width=2.3pt] (4.24,2.15) -- (7.72,2.15);
\node[anchor=north,align=center] at (1.30,1.50) {robot};
\node[anchor=south,align=center] at (2.57,2.75) {F/T sensor $S$};
\draw[guide] (2.85,2.70) -- (2.85,2.53);
\node[anchor=north] at (5.85,2.03) {rigid tool};

\draw[physical]
  (7.78,1.13) -- (10.70,1.13) -- (10.70,2.61) -- (7.78,2.61) -- cycle;
\draw[physical]
  (7.78,2.61) -- (8.43,3.16) -- (11.35,3.16) -- (10.70,2.61);
\draw[physical]
  (10.70,1.13) -- (11.35,1.68) -- (11.35,3.16);
\draw[physical] (8.32,1.13) -- (8.32,.89) -- (10.30,.89) -- (10.30,1.13);
\draw[physical] (7.98,.83) -- (10.65,.83);
\fill[black] (7.78,2.15) circle (1.7pt);
\draw[physical] (7.72,2.15) -- (7.78,2.15);
\node[anchor=south] at (9.45,3.37) {object $B$: unknown pose $X_B$};

\foreach \ox/\oy in {.48/.73,2.85/2.15,4.04/2.15,9.17/1.84}{
  \draw[axis] (\ox,\oy) -- ++(.60,0);
  \draw[axis] (\ox,\oy) -- ++(0,.55);
}
\node[anchor=north east] at (.50,.66) {$W$};
\node[anchor=north east] at (2.42,1.92) {\scriptsize $S$};
\node[anchor=north west] at (4.05,2.09) {\scriptsize $A$};
\node[anchor=north west] at (9.18,1.78) {$B$};
\node[anchor=north,align=center] at (4.06,1.57)
  {{equilibrium pose of $A$} \\{$X_A^\star(X_U,X_B)$}};

\draw[reference] (4.04,3.91) -- ++(.60,0);
\draw[reference] (4.04,3.91) -- ++(0,.55);
\node[anchor=south west] at (4.77,3.83) {command reference $U$: $X_U$};
\draw[guide,densely dashed] (3.76,3.63) rectangle (4.31,4.19);
\draw[draw=black,line width=.7pt,
  decorate,decoration={coil,aspect=.42,segment length=3.7pt,amplitude=2.1pt}]
  (4.04,2.76) -- (4.04,3.59);
\draw[guide] (4.04,2.70) -- (4.04,2.76);
\draw[guide] (4.04,3.59) -- (4.04,3.63);
\node[anchor=west,align=left] at (4.62,3.08) {virtual compliance};
\node[anchor=north,font=\footnotesize] at (6.25,.38)
  {$X_A^\star\ne X_U$ in general, $X_U$ is a reference pose.};
\end{tikzpicture}%
}

\caption{{Quasi-static contact setting. The rigid tool on the actual end effector $A$
contacts the object $B$. We denote the force/torque sensor frame by $S$ and the
world frame by $W$. The dashed frame $U$ is the commanded reference pose, and the
spring represents virtual compliance. {The equilibrium pose
$X_A^\star(X_U,X_B)$ generally differs from $X_U$.}}}
\label{fig:contact_scene}
\end{figure}

\begin{definition}[Potential]
\label{def:potential}
Let $W(X_A,X_U,X_B)$ denote the total potential, where
$X_A$ is the {end-effector pose},
$X_U$ is the control input, and
$X_B$ is the object pose. Let $n_c$ be the number of contact points, 
and let $r_A^{(i)}$, $i=1,\ldots,n_c$, be fixed probe points
in frame $A$. Let $F_B:\mathbb{R}^3\to\mathbb{R}$ denote a signed implicit field
representing the object geometry in frame $B$. Superquadric inside--outside functions provide one such
representation
\cite{barr1981superquadrics,yang2024pathplanning}. The control and contact
contributions are
\begin{equation*}
  W
  =
  \frac{1}{2} e_U^\top K_c e_U
  +
  \frac{1}{2}
  \sum_{i=1}^{n_c} k(d_i)\,d_i^2 .
\end{equation*}
The control error is
$
  e_U=\bigl[(\operatorname{Log}(R_U^\top R_A)^\vee)^\top,
              (p_A-p_U)^\top\bigr]^\top.
$
For the contact potential, let
$$
d_i=\frac{F_B(c_B^{(i)})}{\lVert\nabla F_B(c_B^{(i)})\rVert+{\varepsilon_{\mathrm{reg}}}},
$$
where $            
  c_B^{(i)}=R_B^\top\bigl(p_A+R_A r_A^{(i)}-p_B\bigr).$
The quantity $d_i$ is a regularized local distance surrogate.
Following the smooth contact regularization of
Campolo and Cardin \cite{campolo2025geometric},
we use the stiffness law
$k(d)=k_{\min}+\tfrac{k_{\max}}{2}[1-\tanh(d/d_0)]$.
The parameters satisfy
$K_c\succ0$, $\varepsilon_{\mathrm{reg}}>0$, $k_{\min}\geq0$,
$k_{\max}>0$, and $d_0>0$.
The synthetic experiments use
$\varepsilon_{\mathrm{reg}}=10^{-8}\,\mathrm m^{-1}$ as a small
numerical regularization in this denominator.

\end{definition}

The following assumption specifies the local existence and
regularity of the equilibrium map $X_A^\star(X_U,X_B)$ introduced above.

\begin{assumption}
\label{ass:equilibrium_regular}
\label{ass:equilibrium_smooth}
The potential \(W\) is \(C^2\) in a neighbourhood of interest.  For every
\((X_U,X_B)\) in this neighbourhood, there exists a locally unique
quasi-static equilibrium state
$
    X_A^\star(X_U,X_B)
$
satisfying
$
    \operatorname{grad}_{X_A}
    W\bigl(X_A^\star(X_U,X_B),\allowbreak X_U,X_B\bigr)=\nobreak0.
$
Moreover, the Hessian block of \(W\) with respect to \(X_A\) at the
equilibrium is nonsingular.  Hence, by the implicit function theorem~\cite{krantz2013implicit}, \(X_A^\star(X_U,X_B)\) is locally \(C^1\)
in \((X_U,X_B)\).
\end{assumption}

\begin{definition}[Reduced Potential]
\label{def:observation_map}
The reduced potential obtained after eliminating the quasi-static
end-effector state is
\[
    W^\star(X_U,X_B)
    :=
    W\bigl(X_A^\star(X_U,X_B),X_U,X_B\bigr).
\]
The predicted generalized force evaluated at equilibrium, conjugate to
the local control perturbation at the \(k\)-th reference pose, is
\[
    h_k(X_B)=\widehat y_k(X_B)
    :=-\left.\frac{\partial}{\partial\eta}
    W^\star(X_{U,k}\oplus\eta,X_B)\right|_{\eta=0}
    \in\mathbb R^6.
\]
The derivative is a column vector. Here and below,
$\partial_{\xi_U}W^\star(X_{U,k},X_B)$ is shorthand for this recentred
derivative at zero local increment.
\end{definition}

\emph{Coordinate and sensor conventions.}
For comparison, fix a nominal control pose and write
$X_U=\bar X_U\oplus\xi$, with $\xi=[\phi^\top,v^\top]^\top$.
In this fixed chart, a coordinate variation induces the infinitesimal body
rotation $\delta\theta^U=J_r(\phi)\delta\phi$, where $J_r$ is the right
Jacobian of $\mathrm{SO}(3)$~\cite{barfoot2017state}. By virtual work, a
physical control-port load with moment $\tau_U^U$ about $p_U$ and force
$f^W$ has coordinate generalized force
$Q_\xi=[(\tau_U^U)^\top J_r(\phi),\,(f^W)^\top]^\top$.
Definition~\ref{def:observation_map} instead recentres the chart at each
$X_{U,k}$ and differentiates at $\eta=0$, where $J_r(0)=I_3$.
This zero local increment does not require the control deformation $e_U$
to vanish. Derivatives of the logarithm in the potential are included in
the differentiation of $W^\star$.
Thus a change of moment origin includes a moment shift as well as an axis
rotation. Identifying this sensor load with the generalized force
conjugate to the commanded control also requires mechanical consistency
between the sensor port and the control-potential model. The observation model 
below assumes measurements available in this
conjugate representation.

\begin{assumption}
\label{ass:measurement_noise}
The {generalized-force observations in model coordinates} satisfy the following stochastic model,
\[
    y_k=h_k(X_B)+\varepsilon_k,
    \qquad
    \varepsilon_k\sim\mathcal N(0,\Sigma_w),
\]
where \(\Sigma_w\succ0\) is the covariance, and the noises are conditionally independent given
\(X_B\) and \(\mathcal U\).  Thus, the observed residual is
$
    r_k(X_B):=y_k-h_k(X_B).
$
\end{assumption}

\subsection{Coupled {Matrix Fisher--Gaussian} Prior}
\label{subsec:coupled_mfg_prior}

We place a coupled {Matrix Fisher--Gaussian} prior on
\(X_B=(R_B,p_B)\). The rotational marginal is a
{matrix Fisher (MF) distribution}.
We write \(R\sim\mathcal{MF}(F)\) when \(R\) has density
$
    p(R;F)
    = ({1}/{c(F)})
    \exp\!\bigl(\operatorname{tr}(F^\top R)\bigr), R\in\mathrm{SO}(3),
$
where \(c(F)\) is the normalizing constant.  If
\(F=USV^\top\) is a {proper singular value decomposition (SVD)} with
\(U,V\in\mathrm{SO}(3)\), \(S=\operatorname{diag}(s_1,s_2,s_3)\), and
\(s_1\ge s_2\ge |s_3|\), then a
mode is \(UV^\top\), and the singular values determine the concentration
along the principal rotational axes \cite{khatri1977von,lee2018bayesian}.

\begin{remark}
Unlike the standard singular value decomposition, the third diagonal entry \(s_3\) may be negative. The local rotational concentrations along the principal axes are characterized by \(s_2+s_3\), \(s_1+s_3\), and \(s_1+s_2\), respectively.
\end{remark}

\begin{definition}
\label{def:coupled_mfg}
A coupled MFG prior in precision form is specified by
$
    \Theta_0=(F_0,\mu_0,\Lambda_0,\Gamma_0),
$
where \(F_0\in\mathbb R^{3\times3}\), \(\mu_0\in\mathbb R^3\),
\(\Lambda_0\succ0\), and \(\Gamma_0\in\mathbb R^{3\times3}\).  Its density is
\begin{equation}
\label{eq:coupled_mfg_prior_density}
    p_0(R_B,p_B)
    =
    \frac{\sqrt{\det(\Lambda_0)}}{c(F_0)(2\pi)^{3/2}}
    \exp\!\left[-E_0(R_B,p_B)\right],
\end{equation}
with
$ E_0(R_B,p_B)
    =
    -\operatorname{tr}(F_0^\top R_B)
    +\frac12
    \|p_B-\mu_c(R_B)\|_{\Lambda_0}^2$, and $\mu_c(R_B)
    =
    \mu_0+\Gamma_0\nu_R $.
Here \(\|a\|_{\Lambda_0}^2=a^\top\Lambda_0 a\), and
$
    \nu_R=(QS-SQ^\top)^\vee,
    \quad
    Q=U^\top R_BV,
    \quad
    F_0=USV^\top .
$
We write
$
    X_B\sim\mathrm{MFG}(F_0,\mu_0,\Lambda_0,\Gamma_0).
$
The matrix \(\Gamma_0\) encodes
rotation--translation coupling, while \(\Lambda_0\) is the conditional
translation precision.
{The \emph{prior {centre}} is the selected joint mode
$(UV^\top,\mu_0)$ of the prior density $p_0$ in~\eqref{eq:coupled_mfg_prior_density}.}
\end{definition}

\begin{remark}
This is the covariance-form family of \cite{wang2022matrix} with
$\Gamma_0=P$ and $\Lambda_0=\Sigma_c^{-1}$; conditional precision is not
inverse marginal translation covariance. 
\end{remark}

\begin{definition}[Bayesian Estimation and Non-Conjugacy]
\label{def:Bayesian_posterior}
\leavevmode\par\nobreak\noindent
Under \Cref{ass:measurement_noise}, define the batch likelihood 
\begin{equation}
\label{eq:batch_likelihood}
    p(Y\mid X_B,\mathcal U)
    \propto
    \exp\!\left(
    -\frac12
    \sum_{k=1}^K
    \|r_k(X_B)\|_{\Sigma_w^{-1}}^2
    \right),
\end{equation}
where \(Y:=\{y_k\}_{k=1}^K\) and
\(\|a\|_{\Sigma_w^{-1}}^2=a^\top\Sigma_w^{-1}a\). The corresponding data
energy is
\[
    J_{\rm data}(X_B)
    :=
    \frac12
    \sum_{k=1}^K
    \|r_k(X_B)\|_{\Sigma_w^{-1}}^2 .
\]
Combining \eqref{eq:batch_likelihood} with the prior in
\eqref{eq:coupled_mfg_prior_density} gives the posterior 
\begin{equation}
    p(X_B\mid Y,\mathcal U)
    =
    \frac{
    p(Y\mid X_B,\mathcal U)p_0(X_B)}
    {
    \int_G p(Y\mid \widetilde X,\mathcal U)p_0(\widetilde X)\,
    d\widetilde X
    } .
    \label{eq:bayesian_posterior}
\end{equation}
For the distributional error analysis, denote the corresponding joint
posterior probability measure by
\begin{equation}
 \pi(dX_B)
 :=
 p(X_B\mid Y,\mathcal U)\,d\mu(X_B),
 \; X_B=(R_B,p_B)\in G,
\label{eq:joint-posterior-measure}
\end{equation}
where $\mu$ is the product of Haar measure on ${\mathrm{SO}(3)}$ and Lebesgue
measure on $\mathbb R^3$.
\end{definition}

The implicit-equilibrium energy $\sum_k\|y_k-h_k(X_B)\|_{\Sigma_w^{-1}}^2$
is generally not affine in the MFG sufficient statistics. Thus
\eqref{eq:bayesian_posterior} is not globally MFG-conjugate. 

\section{{Generalized-Force Sensitivity and Information Structure}}
\label{sec:wrench_sensitivity_information}

We differentiate the observation model of
Section~\ref{subsec:force_observation_model}.
Unless specified, $\|\cdot\|$ denotes an arbitrary fixed norm on
$\mathbb R^6$.

\subsection{Sensitivity of the Implicit Equilibrium Map}
\label{subsec:implicit_equilibrium_sensitivity}

Fix a nominal triple
$
    (\bar X_A,\bar X_U,\bar X_B)\in G\times G\times G
    $
satisfying the quasi-static equilibrium condition.
Let \(\xi_A,\xi_U,\xi_B\in\mathbb R^6\) denote local perturbations of
\(X_A,X_U,X_B\), respectively. Around the nominal triple, define the local
equilibrium operator as
$
    \Phi(\xi_A,\xi_U,\xi_B)
    :=
    \partial_{\xi_A}
    W(\bar X_A\oplus\xi_A,\bar X_U\oplus\xi_U,\bar X_B\oplus\xi_B).
$
At the nominal point, \(\Phi(0,0,0)=0\). We denote the Hessian blocks of the
potential $W$ by
\begin{equation*}
    (W_{\alpha\beta})_{ij}
    :=\left.
    \frac{\partial^2W}
    {\partial(\xi_\alpha)_i\,\partial(\xi_\beta)_j}
    \right|_{\xi_A=\xi_U=\xi_B=0},
    \; \alpha,\beta\in\{A,U,B\}.
\end{equation*}
Gradients are columns: rows of $W_{\alpha\beta}$ index $\xi_\alpha$
and columns index $\xi_\beta$. Assume the nominal triple satisfies
\Cref{ass:equilibrium_regular}, so $W_{AA}$ is nonsingular.

\begin{lemma}[First-order equilibrium sensitivity]
\label{lem:first_order_equilibrium}
\leavevmode\par\nobreak\noindent
Under \Cref{ass:equilibrium_regular}, the first-order variation
$\xi_A$ of the equilibrium state $X_A^\star(X_U,X_B)$,
induced by perturbations $\xi_U$ and $\xi_B$, satisfies
\begin{equation*}
    W_{AA}\xi_A+W_{AU}\xi_U+W_{AB}\xi_B=0,
\end{equation*}
where $W_{AA}$, $W_{AU}$, and $W_{AB}$ denote the
corresponding second-order derivative blocks of $W$
in local pose coordinates, evaluated at
$(X_A^\star(X_U,X_B),X_U,X_B)$.
Consequently, for fixed control input $X_U$
(i.e., $\xi_U=0$),
$
    \xi_A=-W_{AA}^{-1}W_{AB}\xi_B.
$
\end{lemma}

\begin{proof}
The proof is given in Appendix~\ref{app:equilibrium_sensitivity}.
\end{proof}

\subsection{{Reduced generalized-force sensitivity and local Fisher information}}
\label{subsec:reduced_wrench_sensitivity}

Recalling the reduced potential $ W^\star(X_U,X_B) $ in \Cref{def:observation_map} 
and the predicted {generalized force evaluated at equilibrium} $\widehat y_k(X_B)$, the residual $ r_k(X_B)$ has the special form \cite{yang2026adaptive}, 
$
    r_k(X_B)
    =
    y_k-\widehat y_k(X_B)
    =
    y_k+\partial_{\xi_U}W^\star(X_{U,k},X_B).
$

\begin{definition}[Residual Jacobian and Residual Merit]

For a nominal object pose \(\bar X_B\), define the residual Jacobian as
\begin{equation*}
    J_k
    :=
    \left.
    \frac{\partial r_k(\bar X_B\oplus\xi_B)}
    {\partial \xi_B}
    \right|_{\xi_B=0}
    \in\mathbb R^{6\times6}.
\end{equation*}
Furthermore, for \(K\) sets of contact measurements, consider each control input $X_{U,k}$,
$k=1,\ldots,K$. Let $\bar X_{A,k}^\star
= X_A^\star(X_{U,k},\bar X_B)$
be the corresponding equilibrium configuration.  Stack $r=(r_1^\top,\ldots,r_K^\top)^\top$ and
$J=(J_1^\top,\ldots,J_K^\top)^\top$, with
$\Sigma_K=I_K\otimes\Sigma_w$. Let
$
 \tilde r_k=\Sigma_w^{-1/2}r_k,$ and $
 \tilde r=(\tilde r_1^\top,\ldots,\tilde r_K^\top)^\top,
$
so the residual merit admits the form of 
\begin{equation}
    \rho(X_B)
    :=
    \|\tilde r(X_B)\|_2
    =
    \left(
    \sum_{k=1}^K
    r_k(X_B)^\top\Sigma_w^{-1}r_k(X_B)
    \right)^{1/2}.
    \label{eq:residual_merit}
\end{equation}

\end{definition}

\begin{theorem}[{Reduced Generalized-Force Jacobian and Fisher \mbox{Information}}]\label{thm:reduced_wrench_fisher_information}
Suppose \Cref{ass:equilibrium_regular} holds and, for each
$k=1,\ldots,K$, the residual map
$r_k:G\to\mathbb{R}^6$ is $C^2$ in a neighbourhood
of $\bar X_B$ under the right-perturbation
coordinate $\xi_B$.
Let $\bar r_k:=r_k(\bar X_B)$.
Then the following statements hold.
\begin{resultparts}
\item {\resultheading{Equilibrium-reduced residual Jacobian.}}
For each prescribed control input $X_{U,k}$,
the residual Jacobian has a reduced form
\begin{equation*}
    J_k
    =W_{UB}-W_{UA}W_{AA}^{-1}W_{AB},
\end{equation*}
where all derivative blocks are evaluated at
$(\bar X_{A,k}^\star,X_{U,k},\bar X_B)$
in the corresponding local pose coordinates. If the potential separates as
$
    W(X_A,X_U,X_B)
    =W_{\mathrm{ctrl}}(X_A,X_U)
     +W_{\mathrm{int}}(X_A,X_B),
$
then $W_{UB}=0$, and hence
$
    J_k=-W_{UA}W_{AA}^{-1}W_{AB}.
$
\item {\resultheading{Local Gauss--Newton data model.}}
The residual admits the expansion
\begin{equation*}
    r_k(\bar X_B\oplus\xi_B)
    =\bar r_k+J_k\xi_B
     +O(\|\xi_B\|^2).
\end{equation*}
Let
${J_{\mathrm{data}}}(X_B)
=\frac12\rho(X_B)^2$.
The Gauss--Newton quadratic approximation obtained
by linearizing the residuals is
\begin{equation}
\begin{aligned}
    {\widehat J_{\mathrm{data}}}
    (\bar X_B\oplus\xi_B)
    &:=\frac12\sum_{k=1}^K
    \|\bar r_k+J_k\xi_B\|_{\Sigma_w^{-1}}^2\\
    &={J_{\mathrm{data}}}(\bar X_B)
      +g_{\mathrm{data}}^\top\xi_B
      \\&\quad+\frac12\xi_B^\top H_{\mathrm{data}}\xi_B,
\end{aligned}
\label{eq:gauss_newton_data_model}
\end{equation}
where
{$
    g_{\mathrm{data}}=\sum_{k=1}^K J_k^\top\Sigma_w^{-1}\bar r_k,
    \quad
    H_{\mathrm{data}}=\sum_{k=1}^K J_k^\top\Sigma_w^{-1}J_k.
$}

\item {\resultheading{Fisher information.}}
{Under \Cref{ass:measurement_noise}, with independent Gaussian noise
of covariance $\Sigma_w\succ0$,} the measurement Fisher
information at pose $\bar X_B$ is
$
    I(\bar X_B)
    =J^\top\Sigma_K^{-1}J
    =H_{\mathrm{data}},
    \Sigma_K=I_K\otimes\Sigma_w.
$
The linearized observation model is identifiable
in all six pose directions if and only if
$J$ has full column rank, equivalently,
$I(\bar X_B)\succ0$.
If $I(\bar X_B)$ is rank deficient, there
exists a nonzero direction $\xi_B$ satisfying
$
    J_k\xi_B=0,
     k=\nobreak1,\ldots,K.
$
Along this direction, the linearized
observations have zero first-order sensitivity,
and the Gauss--Newton data model has zero
quadratic curvature.
\end{resultparts}
\end{theorem}

\begin{proof}
The proof is given in Appendix~\ref{app:reduced_information}.
\end{proof}

\begin{remark}[Gauss--Newton and exact data curvature]
Under the assumed additive Gaussian noise model,
$H_{\mathrm{data}}$ coincides with the measurement
Fisher information. However, the exact Hessian
of the nonlinear data energy satisfies
$
\nabla_{\xi_B}^2
{J_{\mathrm{data}}}(\bar X_B)
=H_{\mathrm{data}}+H_{\mathrm{curv}},
$
where $H_{\mathrm{curv}}$ contains residual-weighted
second derivatives. Gauss--Newton retains only the positive semidefinite local Fisher matrix
$H_{\rm data}$ \cite{nocedal2006numerical,kay1993fundamentals,jauffret2007observability}. {Thus, the Gauss--Newton curvature equals the exact data Hessian
at the nominal pose when $H_{\mathrm{curv}}=0$. It approximates that Hessian
when the residual-weighted second-derivative contribution is small.}
The matrix $H_{\mathrm{data}}$ excludes
the curvature contributed by the prior.
\end{remark}

Eliminating translation as a nuisance variable gives the following
Schur-complement diagnostic theorem \cite{kay1993fundamentals}.

\begin{theorem}[Translation-compensated rotational information]
\label{thm:rotational_information_score}
\leavevmode\par\nobreak\noindent
Let
$I=I(\bar X_B)=J^\top\Sigma_K^{-1}J\succeq0$
denote the measurement Fisher information in the
local pose coordinates
$\xi_B=(\phi^\top,v^\top)^\top$,
where $\phi,v\in\mathbb R^3$.
{Partition $I=(I_{ab})_{a,b\in\{\phi,v\}}$ into $3\times3$ blocks.}
Suppose $I_{vv}\succ0$ and define
$
I_{\mathrm{rot}}
:=\nobreak I_{\phi\phi}
-I_{\phi v}I_{vv}^{-1}I_{v\phi}.
$
Then $0\preceq I_{\mathrm{rot}}\preceq I_{\phi\phi}$,
and for every $\phi\in\mathbb R^3$,
{$\min_{v\in\mathbb R^3}\xi_B^\top I\xi_B
=\nobreak\phi^\top I_{\mathrm{rot}}\phi$, with unique minimizer
$v^\star(\phi)=-I_{vv}^{-1}I_{v\phi}\phi$.} Consequently, the score
\begin{equation}
\label{eq:rotational_information_score}
s_{\mathrm{rot}}
:=\lambda_{\min}(I_{\mathrm{rot}})
=\min_{\|\phi\|_2=1}
\phi^\top I_{\mathrm{rot}}\phi
\end{equation}
is the sharp lower bound satisfying
{$\min_{v\in\mathbb R^3}\xi_B^\top I\xi_B
\ge s_{\mathrm{rot}}\|\phi\|_2^2$.}
Moreover, $s_{\mathrm{rot}}>0$ if and only if
$I\succ0$. Any unit eigenvector $\phi_{\min}$
associated with $s_{\mathrm{rot}}$, together with
$v^\star(\phi_{\min})$, identifies a
least-informed translation-compensated
rotational perturbation.
\end{theorem}
\begin{proof}
The proof is given in Appendix~\ref{app:rotational_information}.
\end{proof}

\begin{remark}[Interpretation of rotational information]
The score $s_{\mathrm{rot}}=\lambda_{\min}(I_{\mathrm{rot}})$
provides a local E-optimality criterion for rotational
information after translation compensation \cite{pukelsheim2006optimal,atkinson2007optimum}.
A small $s_{\mathrm{rot}}$ indicates weak information
in at least one rotational direction, but does not
necessarily imply poor conditioning or explain a
particular estimation failure. Prior regularization
cannot restore identifiability from the measurements.
For nested control batches, $s_{\mathrm{rot}}(K)$
depends on the nominal pose, geometry, and noise model.
Details of the empirical selection of $K=10$ are provided
in Supplementary Section~\SMref{supp:s2-selection}.
\end{remark}

\section{{Local Quadratic Modelling and the Coupled MFG update}}
\label{sec:local_bayesian_closure}

In \Cref{sec:wrench_sensitivity_information}, we derive the local information structure induced by the {generalized-force observations} under quasi-static implicit equilibrium.  We now combine this data information with the coupled
MFG prior introduced in~\Cref{subsec:coupled_mfg_prior} to obtain a structure-preserving local
posterior update. Since the nonlinear {generalized-force} likelihood is not globally
conjugate to the coupled MFG prior, the update is constructed by matching a
local quadratic posterior model back to the coupled MFG family. The reconstruction is formulated through a
conditional-mean-linearized (CL) MFG surrogate. {{Theorem~\ref{thm:closed_form_reconstruction} defines the Coupled MFG update}.
Optional freezing through reuse of the prior statistic is
discussed separately in Section~\ref{sec:safeguarded_refinement} and
Appendix~\ref{app:opt-freezing}.} 

\subsection{Local Quadratic Posterior Model}
\label{subsec:local_quadratic_posterior}

The maximum a posteriori (MAP) objective is defined
as the negative log posterior density, up to an
additive constant independent of $X_B$:
$
    J_{\mathrm{MAP}}(X_B)
    := J_{\mathrm{data}}(X_B)+E_0(X_B),
$
{where $J_{\mathrm{data}}(X_B):=\rho(X_B)^2/2$ is the exact
noise-weighted data misfit defined using~\eqref{eq:residual_merit}.
Its Gauss--Newton approximation is $\widehat J_{\mathrm{data}}$
in~\eqref{eq:gauss_newton_data_model}, and $E_0$ is the negative log density
of the coupled MFG prior.}

\begin{assumption}[Prior regularity]
\label{ass:prior_regularity}
For each nominal pose $\bar X_B$ under consideration,
the local representation
$
    e_0(\xi_B)
    :=E_0(\bar X_B\oplus\xi_B)
$
is $C^3$ in a neighbourhood of $\xi_B=0$.
\end{assumption}

\begin{proposition}[Local quadratic posterior model]
\label{prop:local_quadratic_posterior}
\leavevmode\par\nobreak\noindent
Suppose~\Cref{ass:prior_regularity} and the hypotheses of~\Cref{thm:reduced_wrench_fisher_information}\textup{(ii)} hold. Define the exact local prior derivatives
$
g_{\mathrm{prior}}:=\nabla_{\xi_B}e_0(0),
\quad
H_{\mathrm{prior}}^{\mathrm{ex}}
:=\nabla_{\xi_B}^2e_0(0).
$
For any chosen symmetric prior-curvature
matrix $H_{\mathrm{prior}}$, define
\begin{equation}
\begin{aligned}
\widehat E_0(\bar X_B\oplus\xi_B)
:={}&E_0(\bar X_B)
+g_{\mathrm{prior}}^\top\xi_B\\
&+\frac12\xi_B^\top H_{\mathrm{prior}}\xi_B.
\end{aligned}
\label{eq:local_prior_model}
\end{equation}
Then
$\widehat E_0$ is a local quadratic model of the prior energy, and the 
discrepancy between the exact and approximated prior models is
\begin{equation}
\begin{aligned}
&E_0(\bar X_B\oplus\xi_B)
-\widehat E_0(\bar X_B\oplus\xi_B)\\
&\quad=
\frac12\xi_B^\top
\bigl(H_{\mathrm{prior}}^{\mathrm{ex}}
-H_{\mathrm{prior}}\bigr)\xi_B
+O(\|\xi_B\|^3).
\end{aligned}
\label{eq:prior_model_discrepancy}
\end{equation}
Combining the local quadratic prior model~\eqref{eq:local_prior_model}
with the Gauss--Newton data model~\eqref{eq:gauss_newton_data_model}
gives the local quadratic posterior model
\begin{equation}
\begin{aligned}
&\widehat J_{\mathrm{MAP}}
(\bar X_B\oplus\xi_B)\\
&\quad:=
\widehat J_{\mathrm{data}}
(\bar X_B\oplus\xi_B)
+\widehat E_0(\bar X_B\oplus\xi_B)\\
&\quad=
J_{\mathrm{MAP}}(\bar X_B)
+g_{\mathrm{post}}^\top\xi_B
+\frac12\xi_B^\top H_{\mathrm{post}}\xi_B,
\end{aligned}
\label{eq:local_posterior_quadratic}
\end{equation}
where $
g_{\mathrm{post}}
=g_{\mathrm{data}}+g_{\mathrm{prior}},$ and $
H_{\mathrm{post}}
=H_{\mathrm{data}}+H_{\mathrm{prior}}.
$
If $H_{\mathrm{post}}\succ0$, this quadratic
model has the unique minimizer
$
\xi_{\mathrm{quad}}^\star
=-H_{\mathrm{post}}^{-1}g_{\mathrm{post}},
$
and the associated Gaussian information model
in Euclidean local coordinates is
$
q_{\mathrm{quad}}
=\mathcal N\left(
-H_{\mathrm{post}}^{-1}g_{\mathrm{post}},
H_{\mathrm{post}}^{-1}
\right).
$
\end{proposition}

\begin{proof}
The proof is given in Appendix~\ref{app:local_posterior}.
\end{proof}

\begin{remark}
The above quadratic model combines Gauss--Newton data curvature with a prior curvature 
and therefore need not reproduce the exact Hessian of the nonlinear posterior energy. 
In particular, the prior approximation has a third-order remainder 
when \(H_{\mathrm{prior}}=H_{\mathrm{prior}}^{\mathrm{ex}}\); otherwise, 
a second-order discrepancy may remain.     
\end{remark}

\subsection{{Tangent-Space} Information Form of the Coupled MFG Prior}
\label{subsec:prior_information_form}

For the posterior update, we derive the prior
gradient and a curvature matrix in local coordinates.
Let
$
    \Theta_0=(F_0,\mu_0,\Lambda_0,\Gamma_0)
$
denote the coupled MFG prior parameters in~\Cref{def:coupled_mfg}.
Up to a pose-independent additive constant, the negative log prior is 
$
E_0(R_B,p_B)
=
-\operatorname{tr}(F_0^\top R_B)
+\frac12
\|p_B-\mu_0-\Gamma_0\nu_{R_B}\|_{\Lambda_0}^2.
$
Fix a nominal pose
$\bar X_B=(\bar R_B,\bar p_B)$ and introduce the local
product coordinates
$
R_B(\phi)=\bar R_B\exp(\widehat\phi),
p_B(v)=\bar p_B+v,
\xi=[\phi^{\top}, v^{\top}]^{\top}.
$
Let $F_0=USV^\top$ denote the proper SVD used in~\Cref{def:coupled_mfg}.  Then we define
$
\bar Q=U^\top\bar R_B V,$\quad $
\bar\nu=(\bar Q S-S\bar Q^\top)^\vee,$ and $
\bar e=\bar p_B-\mu_0-\Gamma_0\bar\nu,
$
and the rotational Jacobian
\begin{equation}
J_\nu
:=
\left.
\frac{\partial
\nu_{\bar R_B\exp(\widehat\phi)}}
{\partial\phi}
\right|_{\phi=0}
\in\mathbb R^{3\times3},
\label{eq:prior_statistic_jacobian}
\end{equation}
which denotes the Jacobian of \(\nu_{R_B}\) with respect to the rotational perturbation
\(\phi\). Specifically, for any $\phi\in\mathbb R^3$,
\begin{equation*}
J_\nu\phi
=
\left(
\bar Q\,\widehat{(V^\top\phi)}\,S
+
S\,\widehat{(V^\top\phi)}\,\bar Q^\top
\right)^\vee.
\end{equation*}
Its explicit expression under the proper SVD convention is derived
in Appendix~\ref{app:prior_derivation}.

\begin{proposition}[Prior gradient and Gauss--Newton curvature]
\label{prop:prior_information}
The exact gradient of the negative log prior at
$\bar X_B$, in the above local coordinates, is
\begin{equation*}
g_{\mathrm{prior}}
=
\begin{bmatrix}
-(\bar R_B^\top F_0-F_0^\top\bar R_B)^\vee
-J_\nu^\top\Gamma_0^\top\Lambda_0\bar e\\[2mm]
\Lambda_0\bar e
\end{bmatrix}.
\end{equation*}
Define
$
B_0=
\begin{bmatrix}
-\Gamma_0J_\nu&I_3
\end{bmatrix}.
$
The Gauss--Newton curvature matrix is
\begin{equation*}
H_{\mathrm{prior}}^{\mathrm{GN}}
=
\begin{bmatrix}
H_{\mathrm{MF}}&0\\
0&0
\end{bmatrix}
+
B_0^\top\Lambda_0B_0,
\end{equation*}
where
$
H_{\mathrm{MF}}
=
\operatorname{tr}(F_0^\top\bar R_B)I_3
-
\operatorname{sym}(F_0^\top\bar R_B)
$
{is the exact local curvature of the negative log density of the matrix Fisher distribution}, with
$\operatorname{sym}(A)=(A+A^\top)/2$.
Equivalently,
\begin{equation*}
H_{\mathrm{prior}}^{\mathrm{GN}}
=
\begin{bmatrix}
H_{\mathrm{MF}}
+J_\nu^\top\Gamma_0^\top
\Lambda_0\Gamma_0J_\nu
&
-J_\nu^\top\Gamma_0^\top\Lambda_0\\
-\Lambda_0\Gamma_0J_\nu
&
\Lambda_0
\end{bmatrix}.
\end{equation*}
\end{proposition}

\begin{proof}
The derivation of the Jacobian and the curvature
expressions is provided in Appendix~\ref{app:prior_derivation}.
\end{proof}

\begin{remark}[Gauss--Newton prior approximation]
\label{rem:prior-curvature}
Let ${H_{\mathrm{prior}}^{\mathrm{ex}}}$ denote the exact Hessian
of $E_0(\bar X_B\oplus\xi_B)$ at $\xi_B=0$.
The approximation $H_{\mathrm{prior}}^{\mathrm{GN}}$ in
\Cref{prop:prior_information} uses the first-order linearization of the
conditional translation mean
$\mu_c(R_B)=\mu_0+\Gamma_0\nu_{R_B}$ from~\Cref{def:coupled_mfg}.
Here
$\bar e=\bar p_B-\mu_c(\bar R_B)
=\bar p_B-\mu_0-\Gamma_0\bar\nu$
is the nominal conditional translation residual introduced in
Section~\ref{subsec:prior_information_form}.
The weighted least-squares Hessian identity
\cite[Chap.~10]{nocedal2006numerical} gives
\begin{equation}
\label{eq:prior-curvature-correction}
  {H_{\mathrm{prior}}^{\mathrm{ex}}}
  =
  H_{\mathrm{prior}}^{\mathrm{GN}}
  -\diag(C_\nu,0_{3\times3}),
\end{equation}
where
$
  (C_\nu)_{ij}
  =
  \bar e^\top\Lambda_0\Gamma_0
  \left.
  \frac{
    \partial^2
    \nu_{\bar R_B\exp(\widehat\phi)}
  }{
    \partial\phi_i\,\partial\phi_j
  }
  \right|_{\phi=0},
 i,j=1,2,3.
$
Here $\diag$ denotes the block-diagonal operator.
Thus, the translational and cross blocks are exact, while the rotational block omits a residual-weighted
second-derivative term.
\end{remark}

\begin{remark}[Conditional-mean linearization and properties of the Gauss--Newton prior curvature]
\label{rem:conditional-mean-curvature}
With $J_\nu$ defined in~\eqref{eq:prior_statistic_jacobian}, the conditional mean expansion is
$
  \mu_c\!\left(\bar R_B\exp(\widehat\phi)\right)
  =\mu_c(\bar R_B)+\Gamma_0J_\nu\phi+O(\|\phi\|^2).
$
This yields the first-order approximation \(\bar e+v-\Gamma_0J_\nu\phi\) 
to the conditional translation residual.
This expansion and the resulting quadratic construction are derived in
Appendix~\ref{app:prior_derivation},
\eqref{eq:B11_identity_new}--\eqref{eq:B15_new}.
The resulting conditional Gaussian curvature contribution is
${J_e^\top\Lambda_0J_e\succeq0}$, where
$J_e=B_0$, whereas the residual-weighted
correction in~\Cref{rem:prior-curvature} has no fixed sign.
This choice preserves the exact prior gradient and
{the curvature of the negative log density of the matrix Fisher distribution} while avoiding second
derivatives of the rotational statistic.
\end{remark}

\subsection{{Closed-form Coupled MFG update}}
\label{subsec:structure_preserving_update}

\begingroup
At a fixed nominal pose $\bar X_B=(\bar R_B,\bar p_B)$, write the
constant-free part of the local quadratic posterior
model~\eqref{eq:local_posterior_quadratic} as
\begin{equation}
q(\xi)=g^\top\xi+\tfrac12\xi^\top H\xi,
\qquad \xi=(\phi^\top,v^\top)^\top.
\label{eq:cmfg_quadratic_input}
\end{equation}
Here $g=g_{\mathrm{post}}$ and
$H=H_{\mathrm{post}}=H_{\mathrm{data}}+H_{\mathrm{prior}}$
are the gradient and curvature supplied by
\Cref{prop:local_quadratic_posterior}.
The Coupled MFG update constructs parameters $\Theta_{\mathrm{post}}$
whose conditional-mean-linearized (CL) surrogate energy has this gradient
and curvature. The same construction applies to any supplied
$g\in\mathbb R^6$ and $H\in\mathbb S^6$ satisfying the conditions below.
In particular, $H$ is the curvature of the chosen quadratic model.
With the Gauss--Newton prior curvature in~\Cref{prop:prior_information},
it need not equal the exact Hessian of the nonlinear posterior energy.
\endgroup

\begin{definition}[Conditional-mean-linearized (CL) MFG surrogate]
\label{def:cl_mfg_reconstruction}
Partition $g,H$ into rotational and translational components $g_\phi,g_v$
and blocks $H_{ab}$, $a,b\in\{\phi,v\}$.
For coupled MFG parameters $\Theta=(F,\mu,\Lambda,\Gamma)$ in the precision
form in~\Cref{def:coupled_mfg}, write $\nu_R(R;F)$ for the rotational
statistic formed using a fixed proper SVD choice of $F$, and set
\[
\bar\nu_F:=\nu_R(\bar R_B;F),
\qquad
J_{\nu,F}:=
\left.
\frac{\partial\nu_R(R_B(\phi);F)}
{\partial\phi}
\right|_{\phi=0}.
\]
The conditional-mean-linearized (CL) surrogate retains {the exact negative log density of the matrix Fisher distribution} and linearizes only the conditional mean:
\begin{equation}
    \label{eq:cl_mfg_surrogate}
\begin{aligned}
\widehat E_{\mathrm{CL}}(\xi;\Theta)
&=-\operatorname{tr}\bigl(F^\top R_B(\phi)\bigr)
+\frac12\|e_{\mathrm{CL}}(\xi;\Theta)\|_\Lambda^2,\\
e_{\mathrm{CL}}(\xi;\Theta)
&=\bar p_B+v-\mu-\Gamma(\bar\nu_F+J_{\nu,F}\phi).
\end{aligned}
\end{equation}
The local matching conditions require the CL surrogate to have the same
gradient $g$ and curvature $H$ as the quadratic model $q$ at $\xi=0$:
\begin{equation}
\label{eq:local_matching}
\nabla_\xi\widehat E_{\mathrm{CL}}(0;\Theta)
=g,
\qquad
\nabla_\xi^2\widehat E_{\mathrm{CL}}(0;\Theta)
=H.
\end{equation}
\end{definition}

{The following theorem gives the Coupled MFG update in closed form:
first the matrix Fisher parameter $F_{\mathrm{post}}$, then its rotational
statistic $\bar\nu_{\mathrm{post}}$ and Jacobian $J_{\nu,\mathrm{post}}$,
and finally the  triple of conditional Gaussian parameters
$(\Lambda_{\mathrm{post}},\Gamma_{\mathrm{post}},\mu_{\mathrm{post}})$.}

\begin{theorem}[{Closed-form Coupled MFG update}]
\label{thm:closed_form_reconstruction}
\begingroup
\leavevmode\par\nobreak\noindent
Use the notation in~\Cref{def:cl_mfg_reconstruction} and assume $H_{vv}\succ0$.
The resulting tuple
$\Theta_{\mathrm{post}}=(F_{\mathrm{post}},\allowbreak\mu_{\mathrm{post}},
\allowbreak\Lambda_{\mathrm{post}},\allowbreak\Gamma_{\mathrm{post}})$ is obtained as follows.
\begin{resultparts}
\item \resultheading{Matrix Fisher parameter.} Define the auxiliary quantities by
the gradient $g$ and Hessian $H$ of $q$ established
in~\eqref{eq:cmfg_quadratic_input} as
\begin{equation}
\begin{aligned}
a_{\mathrm{post}}
&=g_\phi-H_{\phi v}H_{vv}^{-1}g_v,\\
M_{\mathrm{post}}
&=H_{\phi\phi}-H_{\phi v}H_{vv}^{-1}H_{v\phi},\\
C_{\mathrm{post}}
&=\tfrac12\operatorname{tr}(M_{\mathrm{post}})I_3-M_{\mathrm{post}}.
\end{aligned}
\label{eq:cmfg_auxiliary_coefficients}
\end{equation}
We obtain the corresponding matrix Fisher parameter:
\begin{equation}
F_{\mathrm{post}}
=\bar R_B\left(C_{\mathrm{post}}
-\tfrac12\widehat{a_{\mathrm{post}}}\right).
\label{eq:cmfg_fisher_update}
\end{equation}
\item \resultheading{Rotational statistic and Jacobian.}
With $F_{\mathrm{post}}=U_{\mathrm{post}}S_{\mathrm{post}}V_{\mathrm{post}}^\top$
given by the fixed proper SVD selection, write
$\bar Q_{\mathrm{post}}=U_{\mathrm{post}}^\top\bar R_BV_{\mathrm{post}}$.
Then $\bar\nu_{\mathrm{post}}=\nu_R(\bar R_B;F_{\mathrm{post}})$ and
$J_{\nu,\mathrm{post}}=J_{\nu,F_{\mathrm{post}}}$ are given by
\begin{equation}
\begin{aligned}
\bar\nu_{\mathrm{post}}
&=(\bar Q_{\mathrm{post}}S_{\mathrm{post}}
-S_{\mathrm{post}}\bar Q_{\mathrm{post}}^\top)^\vee,\\
J_{\nu,\mathrm{post}}
&=\bar Q_{\mathrm{post}}
\bigl[\operatorname{tr}(S_{\mathrm{post}}\bar Q_{\mathrm{post}})I_3
-S_{\mathrm{post}}\bar Q_{\mathrm{post}}\bigr]V_{\mathrm{post}}^\top.
\end{aligned}
\label{eq:cmfg_statistic_update}
\end{equation}
\item \resultheading{Conditional Gaussian parameters.}
If $J_{\nu,\mathrm{post}}$ is nonsingular, the conditional Gaussian parameters are
\begin{equation}
\begin{aligned}
\Lambda_{\mathrm{post}}&=H_{vv},\\
\Gamma_{\mathrm{post}}
&=-H_{vv}^{-1}H_{v\phi}J_{\nu,\mathrm{post}}^{-1},\\
\mu_{\mathrm{post}}
&=\bar p_B-H_{vv}^{-1}g_v
-\Gamma_{\mathrm{post}}\bar\nu_{\mathrm{post}}.
\end{aligned}
\label{eq:cmfg_gaussian_update}
\end{equation}
\end{resultparts}
Under this nonsingularity condition, the constructed
$\Theta_{\mathrm{post}}$ is the unique tuple, under the fixed proper SVD
selection convention, satisfying the local matching conditions
in~\eqref{eq:local_matching}.
For $q$ in~\eqref{eq:cmfg_quadratic_input}, its CL surrogate satisfies
$
\widehat E_{\mathrm{CL}}(\xi;\Theta_{\mathrm{post}})
-\widehat E_{\mathrm{CL}}(0;\Theta_{\mathrm{post}})
=q(\xi)+O(\|\xi\|^3)$,  as $ \xi\to0.
$\par
\endgroup
\end{theorem}

\begingroup
\begin{proof}[Proof sketch]
For each fixed rotational increment $\phi$, we minimize the quadratic
model $q(\phi,v)$ with respect to the translation increment $v$.
This gives a model in $\phi$ alone, whose gradient $a_{\mathrm{post}}$
at $\phi=0$ and curvature $M_{\mathrm{post}}$ uniquely
determine $F_{\mathrm{post}}$ through~\eqref{eq:cmfg_auxiliary_coefficients}--\eqref{eq:cmfg_fisher_update}.
Using the fixed proper SVD of $F_{\mathrm{post}}$, we evaluate
$\bar\nu_{\mathrm{post}}$ and $J_{\nu,\mathrm{post}}$ at the nominal
rotation $\bar R_B$ according to~\eqref{eq:cmfg_statistic_update}.
Matching the translation and cross terms gives the remaining parameters,
uniquely because $J_{\nu,\mathrm{post}}$ is nonsingular.
Substitution verifies the local matching conditions~\eqref{eq:local_matching}.
The Gaussian part of the CL surrogate is quadratic, and its smooth
matrix Fisher part gives the third-order remainder.
The detailed proof is given in Appendix~\ref{app:closed_form_update}.
\end{proof}
\endgroup

\begin{remark}[{Scope of the closed-form Coupled MFG update}]
\begingroup
Here $H=\nabla_\xi^2q$ is the curvature matrix of the quadratic
model~\eqref{eq:cmfg_quadratic_input}. For the posterior
model~\eqref{eq:local_posterior_quadratic}, it should be taken as
$H_{\mathrm{post}}=H_{\mathrm{data}}+H_{\mathrm{prior}}$
from~\Cref{prop:local_quadratic_posterior}.
Under $H_{vv}\succ0$ and nonsingular $J_{\nu,\mathrm{post}}$,
Theorem~\ref{thm:closed_form_reconstruction} gives exact local matching
for the CL surrogate without requiring $H\succeq0$.
If $H\succ0$, the quadratic model~\eqref{eq:cmfg_quadratic_input}
has the unique minimizer $-H^{-1}g$.
\endgroup
\end{remark}

\begin{remark}[Local curvature of the actual MFG energy]
\label{rem:actual-mfg-curvature}
Let $\mathcal V_u$ be the negative log density of the MFG returned by
Theorem~\ref{thm:closed_form_reconstruction}, up to a constant, with respect to
$d\mu(X)=d\mu_{\mathrm H}(R)\,dp$, where $\mu_{\mathrm H}$ is normalized
Haar measure on $\mathrm{SO}(3)$ and $dp$ is Lebesgue measure on $\mathbb R^3$.
In local coordinates, set
$
\widetilde{\mathcal V}_u(\xi)
:=\mathcal V_u(\bar X_B\oplus\xi)
=-\operatorname{tr}\!(
 F_{\mathrm{post}}^\top \bar R_B\exp(\widehat\phi))
+\frac12
\|
 \bar p_B+v-\mu_{\mathrm{post}}
 -\Gamma_{\mathrm{post}}
  \nu_R(\bar R_B\exp(\widehat\phi);F_{\mathrm{post}})
\|_{\Lambda_{\mathrm{post}}}^2.
$
Define
\[
\begin{aligned}
B_j
&=
D_\phi^2
[
 \nu_R(\bar R_B\exp(\widehat\phi);F_{\mathrm{post}})
]_j\Big|_{\phi=0}, \\
C_u
&=
-\sum_{j=1}^{3}
  (\Gamma_{\mathrm{post}}^\top g_v)_j B_j .
\end{aligned}
\]
Then
$
\nabla_\xi\widetilde{\mathcal V}_u(0)
=g, $ and
$
\nabla_\xi^2\widetilde{\mathcal V}_u(0)
=H+\operatorname{diag}(C_u,0_{3\times3}).
$
The term $C_u$ is the curvature correction to the CL surrogate~\eqref{eq:cl_mfg_surrogate}.
For $\theta=\|\phi\|<\pi$, the chart volume is
$d\mu=j(\phi)\,d\xi$, with $j$ given in~\cite[Appendix~B, p.~1271]{wang2008nonparametric}.
The derivatives above concern the energy relative to $\mu$.
The coordinate-Lebesgue energy $\widetilde{\mathcal V}_u-\log j$
has the same gradient at zero but adds $I_3/6$ to its rotational
Hessian (A proof is given in  Appendix~\ref{app:actual-mfg-curvature}).
\end{remark}

\begin{corollary}[A sufficient condition for statistic-Jacobian nonsingularity]
\label{cor:statistic-jacobian-condition}
With the notation of Theorem~\ref{thm:closed_form_reconstruction},
the statistic Jacobian $J_{\nu,\mathrm{post}}$ can be written as
$U_{\mathrm{post}}^\top\bar R_B$$
(M_{\mathrm{post}}-\frac12\widehat a_{\mathrm{post}})$. If $M_{\mathrm{post}}$$\succ0$, then $\lambda_{\min}(M_{\mathrm{post}})>0$
and $\sigma_{\min}(J_{\nu,\mathrm{post}})\ge\lambda_{\min}(M_{\mathrm{post}})$.
Consequently, $J_{\nu,\mathrm{post}}$ is nonsingular and
$\|J_{\nu,\mathrm{post}}^{-1}\|_2\le\lambda_{\min}(M_{\mathrm{post}})^{-1}$.
In particular, $H\succ0$ is sufficient for the nonsingularity
condition in Theorem~\ref{thm:closed_form_reconstruction}. The theorem also
applies to indefinite $H$ whenever $H_{vv}\succ0$ and
$J_{\nu,\mathrm{post}}$ is nonsingular.
\end{corollary}
\begin{proof}
The proof is given in Appendix~\ref{app:statistic-jacobian-condition}.
\end{proof}

\section{Safeguarded MFG Posterior Refinement}
\label{sec:safeguarded_refinement}

In this section, we describe the single-pass and multi-pass
implementations of the closed-form Coupled MFG update. The single-pass
implementation constructs a local posterior approximation at a fixed
nominal pose. The multi-pass implementation repeats this construction at
successive nominal poses, using the original prior and measurement batch.
Algorithm~\ref{alg:single_pass_mfg_update} constructs one local MFG
approximation; Algorithm~\ref{alg:safeguarded_multipass_mfg_refinement}
repeats this construction. The latter adds the residual
acceptance test introduced in the Introduction. The single pass alone does
not compare successive nominal poses or guarantee residual decrease.

The closed-form Coupled MFG update follows the order in
Theorem~\ref{thm:closed_form_reconstruction}. Compute $F_{\mathrm{post}}$
from~\eqref{eq:cmfg_fisher_update}, evaluate its statistic and Jacobian
using~\eqref{eq:cmfg_statistic_update}, and obtain the remaining
parameters from~\eqref{eq:cmfg_gaussian_update}. The optional freezing
variant introduced below reuses the statistic and Jacobian of the original
prior, evaluated at the same nominal pose, within one reconstruction.

\paragraph{\textbf{Optional reuse of the prior statistic and Jacobian}}
Reusing the prior statistic and Jacobian is a computational 
approximation intended to reduce the local algebraic cost of 
reconstruction. It is not an assumption of 
Theorem~\ref{thm:closed_form_reconstruction}. For the same nominal pose
$\bar X_B=(\bar R_B,\bar p_B)$ and local quadratic input
$q(\xi)$
in~\eqref{eq:cmfg_quadratic_input}, the two constructions have the same
rotational marginal $R_B\sim\mathcal{MF}(F_{\mathrm{post}})$ and
conditional translation covariance
$\operatorname{Cov}(p_B\mid R_B)=\Lambda_{\mathrm{post}}^{-1}=H_{vv}^{-1}$.
The conditional translation means generally differ.
The Coupled MFG update~\eqref{eq:cmfg_gaussian_update} gives the mean
$\mu_{\mathrm{post}}+\Gamma_{\mathrm{post}}\nu_R(R_B;F_{\mathrm{post}})$.
The optional freezing variant~\eqref{eq:frozen_statistic_update} gives the mean
$\mu_{\mathrm{post}}^{\mathrm{frz}}+
\Gamma_{\mathrm{post}}^{\mathrm{frz}}\nu_R(R_B;F_{\mathrm{post}})$.
Proposition~\ref{prop:opt-freezing-certificate} bounds the 
discrepancy between the distributions produced by the freezing variant 
and the Coupled MFG update.  It therefore provides sufficient conditions for 
the optional freezing variant if one wants to meet a prescribed error tolerance. The Coupled MFG update is the theoretical reference. 
Supplementary Table~\SMref{supp:tab:runtime-repeats} reports repeated total-runtime measurements with and without freezing for the seed-44 dataset; Supplementary Sections~\SMref{app:opt-reconstruction-cost}--\SMref{app:opt-distribution-errors} compare the computational costs and distributional differences of the full and frozen reconstructions.

\subsection{{Single-pass implementation of the Coupled MFG update}}
\label{subsec:single_pass_update}

{We design Algorithm~\ref{alg:single_pass_mfg_update} by combining the
generalized-force observations and existing prior information specified by
\(\Theta_{\rm prior}\) to produce a local posterior approximation.}
Specifically, the algorithm takes the prior 
parameters \(\Theta_{\rm prior}\), a nominal pose \(\bar X_B\), 
a batch of control inputs \(\mathcal U\), a batch of measurements \(Y\), 
noise covariance \(\Sigma_w\), and equilibrium warm starts \(\mathcal X_A^{(0)}\) as input. 
It outputs the posterior parameters \(\Theta_{\rm post}\), 
the equilibrium batch \(\mathcal X_A^\star\), the residual merit \(\rho\), 
the rotational information score \(s_{\rm rot}\), the data gradient \(g_{\rm data}\), 
and the data curvature \(H_{\rm data}\).
It computes the local prior gradient and curvature at the nominal pose, 
 then iteratively solves for the implicit equilibrium states corresponding to each control 
 input in the batch, which is the most time-consuming step. It evaluates the residuals and Jacobians, accumulating the data gradient and curvature. 
Finally, it constructs the
 coupled MFG posterior parameters
{using the Coupled MFG update or its optional freezing variant.}

{The equilibrium, {generalized-force} and numerical sensitivity evaluations are
detailed in Appendix~\ref{app:auxiliary_inner}. The returned score
$s_{\rm rot}$ is a numerical counterpart of
\eqref{eq:rotational_information_score}, evaluated as specified in
Appendix~\ref{app:auxiliary_checks}. The undamped Schur-complement result
in~\Cref{thm:rotational_information_score} requires the data block
$I_{vv}\succ0$.}

\begin{algorithm}[!htbp]
\caption{{Single-pass Coupled MFG update with optional freezing}}
\label{alg:single_pass_mfg_update}
\small
\begin{algorithmic}[1]
\REQUIRE Prior parameters \(\Theta_{\rm prior}\); nominal pose
\(\bar X_B\); control batch \(\mathcal U=\{X_{U,k}\}_{k=1}^{K}\);
measurement batch \(Y=\{y_k\}_{k=1}^{K}\); noise covariance \(\Sigma_w\);
equilibrium warm starts \(\mathcal X_A^{(0)}\) as input.

\ENSURE Posterior parameters $\Theta_{\rm post}$; equilibrium batch
$\mathcal X_A^\star$; residual merit $\rho$; rotational information score
$s_{\rm rot}$; data gradient $g_{\rm data}$; data curvature $H_{\rm data}$.

\STATE Compute the local prior gradient and curvature
$g_{\rm prior}$ and $H_{\rm prior}$ at $\bar X_B$
from $\Theta_{\rm prior}$.

\STATE Initialize
$ g_{\rm data}\gets 0,\quad H_{\rm data}\gets 0 .
$

\FOR{$k=1,\ldots,K$}
    \STATE Solve the implicit equilibrium state by a damped Newton iteration:
    $
        X_{A,k}^{\star}
        \gets
        X_A^\star(X_{U,k},\bar X_B)
    $.
    \STATE Evaluate the residual $r_k(\bar X_B)$, 
$\hat y_k(\bar X_B)$, and $J_k$.
    \STATE Accumulate the data gradient and data curvature $g_{\rm data}$, $H_{\rm data} $.   
\ENDFOR

\STATE Form the local posterior information pair $g$ and $H$ using \Cref{prop:local_quadratic_posterior}.

\STATE {Construct $\Theta_{\rm post}$ {using the Coupled MFG update in
\Cref{thm:closed_form_reconstruction}. If the optional fixed-statistic}
variant is taken, use \Cref{cor:frozen_statistic_reconstruction}
with the scope and checks in Appendix~\ref{app:opt-freezing}.}

\STATE Compute the scalar whitened residual using \eqref{eq:residual_merit}:
$
\rho \leftarrow \rho(\bar X_B)
$

\STATE {Compute the numerical rotational information score
$s_{\rm rot}$ from $H_{\rm data}$ as specified in
Appendix~\ref{app:auxiliary_checks}.}

\RETURN \((\Theta_{\rm post},\mathcal X_A^\star,\rho,s_{\rm rot}, g_{\rm data}, H_{\rm data})\).
\end{algorithmic}
\end{algorithm}

\subsection{Residual-Safeguarded {Multi-pass} Refinement}
\label{subsec:safeguarded_outer_refinement}

Keeping the original prior and observation
batch fixed is part of the refinement design. Algorithm~\ref{alg:safeguarded_multipass_mfg_refinement} reuses the fixed
initial prior $\Theta^{(0)}$ and batch $\mathcal D=(\mathcal U,Y,\Sigma_w)$
at every {nominal pose}, avoiding repeated assimilation of the same
data. {Here $\textsc{SP}$ stands for \emph{single pass} and
denotes Algorithm~\ref{alg:single_pass_mfg_update} with $\mathcal D$ fixed.
It takes a prior, a nominal pose and an equilibrium warm-start batch,
and returns the posterior parameters, the equilibrium batch, the residual, and 
the rotational information score. We denote the output of the 
single-pass routine at iteration $t$ as
$$
 \mathfrak S^{(t)}
 =(\Theta_{\rm post}^{(t)},\mathcal X_A^{\star,(t)},\rho_t,
   s_{\rm rot}^{(t)},g_{\rm data}^{(t)},H_{\rm data}^{(t)}).
$$

At iteration $t$, let
$F_{\rm post}^{(t)}=U_tS_tV_t^\top$ be the proper SVD.
The routine $\textsc{RecoverPose}$ returns a joint mode under the
reconstructed MFG distribution $\Theta_{\rm post}^{(t)}$:
$$
 \textsc{RecoverPose}(\Theta_{\rm post}^{(t)})
 =(R_{\rm mode}^{(t)},p_{\rm mode}^{(t)})
 =(U_tV_t^\top,\mu_{\rm post}^{(t)}).
$$
By \Cref{def:coupled_mfg}, the rotational marginal is maximized at
$U_tV_t^\top$, where
$\nu_R(U_tV_t^\top;F_{\rm post}^{(t)})=0$ and the conditional Gaussian
mode is $\mu_{\rm post}^{(t)}$.
$\textsc{Branches}$ uses this pose to form the full, translation-only
and rotation-only endpoints
$
 (R_{\rm mode}^{(t)},p_{\rm mode}^{(t)}),
 (\bar R_B^{(t)},p_{\rm mode}^{(t)}),$ and $
 (R_{\rm mode}^{(t)},\bar p_B^{(t)}).
$
Only the rotation-only branch uses reduced relaxation steps when
$s_{\rm rot}^{(t)}$ is below the supplied information threshold.

\begin{algorithm}[!htbp]
\caption{Residual-Safeguarded Recentred Local MFG Estimator}
\label{alg:safeguarded_multipass_mfg_refinement}
\footnotesize
\begin{algorithmic}[1]
\REQUIRE Initial nominal pose $\bar X_B^{(0)}$; fixed prior
$\Theta^{(0)}$; data $\mathcal D=(\mathcal U,Y,\Sigma_w)$; equilibrium warm
starts $\mathcal X_A^{(0)}$; candidate settings;
pose tolerances $\varepsilon_R,\varepsilon_p\geq0$;
residual tolerances $\varepsilon_{\rm stop}>\varepsilon_{\rm acc}\geq0$; maximum outer
iteration number \(T_{\max}\).
\ENSURE {Refined nominal pose} $X_B^\star$ and {associated MFG parameters} $\Theta^\star$.

\STATE $\mathfrak S^{(0)}
\gets\textsc{SP}(\Theta^{(0)},\bar X_B^{(0)},\mathcal X_A^{(0)})$.
\STATE Initialize
$(X_B^\star,\Theta^\star,\rho^\star)
\gets(\bar X_B^{(0)},\Theta_{\rm post}^{(0)},\rho_0)$.

\FOR{$t=0,\ldots,T_{\max}-1$}

    \STATE $\mathcal B_t\gets
    \textsc{Branches}(\bar X_B^{(t)},
    \textsc{RecoverPose}(\Theta_{\rm post}^{(t)}),s_{\rm rot}^{(t)})$.

    \STATE $\mathcal C_t\gets
    \textsc{Cand}(\bar X_B^{(t)},\mathfrak S^{(t)},\mathcal B_t,\mathcal D)$;
    retain only candidates with $\rho\le \rho_t-\varepsilon_{\rm acc}$.

    \IF{$\mathcal C_t=\emptyset$}
        \STATE $\mathcal C_t\gets
        \textsc{Fallback}(\bar X_B^{(t)},\mathfrak S^{(t)},\mathcal D)$.
    \ENDIF

    \IF{$\mathcal C_t=\emptyset$}
        \RETURN $(X_B^\star,\Theta^\star)$.
    \ENDIF

    \STATE Sort $\mathcal C_t$ by candidate residual.
    \STATE For candidates $(X_c,\mathcal X_{A,c}^\star,\rho_c)$ in that order,
    compute $\mathfrak S_{\rm try}\gets
    \textsc{SP}(\Theta^{(0)},X_c,\mathcal X_{A,c}^\star)$.
    \STATE Select the first candidate passing the posterior-health checks
    and $\rho_{\rm try}\le\rho_t-\varepsilon_{\rm acc}$.

    \IF{no candidate survives this final recomputation}
        \RETURN $(X_B^\star,\Theta^\star)$.
    \ENDIF

    \STATE $(\bar X_B^{(t+1)},\mathfrak S^{(t+1)})
    \gets(X_c,\mathfrak S_{\rm try})$ for the selected candidate.

    \STATE Update \((X_B^\star,\Theta^\star,\rho^\star)\) if
    \(\rho_{t+1}<\rho^\star\).

    \IF{\(\textsc{Stop}(\bar X_B^{(t)},\bar X_B^{(t+1)},\rho_t,\rho_{t+1};
        \varepsilon_R,\varepsilon_p,\varepsilon_{\rm stop})\)}
        \STATE \textbf{break}.
    \ENDIF

\ENDFOR

\RETURN $(X_B^\star,\Theta^\star)$.
\end{algorithmic}
\end{algorithm}

For a branch endpoint $X_{B,b}=\bar X_B\oplus\xi_b$, define
\[
 \textsc{Relax}(\bar X_B,X_{B,b},\alpha)
 :=\bar X_B\oplus(\alpha\xi_b),\quad \alpha\in[0,1].
\]
For each trial, $\EvalRes(X_B,\mathcal D,\mathcal X_A^{\rm warm})$
solves the full equilibrium batch from
$\mathcal X_A^{\rm warm}$, evaluates $r_k(X_B)$, and returns the updated
batch, the scalar merit $\rho(X_B)$ in~\eqref{eq:residual_merit},
and the batch diagnostics used for screening. The branch search screens candidates using these diagnostics and their
residuals. If the pool is empty, a damped residual-descent fallback is
searched. Candidates are sorted by residual and then tested by a
$\textsc{SP}$ with the original prior $\Theta^{(0)}$.
For a supplied tolerance $\varepsilon_{\rm acc}\geq0$, the first
candidate passing the posterior-health checks and
$\rho_{\rm try}\leq\rho_t-\varepsilon_{\rm acc}$ is accepted;
if none passes, the best state is returned. The returned $X_B^\star$ minimizes the residual over the initial
and accepted nominal poses. 
The numerical checks and search routines are detailed in
Appendix~\ref{app:auxiliary_algorithms}.
}

After an accepted update, $\textsc{Stop}$ tests the rotation and
translation changes
$
 \delta_R^{(t)}
 =d_{\mathrm{SO}(3)}(\bar R_B^{(t)},\bar R_B^{(t+1)}) $ and $
 \delta_p^{(t)}
 =\|\bar p_B^{(t+1)}-\bar p_B^{(t)}\|_2,
$
together with the residual improvement $\rho_t-\rho_{t+1}$. {The stopping rule requires that
$\delta_R^{(t)}\leq\varepsilon_R$,
$\delta_p^{(t)}\leq\varepsilon_p$, and
$\rho_t-\rho_{t+1}\leq\varepsilon_{\rm stop}$ all hold,}
as specified in Appendix~\ref{app:auxiliary_checks}. The search also ends when no candidate survives the final checks, 
or the iteration limit is reached.
The residual acceptance condition gives a nonincreasing accepted
sequence, with strict decrease when $\varepsilon_{\rm acc}>0$.

In the following prior-mismatch Coupled MFG setting and the paired
benchmarks, we set $\varepsilon_R=10^{-6}$ rad,
$\varepsilon_p=10^{-6}$ m, $\varepsilon_{\rm acc}=10^{-6}$ and
$T_{\max}=20$. These Coupled MFG runs used
$\varepsilon_{\rm acc}$ also as the residual threshold in the joint
pose-change and residual stopping test, without a separate stopping threshold. All five prior-mismatch MFG runs reached
$T_{\max}$. The paired-study stopping implementation is supplied in
Supplementary Section~\SMref{supp:s3-cost}.
Settings for the additional
comparisons are given in Supplementary
Sections~\SMref{supp:s3-cost} and~\SMref{app:opt-reconstruction-cost}.
The following lemma states the residual--monotonicity.

\Needspace{3\baselineskip}
\begin{lemma}[Accepted residual monotonicity]
\label{lem:residual_monotonicity}
Let \(\rho(X_B)\) be the scalar whitened residual merit defined in
\eqref{eq:residual_merit}.
Assume that \(\varepsilon_{\rm acc}\ge0\), that each candidate residual is
recomputed with the same batch \((\mathcal U,Y,\Sigma_w)\) used to define
\(\rho\), and that Algorithm~\ref{alg:safeguarded_multipass_mfg_refinement}
accepts a candidate \(X_B^{(t+1)}\) only if
\(\rho(X_B^{(t+1)})\le\rho(X_B^{(t)})-\varepsilon_{\rm acc}\).
Then every accepted outer iteration satisfies
\begin{equation}
    \rho(X_B^{(t+1)})
    \le
    \rho(X_B^{(t)})-\varepsilon_{\rm acc}.
    \label{eq:monotone_residual}
\end{equation}
In particular, if \(\varepsilon_{\rm acc}=0\), the accepted residual sequence is
nonincreasing. If \(\varepsilon_{\rm acc}>0\), the number \(N_{\rm acc}\) of
strict accepted updates is finite and satisfies
$
    N_{\rm acc}
    \le
    \rho(X_B^{(0)})/\varepsilon_{\rm acc}
$
\end{lemma}
\begin{proof}
The proof is given in Appendix~\ref{app:residual_monotonicity}.
\end{proof}

\begin{remark}[Residual safeguard]
{Algorithm~\ref{alg:single_pass_mfg_update} supplies a local reconstruction
with numerical checks. Algorithm~\ref{alg:safeguarded_multipass_mfg_refinement}
adds a pose-level acceptance test: it recomputes the full equilibrium
batch and requires
$\rho_{\rm try}\leq\rho_t-\varepsilon_{\rm acc}$ before accepting a new
nominal pose. Here $\rho_{\rm try}$ is the whitened residual norm
in~\eqref{eq:residual_merit}, recomputed at the candidate nominal pose by
the final single-pass result. The value $\rho_t$ is the residual at the
current nominal pose. The guarantee in~\Cref{lem:residual_monotonicity}
concerns these accepted residuals.} {The numerical score $s_{\rm rot}$ approximates the diagnostic
in~\Cref{thm:rotational_information_score} and is used to assess weak
translation-compensated rotational sensitivity.}
\end{remark}

\section{\texorpdfstring{{Posterior Error Bounds and Experimental Evaluation}}{Posterior Error Bounds and Experimental Evaluation}}
\label{sec:evaluation}

\setlength{\textfloatsep}{12pt plus 2pt minus 2pt}

{%
We now investigate the accuracy and computational
cost of the local closed-form Coupled MFG
update for wrench-based pose estimation.
\Cref{subsec:original_false_confidence} tests refinement policy especially from an inaccurate,
overconfident prior setting.
\Cref{subsec:laplace_historical} compares pose estimates and residuals with a tangent-space Gaussian (Laplace-type) update~\cite{barfoot2014associating,schillings2020convergence}.
{Small pose errors do not establish accurate posterior probabilities,
and increasing the sample size cannot by itself correct a biased posterior
approximation. This motivates \Cref{sec:sec63-comparison}, which asks what
additional samples improve and what mechanical computation they require.
{Proposition~\ref{prop:sec63-compact-comparison}\textup{(i)} separates Coupled MFG
approximation bias from sampling variance. More independent draws from a
fixed MFG approximation reduce only the sampling variance of the sample-average
estimator defined in Definition~\ref{def:sec63-posterior-errors}.}
{Proposition~\ref{prop:sec63-compact-comparison}\textup{(ii)} gives an
$O(N^{-1/2})$ RMSE bound for $N$ particles under the assumptions stated in
Appendix~\ref{app:sec63-particles}.}
Corollary~\ref{cor:sec63-mechanical-work} shows that further draws from a
completed Coupled MFG approximation require no additional equilibrium solves or
reduced Jacobians, whereas particle equilibrium-solve counts grow linearly
with $N$ under the stated evaluation model and fixed particle paths.}{
\Cref{subsec:shape_pose_clouds} evaluates the numerical applicability of
the closed-form update in \Cref{thm:closed_form_reconstruction} to
geometries ranging from an ellipsoid to a smoothed Stanford bunny,
and examines how the number of wrench measurements affects posterior
concentration and pose-estimation accuracy.}
\Cref{subsec:original_physical} evaluates pose estimation with real robot
wrench measurements on a cube.
Supplementary Section~\SMref{supp:sec:geometry-contact} describes the geometry
and contact model. The information-based measurement-number selection rule
and a nested study with $K\in\{6,8,10,12,16,20,24\}$ measurements are
reported in Supplementary Sections~\SMref{supp:s2-selection}
and~\SMref{supp:s2-nested-results}, respectively.}

\subsection{Synthetic Prior-Mismatch Stress Test}
\label{subsec:original_false_confidence}
\begingroup
\captionsetup{hypcap=false}
\begin{table*}[!tp]
\centering
\makebox[\linewidth][c]{%
\begin{minipage}{\linewidth}
\renewcommand{\baselinestretch}{1}\selectfont
\begingroup
\caption{{Prior-mismatch stress test with $K=10$ {generalized-force}
observations. Entries are sample means $\pm$ sample standard deviations
over five noise seeds.}}
\label{tab:original_false_confidence}
\centering
\begingroup
\normalsize
\setlength{\tabcolsep}{1.5pt}
\renewcommand{\arraystretch}{1.08}
\begin{tabular*}{\linewidth}{@{\extracolsep{\fill}}*{12}{c}@{}}
\toprule
\multicolumn{3}{c}{\makecell{{Rotation error}\\
{$\downarrow\,\bar e_R$ (rad), \eqref{eq:pose_errors}}}}
& \multicolumn{3}{c}{\makecell{{Translation error}\\
{$\downarrow\,\bar e_p$ ($10^{-3}$ m), \eqref{eq:pose_errors}}}}
& \multicolumn{3}{c}{\makecell{{Residual merit}\\
{$\downarrow\,\bar\rho$, \eqref{eq:residual_merit}}}}
& \multicolumn{3}{c}{\makecell{{Rotational information}\\
{$\uparrow\,\bar s_{\rm rot}$, \eqref{eq:rotational_information_score}}}} \\
\cmidrule(lr){1-3}\cmidrule(lr){4-6}\cmidrule(lr){7-9}\cmidrule(lr){10-12}
Alg.~\ref{alg:single_pass_mfg_update} & {\textbf{Alg.~\ref{alg:safeguarded_multipass_mfg_refinement}}} & {$\downarrow$ (\%)}
& Alg.~\ref{alg:single_pass_mfg_update} & {\textbf{Alg.~\ref{alg:safeguarded_multipass_mfg_refinement}}} & {$\downarrow$ (\%)}
& Alg.~\ref{alg:single_pass_mfg_update} & {\textbf{Alg.~\ref{alg:safeguarded_multipass_mfg_refinement}}} & {$\downarrow$ (\%)}
& Alg.~\ref{alg:single_pass_mfg_update} & {\textbf{Alg.~\ref{alg:safeguarded_multipass_mfg_refinement}}} & {$\uparrow$ ($\times$)} \\
\midrule
$0.2926$ & $\bm{0.2459}$ & $15.87$
& $35.365$ & {$\bm{1.078}$} & {$96.95$}
& $17.890$ & {$\bm{6.778}$} & {$62.09$}
& $0.299$ & {$\bm{9.124}$} & {$30.52$} \\
$\pm0.00797$ & $\pm0.00853$ & $\pm5.19$
& $\pm1.59$ & {$\pm0.164$} & {$\pm0.44$}
& $\pm0.711$ & {$\pm0.510$} & {$\pm2.89$}
& $\pm0.000$ & {$\pm0.985$} & {$\pm3.30$} \\
\bottomrule
\end{tabular*}
\endgroup
\par\vspace{3pt}
{\normalsize\raggedright
{\emph{Note:} Bold entries mark the better mean in each pair.
For each seed, error and residual reductions are
$100(x_{\mathrm{Alg1}}-x_{\mathrm{Alg2}})/x_{\mathrm{Alg1}}$;
the score gain is $x_{\mathrm{Alg2}}/x_{\mathrm{Alg1}}$.}\par}
\endgroup
\end{minipage}%
}
\end{table*}
\endgroup

\begingroup
\captionsetup{hypcap=false}
\begin{figure*}[!tp]
\centering
\makebox[\linewidth][c]{%
\begin{minipage}{\linewidth}
\renewcommand{\baselinestretch}{1}\selectfont
\centering
\includegraphics[width=0.88\linewidth]{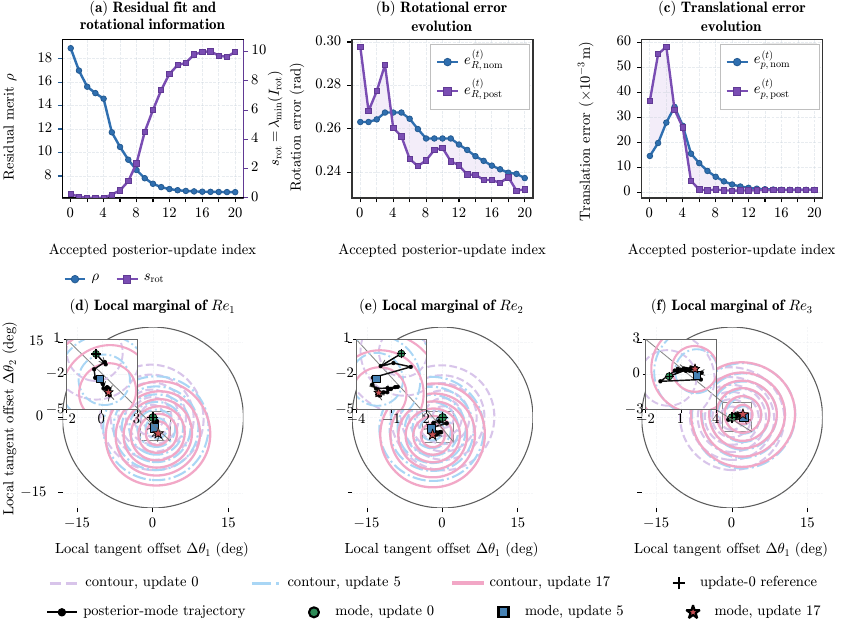}
\caption{{Refinement dynamics and rotational-marginal evolution
in the seed-44 stress test.
Panels (a)--(c) show the residual, rotational information and pose errors.
Panels (d)--(f) show local tangent-plane density contours for
$Re_1$, $Re_2$ and $Re_3$, estimated from {samples from the matrix Fisher distribution} and
normalized separately for each axis.
Contours and coloured mode markers identify updates 0, 5 and 17, and
the {MFG posterior-mode trajectory} covers updates 0--20.
Plus signs mark the update-0 projected mode directions at the chart origins.
Insets enlarge the central migration regions.}}
\label{fig:original_refinement_dynamics}
\label{fig:rotational_marginal_evolution}
\end{minipage}%
}
\end{figure*}
\endgroup

The results in this stress test use the optional freezing
(fixed-statistic) implementation.
Timing and output-distribution comparisons are reported in
Supplementary Sections~\SMref{app:opt-reconstruction-cost}--\SMref{app:opt-distribution-errors}, 
and the details of the freezing construction 
are given in Appendix~\ref{app:opt-freezing}. Supplementary Tables~\SMref{supp:tab:k-rotation}
and~\SMref{supp:tab:k-translation} report results with large initial
errors.

{%
\paragraph{\textbf{Experimental setup}}
{In this stress test, we use a prior} with a large pose error and high confidence,
which means that the distribution is sharply concentrated around an incorrect pose.
The object is represented by an ellipsoid-like field with semi-axes
$(0.04,0.03,0.05)$ m and shape exponents $(1,1)$.
We use an asymmetric contact patch of scale $0.010$ m and control
stiffness matrices $(K_R,K_p)=(8I_3,600I_3)$.
The {generalized-force observations evaluated at equilibrium} include independent Gaussian noise
with standard deviations $0.10$ N\,m for torque and $0.70$ N for force.
The true pose is $R_*=R_z(14^\circ)R_y(-6^\circ)R_x(4^\circ)$ and
$p_*=(0.015,-0.010,0.020)^\top$ m.} {%
We centre the MFG prior in~\eqref{eq:coupled_mfg_prior_density} at the
incorrect initial nominal pose
$\bar X_B^{(0)}=(\bar R_B^{(0)},\bar p_B^{(0)})$ and  further set
$
\bar R_B^{(0)}
 =R_*\exp\!\left(2\operatorname{Log}\!\bigl(
 R_z(-6^\circ)R_y(4^\circ)R_x(2^\circ)\bigr)\right),$
 $
\bar p_B^{(0)}
 =p_*+(-0.010,0.007,-0.008)^\top\ \mathrm m,$
 $
F_0=120\bar R_B^{(0)}, \mu_0=\bar p_B^{(0)},$ and $
\Lambda_0=12000I_3, \Gamma_0=0.
$
{Recentring changes the nominal pose, while this prior remains fixed.}
Here $F_0$ sets the rotational concentration and $\Lambda_0$ the
conditional translation precision, giving high confidence in the
incorrect pose.
The initial rotation and translation errors are $0.26312$ rad and
$14.595$ mm, respectively, and the experiment uses $K=10$ {generalized-force observations}.}

{%
\paragraph{\textbf{Results summary}}
Across the five noise seeds, Algorithm~\ref{alg:safeguarded_multipass_mfg_refinement}
reduces the rotation error, translation error and residual merit by
$15.87\pm5.19\%$, $96.95\pm0.44\%$ and $62.09\pm2.89\%$, respectively,
relative to Algorithm~\ref{alg:single_pass_mfg_update}
(shown in Table~\ref{tab:original_false_confidence}).
The rotational information score increases by a factor of $30.52\pm3.30$.
Relative to the initial pose, the rotation and translation errors decrease
by $6.55\pm3.24\%$ and $92.61\pm1.12\%$.
Thus, recentring improves the local {generalized-force} fit and reduces pose errors
in this test despite a sharply concentrated, incorrect prior.} {%
{Figure~\ref{fig:original_refinement_dynamics} (a)--(c) follows the seed-44 run.
At update $t$, ``nom'' denotes the nominal pose
$\bar X_B^{(t)}$, while ``post'' denotes the joint mode
$(R_{\rm mode}^{(t)},p_{\rm mode}^{(t)})$
recovered from that update's MFG by $\textsc{RecoverPose}$
in~\Cref{subsec:safeguarded_outer_refinement}.
Using the error definitions in~\eqref{eq:pose_errors}, the plotted errors are
$
e_{R,\rm nom}^{(t)}=d_{\mathrm{SO}(3)}(R_*,\bar R_B^{(t)}),
e_{p,\rm nom}^{(t)}=\|\bar p_B^{(t)}-p_*\|_2,
e_{R,\rm post}^{(t)}=d_{\mathrm{SO}(3)}(R_*,R_{\rm mode}^{(t)}),
e_{p,\rm post}^{(t)}=\|p_{\rm mode}^{(t)}-p_*\|_2.
$
The blue curves evaluate the nominal poses, and the purple curves evaluate
the corresponding Coupled MFG modes. A recovered mode is a trial endpoint and
becomes a new nominal pose only after selection and acceptance.
The residual falls as rotational information rises, but the pose
errors need not decrease at each step: the safeguard tests the residual
at the nominal pose. Panels (d)--(f) show the accompanying shift and
reshaping of the rotational marginals,  where the insets magnify the central
migration region.}}

\subsection{\texorpdfstring{{Comparison with a Tangent-Space Gaussian Update}}{Comparison with a Tangent-Space Gaussian Update}}
\label{subsec:laplace_historical}

{%
We compare the proposed framework
{(Algorithms~\ref{alg:single_pass_mfg_update} and~\ref{alg:safeguarded_multipass_mfg_refinement})}
with a tangent-space Gaussian point update under the prior--mismatch setting as in \Cref{subsec:original_false_confidence}.  
In this comparison, the proposed framework gives lower mean pose errors and residuals than the Laplace-type update.
This Laplace-type update uses a damped version of the local quadratic
posterior model in~\Cref{prop:local_quadratic_posterior}, following the
ideas of Gaussian pose uncertainty and Laplace approximation
\cite{barfoot2014associating,schillings2020convergence}.

\paragraph{\textbf{Experimental setup}}
Our implementation recomputes the model at each nominal pose and tests
step lengths $\alpha\in\{1,0.5,0.25,0.1\}$, accepting a step only when
the recomputed noise-weighted residual decreases by more than $10^{-8}$.
The Gaussian baseline stops when the relaxed step norm is below $10^{-6}$, or
no trial is accepted.}
The comparison concerns pose estimates and residuals. Both methods use $K=10$ observations, five noise seeds and the
prior-mismatch configuration in~\Cref{subsec:original_false_confidence}.
The initial errors are $e_R=0.26312$ rad and $e_p=14.595$ mm.
The maximum iteration budget is $T_{\max}=20$ passes.

\begin{table*}[!tp]
\centering
\caption{{Pose-estimation results under prior mismatch.}
Entries are sample means $\pm$ sample standard deviations over five
noise seeds. Positive $\Delta e_R$ and $\Delta e_p$ indicate error
reductions relative to the initial pose.}
\label{tab:laplace_historical}
\normalsize
\setlength{\tabcolsep}{3pt}
\renewcommand{\arraystretch}{1.12}
\begin{tabular*}{\linewidth}{@{\extracolsep{\fill}}lccccc@{}}
\toprule
{Method}
& \makecell{{$\downarrow e_R$}\\{(rad)}}
& \makecell{{$\Delta e_R$}\\{(\%)}}
& \makecell{{$\downarrow e_p$}\\{(mm)}}
& \makecell{{$\Delta e_p$}\\{(\%)}}
& {$\downarrow\rho$} \\
\midrule
\makecell[l]{{Tangent-space Gaussian}\\
{(Laplace-type)~\cite{barfoot2014associating,schillings2020convergence}}}
& \makecell{$0.311$\\$\pm0.023$}
& \makecell{{$-18.2$}\\{$\pm8.8$}}
& \makecell{$8.56$\\$\pm1.50$}
& \makecell{{$41.4$}\\{$\pm10.3$}}
& \makecell{$12.75$\\$\pm1.94$} \\
\addlinespace[3pt]
{\textbf{Proposed framework}}
& \makecell{{$\bm{0.246}$}\\$\pm0.009$}
& \makecell{{$\bm{6.6}$}\\{$\pm3.2$}}
& \makecell{{$\bm{1.08}$}\\$\pm0.16$}
& \makecell{{$\bm{92.6}$}\\{$\pm1.1$}}
& \makecell{{$\bm{6.78}$}\\$\pm0.51$} \\
\bottomrule
\end{tabular*}
\par\vspace{3pt}
\begin{minipage}{\linewidth}
\normalsize\raggedright
{\emph{Note:} {Downward arrows indicate that smaller
$e_R$, $e_p$ and $\rho$ values are better.} For either pose error,
$\Delta e=100(e^{(0)}-e)/e^{(0)}$ is computed for each seed before
averaging. Pose errors and residuals are defined in
\eqref{eq:pose_errors} and~\eqref{eq:residual_merit}.
Bold means identify the better result.}
\end{minipage}
\end{table*}

{%
\paragraph{\textbf{Results summary}}
Table~\ref{tab:laplace_historical} shows that the Gaussian update reduces
mean translation error but increases mean rotation error relative to the
initial pose. The proposed framework reduces both mean pose errors and gives a lower mean
residual, while using more refinement passes. These results support
improved pose estimation under the tested prior-mismatch configuration.}

\subsection{Posterior Approximation and Computational Performance}
\label{sec:sec63-frozen-particle}
\label{sec:sec63-comparison}

{Here, particle methods repeatedly solve the equilibrium problem to
evaluate likelihoods at proposed poses, whereas a fitted MFG distribution
can be sampled directly without further equilibrium solves.
This motivates our comparison of approximation accuracy and computational cost.
We compare the proposed closed-form update framework with resample--move particle
filtering (RM-PF)~\cite{gilks2001resamplemove} and tempered sequential
Monte Carlo (SMC)~\cite{delmoral2006smc}.
RM-PF adds observations through likelihood bridges, resampling and MCMC
moves. Tempered SMC increases the full-batch likelihood power from zero
to one. The bounds concern a common posterior target under the
assumptions stated in each result. The experiment compares the three
methods using fixed sample allocations and the same numerical likelihood.
Results are shown in Table~\ref{tab:sec63-main-results}.}

\paragraph{\textbf{Preliminaries}}
\begingroup
Let $\pi$ be the joint posterior defined in~\eqref{eq:joint-posterior-measure},
and assume the hypotheses of Theorem~\ref{thm:closed_form_reconstruction}.
The Coupled MFG update gives the distribution
$
 Q_u=\operatorname{MFG}(F_{\mathrm{post}},\mu_{\mathrm{post}},
                       \Lambda_{\mathrm{post}},\Gamma_{\mathrm{post}}),$
 where $ \Lambda_{\mathrm{post}}\succ0,
$
with parameters given by~\eqref{eq:cmfg_fisher_update}
and~\eqref{eq:cmfg_gaussian_update}.
\endgroup
RM-PF \cite{gilks2001resamplemove} and tempered SMC
\cite{delmoral2006smc} represent the posterior by weighted poses:
$
\widehat\pi_N:=\sum_{i=1}^Nw_i\delta_{\widetilde X_i}, w_i\geq0,\sum_{i=1}^Nw_i=1.
$
Here $\widetilde X_i=(\widetilde R_i,\widetilde p_i)$ is a sampled pose,
$w_i$ is its weight, and $\delta_{\widetilde X_i}$ puts unit probability
at that pose.

\begingroup
\begin{definition}[Posterior approximation error]
\label{def:sec63-posterior-errors}
For the pose distribution $Q_u$ on $G$, $Q_u(A)$ denotes the probability
that the pose lies in a measurable region $A\subseteq G$.
For a measurable map $f:G\to[0,1]$, write
$Q_u(f):=\int_G f\,dQ_u$; in particular, $Q_u(\mathbf1_A)=Q_u(A)$.
The Total Variation (TV) distance from the target posterior $\pi$ is
\[
 d_{\mathrm{TV}}(Q_u,\pi)
 :=\sup_A|Q_u(A)-\pi(A)|
 =\sup_{0\leq f\leq1}|Q_u(f)-\pi(f)|.
\]
The suprema range over measurable pose regions $A$ and measurable maps
$f:G\to[0,1]$, respectively. {For the fixed $Q_u$, every such $f$ satisfies
$|Q_u(f)-\pi(f)|\leq d_{\mathrm{TV}}(Q_u,\pi)$.
Let $S$ be a positive integer and let $X_1,\ldots,X_S$ be independent and
identically distributed exact draws from the fixed law $Q_u$.
The MFG sample-average estimator of $Q_u(f)$ is
$\widehat Q_S(f):=S^{-1}\sum_{s=1}^S f(X_s)$.} For the particles, set
$\widehat\pi_N(f):=\sum_{i=1}^Nw_i f(\widetilde X_i)$. For either estimator
$\widehat I\in\{\widehat Q_S,\widehat\pi_N\}$,
define the root-mean-square error by
\begin{equation*}
 \operatorname{RMSE}_f(\widehat I)
 :=\sqrt{\mathbb E_{\mathrm{MC}}\!\left[
 (\widehat I(f)-\pi(f))^2\right]}.
\end{equation*}
Here, $\mathbb E_{\mathrm{MC}}$ denotes expectation over the
sampling randomness, with the observations being fixed.
For the MFG estimator, the distribution $Q_u$ is also held fixed.

\end{definition}
\endgroup

\paragraph{\textbf{Main results}}
\begingroup
\textbf{(i) Error decomposition.}
For the MFG sample-average estimator $\widehat Q_S(f)$ defined in
Definition~\ref{def:sec63-posterior-errors},
Proposition~\ref{prop:sec63-compact-comparison} separates squared
approximation bias from sampling variance. Increasing $S$ reduces
the sampling variance within the fixed $Q_u$, but does not remove
its approximation bias relative to $\pi$.
\textbf{(ii) Conditional posterior-distance bound.}
In Appendix~\ref{app:opt-gap}, Proposition~\ref{prop:opt-actual-law}
bounds $d_{\mathrm{TV}}(Q_u,\pi)$ using the centred energy-gap
condition~\eqref{eq:opt-direct-gap} on a local pose region
$D\subseteq G$, together with $Q_u(D^c)$ and $\pi(D^c)$,
the probability masses assigned outside $D$ by the reconstructed
Coupled MFG distribution and the target posterior, respectively.
The bound uses the actual negative log density of $Q_u$,
without a separate representation-discrepancy term.
For the full update of Theorem~\ref{thm:closed_form_reconstruction},
the actual MFG energy has the curvature correction described in
Remark~\ref{rem:actual-mfg-curvature}.
Proposition~\ref{prop:local-gap-actual-mfg} bounds this centred
energy gap under its stated assumptions, including local $C^3$
regularity of the energy difference, thereby supplying the local
term in the conditional TV bound.
\endgroup

\Needspace{5\baselineskip}
\begin{proposition}[{Error bounds for posterior expectation estimates}]
\label{prop:sec63-compact-comparison} 
Fix the observations and a measurable map $f:G\to[0,1]$.
\begin{resultparts}
\item {\resultheading{Coupled MFG.}}
\begingroup
Fix the Coupled MFG distribution $Q_u$ defined above.
For $S\geq1$ independent exact draws from this fixed law,
\begin{equation}
\begin{aligned}
 \operatorname{RMSE}_f(\widehat Q_S)^2
 &= [Q_u(f)-\pi(f)]^2+\frac{\operatorname{Var}_{Q_u}(f)}{S}\\
 &\leq d_{\mathrm{TV}}(Q_u,\pi)^2+\frac{1}{4S}.
\end{aligned}
\label{eq:sec63-mfg-risk}
\end{equation}
An explicit conditional bound for the TV term is
given in~\eqref{eq:sec63-total-bound} with $Q=Q_u$.
\endgroup

\item {\resultheading{Particle filtering.}}
For a particle approximation targeting the same $\pi$ and
satisfying the finite-stage assumptions in
Appendix~\ref{app:sec63-particles}, there exists a finite
constant $C_{\mathrm{path}}$, independent of $N$ and potentially different
for RM-PF and SMC, such that
\begin{equation}
 \operatorname{RMSE}_f(\widehat\pi_N)
 \leq C_{\mathrm{path}}N^{-1/2},
 \qquad N\geq1.
\label{eq:sec63-particle-risk}
\end{equation}
\end{resultparts}
\end{proposition}

\begin{proof}
{Part~\textup{(i)} is proved in
Appendix~\ref{app:sec63-posterior-proof}, and part~\textup{(ii)} in
Appendix~\ref{app:sec63-particles}.}
\end{proof}

\paragraph{\textbf{Computational comparison}}
{%
Here, $n_{\mathrm{eq}}$
counts equilibrium solves and $n_J$ additional reduced Jacobian
evaluations. Let $n_{\mathrm{rec}}$ count all full-batch MFG reconstructions
and $n_{\mathrm{res}}$ separate full-batch forward-only residual
checks.
For RM-PF, $M_k$ counts proposals per particle across all bridges for
observation $k$, and for a $T$-stage tempered SMC path, $M_t$ counts proposals
per particle at stage $t$.}

\begingroup
\begin{corollary}[Mechanical work for posterior estimation]
\label{cor:sec63-mechanical-work}
Fix the observations, the completed MFG reconstruction, and the particle
paths and move counts independently of $N$.
Under the full-evaluation model in
Appendix~\ref{app:sec63-work}, with derivative-free particle moves, the counts are
\begin{equation}
\begin{array}{c|cc}
 & n_{\mathrm{eq}} & n_J \\ \hline
 \text{Proposed framework}
  & K(n_{\mathrm{rec}}+n_{\mathrm{res}}) & Kn_{\mathrm{rec}} \\[1mm]
 \text{RM-PF~\cite{gilks2001resamplemove}}
  & N\bigl(K+\sum_{k=1}^{K}kM_k\bigr) & 0 \\[1mm]
 \text{Tempered SMC~\cite{delmoral2006smc}}
  & NK\bigl(1+\sum_{t=1}^{T}M_t\bigr) & 0 .
\end{array}
\label{eq:sec63-mechanical-counts}
\end{equation}
The MFG counts are independent of $S$, while the particle counts are linear in $N$.
\end{corollary}
\endgroup
\begin{proof}
{The work counts and their independence from $S$ are
proved in Appendix~\ref{app:sec63-work}.}
\end{proof}

Evaluating the new $3\times3$ statistic and Jacobian changes local
computational work, not the per-reconstruction mechanical counts in
\eqref{eq:sec63-mechanical-counts}. Optional freezing can nevertheless
change the refinement trajectory and hence $n_{\rm rec},n_{\rm res}$.

{%
We compared three methods in~\eqref{eq:sec63-mechanical-counts} on
{the fixed bank of 24 synthetic datasets with identifiers
$\mathcal C=\{0,\ldots,11\}\cup\{13,\ldots,24\}$ (excluding $12$)}, each with $K=10$
{generalized-force observations} and
{four runs per method per dataset ($96$ runs per method)}. All methods shared the same
prior, measurements, and equilibrium solver with fixed settings $\eta$.
{The numerical posterior $\pi_\eta$ is defined by~\eqref{eq:joint-posterior-measure},
replacing $h_k$ in~\eqref{eq:batch_likelihood} by the solver responses under a
common measurable equilibrium-selection rule and assuming a finite, positive
normalizing constant. Transferring error bounds from $\pi_\eta$ to the ideal
posterior $\pi$ additionally requires control of $d_{\rm TV}(\pi_\eta,\pi)$.}
The proposed framework
(Algorithm~\ref{alg:safeguarded_multipass_mfg_refinement}) used
$S=16\,384$ samples, RM-PF and tempered SMC each used $N=256$ particles.

The Coupled MFG estimations use the stopping
implementation in Supplementary Section~\SMref{supp:s3-cost},
with $\varepsilon_R=10^{-6}$ rad, $\varepsilon_p=10^{-6}$ m, and
$\varepsilon_{\rm acc}=10^{-6}$. 
For these 24 fixed inputs, fresh fits with
$\varepsilon_{\rm stop}\in\{10^{-5},10^{-4},10^{-3},10^{-2}\}$
gave identical accepted-pass counts, returned poses,
fitted distribution parameters, residuals, and counts of equilibrium solves
and Newton iterations in all 96 comparisons across these $\varepsilon_{\rm stop}$ values. These observations are
limited to the tested inputs and thresholds. Per-case records and paired comparisons are included in the supporting data. Particle settings and
equilibrium-solver initialization are described in Supplementary
Section~\SMref{supp:s3-cost}.}

{%
Table~\ref{tab:sec63-main-results} reports pose errors~\eqref{eq:pose_errors}
computed from the weighted intrinsic rotation mean and weighted
translation mean of each method's output samples. RM-PF gives the smallest
mean pose errors, while the proposed framework has the lowest measured
runtime at the chosen sample allocations. Its median construction uses 660 equilibrium solves and 660
additional reduced Jacobian evaluations. Consistent with
Corollary~\ref{cor:sec63-mechanical-work}, further MFG samples require
neither operation, whereas particle proposals require equilibrium much more often. The proposed framework is 
therefore more efficient in terms of the number of equilibrium
solves.}

\begin{figure*}[!tp]
\centering
\setlength{\abovecaptionskip}{5pt}
\setlength{\belowcaptionskip}{0pt}
\includegraphics[width=0.88\linewidth]{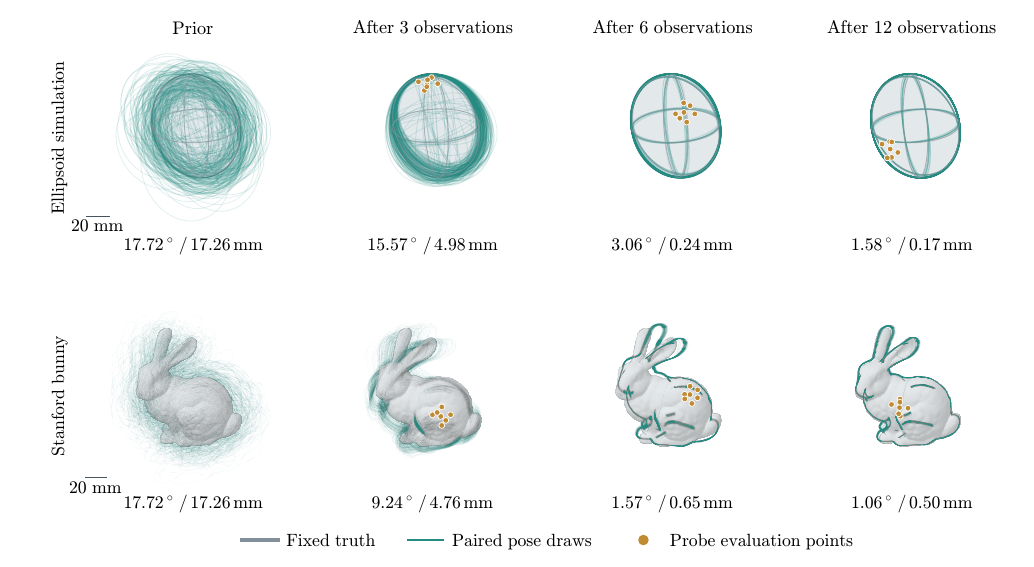}
\caption{{Joint pose clouds for the ellipsoid simulation
(top) and Stanford-bunny simulation (bottom). At stage $k>0$,
each cloud incorporates observations $1{:}k$, while the probe markers
show only observation $k$. Each panel displays 180 paired MFG
draws. Its values give rotation/position RMS about the estimated pose
using all 2,000 draws.}}
\label{fig:pose_cloud_geometries}
\end{figure*}
\begin{table}[!t]
\centering
\caption{Pose-estimation results and computational costs on 24
synthetic datasets, each with $K=10$ observations and four runs per method.}
\label{tab:sec63-main-results}
\renewcommand{\arraystretch}{1.9}
\normalsize
\setlength{\tabcolsep}{1.5pt}
\begin{tabular*}{\linewidth}{@{\extracolsep{\fill}}lrrrr@{}}
\toprule
& \multicolumn{2}{c}{Mean pose error}
& \multicolumn{2}{c}{Computational cost}\\
\cmidrule(lr){2-3}\cmidrule(l){4-5}
Method & \shortstack{Rotation\\(deg)}
& \shortstack{Translation\\(mm)}
& \shortstack{Full time\\(s)} & $n_{\mathrm{eq}}/K$\\
\midrule
\makecell[l]{Proposed\\framework} & 4.449 & 2.684 & 2.39 & 66.0\\
{RM-PF}~\cite{gilks2001resamplemove} & 4.028 & 1.298 & 251.50 & 32\,204.8\\
\makecell[l]{Tempered\\SMC~\cite{delmoral2006smc}} & 4.555 & 1.320 & 187.11 & 24\,832.0\\
\bottomrule
\end{tabular*}
\par\vspace{3pt}
\begin{minipage}{\linewidth}
\normalsize
{\emph{Note:} Pose errors are averaged equally over datasets and runs.
For time and work, we take the median of the four runs in each dataset,
then the median across datasets. Costs include construction, sampling, and posterior probability estimation.
The quantity $n_{\mathrm{eq}}/K$ is the equilibrium-solve count divided by
the number of observations.}
{The proposed-framework row uses the regularized fixed-statistic
implementation. RM-PF and SMC use the same numerical likelihood and
derivative-free MCMC moves.}
\end{minipage}
\end{table}

\paragraph{\textbf{Further discussion}}
The bias--variance identity used in
Proposition~\ref{prop:sec63-compact-comparison}\textup{(i)} also applies
to the fixed MFG distribution used in the sampling diagnostic.
Additional draws reduce sampling variance while leaving the
approximation bias unchanged. In the sampling diagnostic, increasing $S$
from 256 to $16\,384$ reduces mean-square error against independent
Monte Carlo estimates of the same MFG probabilities from
$3.779\times10^{-4}$ to $5.390\times10^{-6}$ without further mechanical
evaluations. Mean-square differences from the numerical SMC reference
probabilities remain near $2.3\times10^{-2}$.  The numerical SMC reference probabilities did not meet the stability criteria reported in Supplementary Section~\SMref{supp:s3-reference-checks}. Consequently, discrepancies from this reference are interpreted as descriptive
comparisons and do not provide a reliable measure of MFG approximation error relative to the exact
nonlinear posterior. This limitation does not affect the pose-estimation comparisons against known
ground truth or the computational-cost results.

This illustrates improved
integration within the fixed approximation. 
We provide the solver and particle settings in
Supplementary Section~\SMref{supp:s3-cost}, {query definitions and error metrics} in
Section~\SMref{supp:s3-probability-errors}, reference checks in
Section~\SMref{supp:s3-reference-checks}, and sampling diagnostics in
Section~\SMref{supp:s3-sampling-diagnostics}.

\subsection{Joint Pose-Cloud Visualization on Two Geometries}
\label{subsec:shape_pose_clouds}

We illustrate the joint pose approximation using a stationary ellipsoid
with semi-axes $(40,30,50)$ mm and a smoothed Stanford-bunny implicit
field derived from the Stanford Computer Graphics Laboratory
reconstruction~\cite{stanfordBunnyRepository}. Both examples use twelve simulated wrench observations, the seven-point
probe and mechanical coefficients in Supplementary
Table~\SMref{supp:tab:baseline}, and the full reconstruction in
Theorem~\ref{thm:closed_form_reconstruction}, without freezing. At stage $k$, Algorithms~\ref{alg:single_pass_mfg_update}
and~\ref{alg:safeguarded_multipass_mfg_refinement} use observations
$1{:}k$ with the fixed prior. The prior has $F_0=16R_0$,
$\Lambda_0=10^4I_3\,\mathrm{m}^{-2}$ and $\Gamma_0=0$.
The illustrative noise standard deviations are
$15\,\mu\mathrm{N\,m}$ for torque and $1.5\,\mathrm{mN}$ for force,
which differ from the benchmark settings. Commands are designed using
the synthetic truth and held fixed under alternative pose hypotheses. 
Supplementary Section~\SMref{supp:shape-clouds} gives the geometry and probing strategy.

For the ellipsoid, we implement the seven-point quasi-static
contact potential in Definition~\ref{def:potential}, using the analytic
field $F_B(c)=\sum_{j=1}^3(c_j/a_j)^2-1$ with the semi-axes given above.
Twelve rays in the object frame define surface contacts and
outward normals. Each commanded tool origin lies 6 mm inward from its
contact point, and its $z$-axis follows the normal. 
The commands are constructed
using the synthetic true pose and held fixed during inference.
Supplementary Section~\SMref{supp:shape-geometry-controls} lists the
ordered directions, the complete command construction, and the simulation
code and saved-data files. The experiment uses these controls
and noise seed 44.

\begin{table}[!tb]
\centering
\caption{Pose-cloud dispersion and {pose-estimation error} for the 
ellipsoid and bunny simulation.
RMS values use all 2,000 paired draws per stage.}
\label{tab:shape_cloud_diagnostics}
\normalsize
\setlength{\tabcolsep}{1.3pt}
\renewcommand{\arraystretch}{1.58}
\begin{tabular*}{\linewidth}{@{\extracolsep{\fill}}lcrrrr@{}}
\toprule
 & & \multicolumn{2}{c}{\makecell{Sample dispersion\\(RMS)}}
 & \multicolumn{2}{c}{\makecell{Pose-estimation\\error}}\\
\cmidrule(lr){3-4}\cmidrule(lr){5-6}
Geometry & $k$ & \makecell{Rotation\\($^\circ$)} & \makecell{Translation\\(mm)}
    & {$e_R$ ($^\circ$)} & {$e_p$ (mm)}\\
\midrule
\multirow{4}{*}{Ellipsoid}
 & 0  & 17.721 & 17.260 & 10.770 & 8.124\\
 & 3  & 15.568 & 4.984  & 5.689  & 2.035\\
 & 6  & 3.058  & 0.236  & 3.076  & 0.162\\
 & 12 & 1.578  & 0.173  & 2.717  & 0.051\\
\midrule
\multirow{4}{*}{Bunny}
 & 0  & 17.721 & 17.260 & 10.770 & 8.124\\
 & 3  & 9.238  & 4.761  & 3.809  & 4.870\\
 & 6  & 1.575  & 0.651  & 10.560 & 3.288\\
 & 12 & 1.062  & 0.497  & 3.186  & 0.872\\
\bottomrule
\end{tabular*}
\par\smallskip
\begin{minipage}{\linewidth}
\normalsize
{\textit{Note:} The pose estimate is the joint mode
(maximum-density pose) of the constructed Coupled MFG distribution.
RMS measures sample dispersion around this estimate; $e_R$ and $e_p$
are its rotation and translation errors relative to the true pose,
as defined in~\eqref{eq:pose_errors}. Rotation deviations are geodesic
angles, and translation deviations are Euclidean distances.}
\end{minipage}
\end{table}
{For the bunny, the same contact potential uses a smooth implicit
field constructed from the Stanford Bunny triangle mesh, which combines
multiple aligned range scans~\cite{stanfordBunnyRepository}. We centre the mesh at
its bounding-box midpoint, align its original $+y$-axis with the body
$+z$-axis, scale its source height to 100 mm, and close the five holes
in the mesh by adding triangular patches. Signed mesh distance is sampled on a 0.8 mm grid and
Gaussian-smoothed with standard deviation 0.8 mm. We interpolate these
grid values using products of one-dimensional, degree-five B-spline basis
functions along the three coordinate axes~\cite{deboor2001splines}. This tensor-product quintic
B-spline defines a scalar field $s(c)$, giving
$F_B(c)=s(c)/(0.05\,\mathrm{m})$.
Twelve rays from an interior torso point define surface contact points on
$F_B=0$: the first three probe nearby regions, and the remaining nine
provide varied directions. Each commanded tool origin lies 3 mm inward
along the outward normal, which also defines its $z$-axis. Commands use
the synthetic true pose and remain fixed during inference; the noise
seed is 44. Supplementary Section~\SMref{supp:shape-geometry-controls}
specifies the geometry construction, control records, simulation code
and saved-data files.}

Figure~\ref{fig:pose_cloud_geometries} preserves the pairing between
matrix Fisher rotation draws and their conditional Gaussian
translations. {The gold markers denote the seven fixed probe
evaluation points at the data-generating equilibrium for the current
observation. Together they yield one six-component wrench and need not
lie on the object surface.
The first three observations use nearby but distinct probing configurations.
All panels in each row use the same viewing direction and spatial scale.}
At $k=3$, the position RMS about the {estimated pose}
is $4.984$ mm for the ellipsoid and $4.761$ mm for the bunny, whereas
the corresponding conditional translation RMS values
$\sqrt{\tr(\Lambda_3^{-1})}$ are $0.901$ mm and $0.547$ mm.
These differences illustrate the contribution of the rotation-dependent
conditional mean to the marginal position clouds.

Table~\ref{tab:shape_cloud_diagnostics} separates sample dispersion
from pose estimation error. For both geometries, the rotational and
translational RMS values decrease across the stages $k=0,3,6,12$.
This empirical narrowing of the fitted posterior clouds illustrates the
closed-form coupled MFG update in \Cref{thm:closed_form_reconstruction}
as wrench observations accumulate. After $K=12$ observations, the
errors are $e_R=2.717^\circ$ and $e_p=0.051$ mm for the
ellipsoid, and $e_R=3.186^\circ$ and $e_p=0.872$ mm for the bunny. Both components
are below their prior-stage values. These single-realization results use
geometry-specific controls. Concentration and accuracy remain distinct:
the bunny at $k=6$ has rotational RMS $1.575^\circ$ but
$e_R=10.560^\circ$, showing that a narrow local approximation can
remain inaccurately centred.

The ellipsoid is invariant under body-axis half-turns, so its cloud
describes the local branch favoured by the prior rather than global
identification of a unique oriented pose. The bunny extends the
illustration to a scan-derived geometry with field derivatives through
third order. Supplementary Section~\SMref{supp:shape-clouds} reports numerical checks of the geometry derivatives, equilibrium sensitivities, reconstruction formulas, and sampled moments.

\subsection{Real-World Validation}
\label{subsec:original_physical}

\begin{figure}[!ht]
\centering
\includegraphics[width=\linewidth,trim={60bp 1270bp 60bp 30bp},clip]{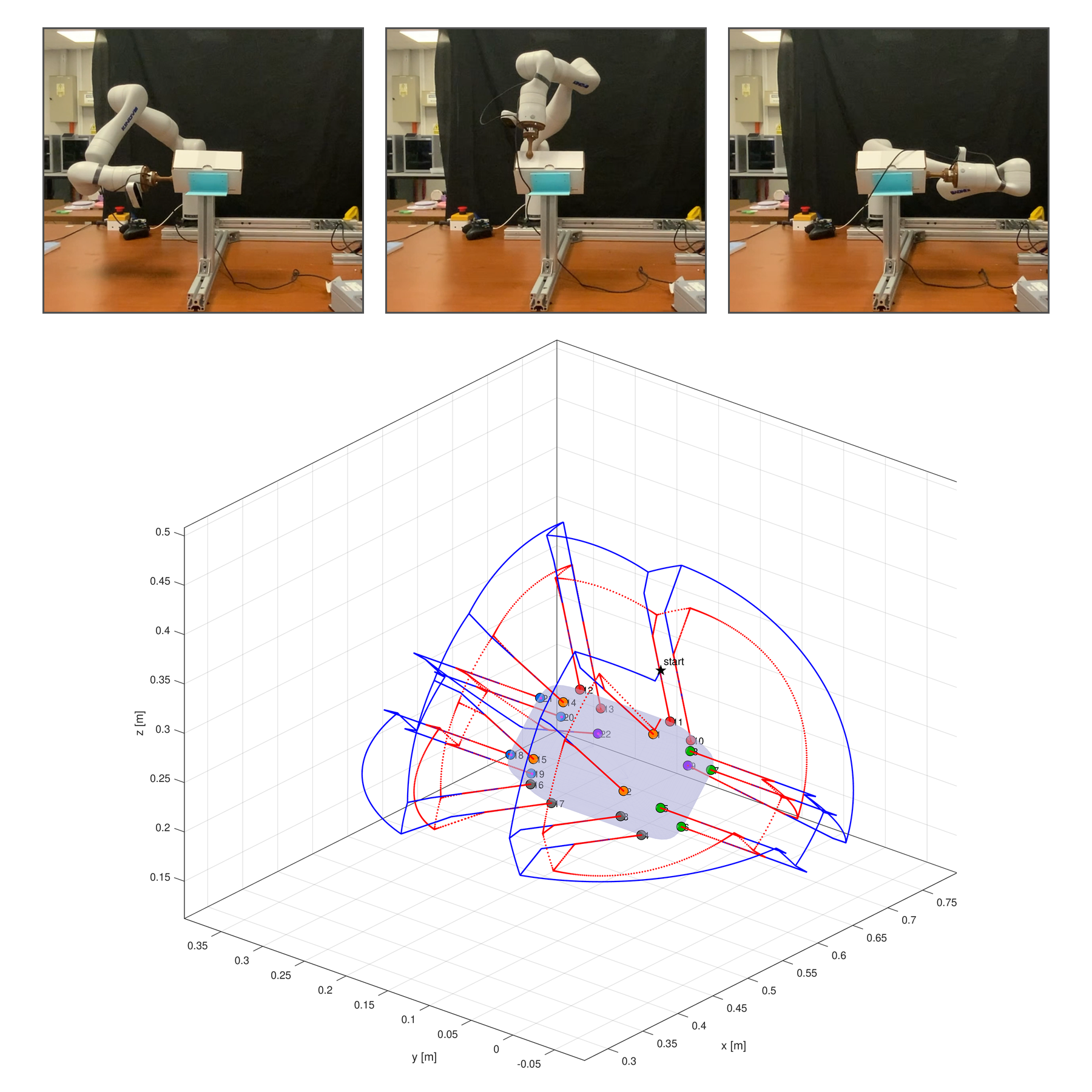}
\caption{Tilted-box probing video frames at different faces.
Supplementary Section~\SMref{supp:s4:apparatus} gives the apparatus
schematic and full probing trajectory, whose numbered markers
indicate the touch order.}
\label{fig:physical_setup}
\end{figure}

We probe a known 3D-printed tilted box with a Kinova Gen3
manipulator~\cite{kinovaGen3UserGuide} and ATI Mini40 force/torque
sensor~\cite{atiMini40} under quasi-static Cartesian impedance control,
with $(K_R,K_p)=(8I_3,600I_3)$. The object is modelled by the smooth
superquadric field. Algorithm~\ref{alg:safeguarded_multipass_mfg_refinement}
uses two wrench measurements per face in three-, four- and six-face
batches ($K=6,8,12$). The reference pose is obtained from CAD geometry
and manual fixture measurements; Figure~\ref{fig:physical_setup} shows
the probing video frames.
\begin{figure*}[!tp]
\centering
\setlength{\abovecaptionskip}{5pt}
\setlength{\belowcaptionskip}{0pt}
\includegraphics[width=0.9\linewidth]{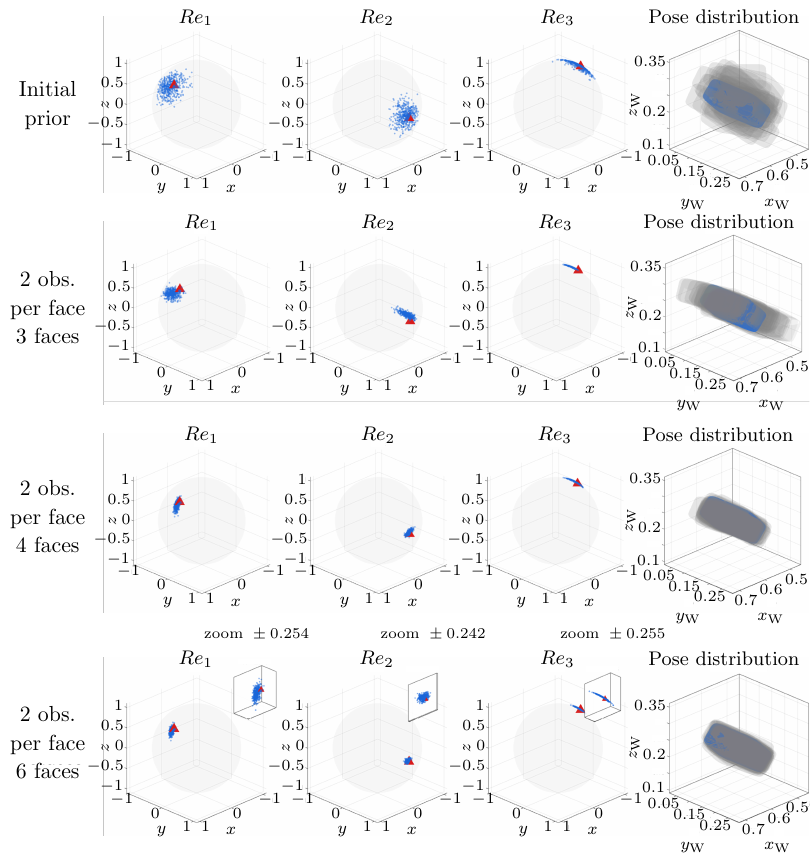}
\caption{MFG samples for the tilted box. Rows show the initial
prior and results for three ($K=6$), four ($K=8$) and six ($K=12$)
probed faces. Columns 1--3 show rotation marginals $Re_1$, $Re_2$ and
$Re_3$ on $S^2$; column 4 shows pose samples. Blue points denote samples,
red triangles the reference axes, and {gray} cuboids MFG draws. Insets enlarge the
six-face rotation marginals.}
\label{fig:original_tilted_posterior}
\end{figure*}

Table~\ref{tab:physical_pose_errors} shows decreasing rotation error
across the tested configurations. Translation error falls from $8.875$ mm
with three faces to $3.131$ mm with four, then rises to $3.465$ mm with six.
Figure~\ref{fig:original_tilted_posterior} shows broad MFG uncertainty
with three faces and more concentrated samples with four and six. The world--frame coordinates $x_W$, $y_W$ and $z_W$ in column $4$ are expressed in metres. The Cartesian coordinates in columns 1--3, including the insets, are dimensionless components of the unit vectors $R{e_1}$, $Re_2$ and $Re_3$. Insets just provide magnified views of the six-face rotation marginals.
The latter two cases illustrate that concentration need not imply
smaller translation error.

These results demonstrate refinement from real wrench measurements for
known geometry. Face coverage and measurement count increase together,
so only their combined effect is assessed.
Supplementary Section~\SMref{supp:s4:results} provides horizontal-object
results and processing details.

\begin{table}[H]
\centering
\setlength{\abovecaptionskip}{5pt}
\setlength{\belowcaptionskip}{0pt}
\caption{{Pose-estimation results for the tilted-box experiment,
with two wrench measurements per probed face.}}
\label{tab:physical_pose_errors}
\normalsize
\setlength{\tabcolsep}{3pt}
\renewcommand{\arraystretch}{1.5}
\begin{tabular*}{\linewidth}{@{\extracolsep{\fill}}lccc@{}}
\toprule
Configuration & $K$ & $e_R$ ($^\circ$) & $e_p$ (mm)\\
\midrule
Initial nominal pose & --- & $7.21$ & $8.77$\\
\addlinespace
Three faces & $6$ & $5.362$ & $8.875$\\
Four faces & $8$ & $5.011$ & $3.131$\\
Six faces & $12$ & $3.746$ & $3.465$\\
\bottomrule
\end{tabular*}
\par\vspace{3pt}
\begin{minipage}{\linewidth}
\normalsize
{\emph{Note:} $K$ is the number of wrench measurements.
Rotation and translation errors are defined in~\eqref{eq:pose_errors},
using the CAD and fixture-based reference pose.}
\end{minipage}
\end{table}

\section{Conclusion and Future Work}
\label{sec:conclusions_scope}

We developed local Bayesian pose inference from contact forces and
torques, representing uncertainty by a coupled MFG distribution on
$\mathrm{SO}(3)\times\mathbb R^3$. Implicit differentiation captures
equilibrium changes and yields a reduced Jacobian for local measurement
information and a rotational score that accounts for translation.
The Coupled MFG update matches a local quadratic model through a
conditional-mean-linearized (CL) surrogate. {Freezing is optional. For this variant, we provide} same-input error bounds and sufficient conditions
for a prescribed tolerance. Nominal-pose refinement retains the original
prior and measurements.

In the prior-mismatch comparison, the framework gave smaller mean pose
errors than the tangent-space Gaussian point update, with more refinement
passes. In the 24-dataset comparison, MFG used less computation than
RM-PF and SMC at the chosen sample counts, while RM-PF gave the smallest
mean pose errors. Additional MFG samples improved probability estimates
within that distribution without further equilibrium solves but retained
differences from the finite posterior references. Ellipsoid and smoothed
Stanford bunny simulations illustrated the closed-form reconstruction:
posterior clouds narrowed across the reported measurement counts, and
final rotation and translation errors were below their prior-stage values.
Concentration alone did not ensure accuracy. Robot experiments
demonstrated pose refinement from real wrench data under controlled
contact conditions.

The local approximation depends on the nominal pose and uses a unimodal
prior. A nonlinear contact likelihood may produce multiple posterior
modes that a single local MFG approximation may not capture. The formulation
assumes known geometry, quasi-static contact and calibrated wrench
measurements.

Future work will address joint pose and shape estimation for unknown
geometry, accounting for shape uncertainty in contact predictions and
pose information. Fisher information from this joint model could guide
contact locations and probing directions when visual measurements are
unavailable. Such policies would seek rotationally informative wrench
responses while avoiding locally redundant or infeasible contacts,
balancing rotational information, translation accuracy, contact
constraints, and robot motion limits to maintain feasible, safe
configurations.

\section*{Funding}
This research is supported by the National Research Foundation, Singapore,
under its NRF Fellowship Program, Award No. NRF-NRFF18-2026-0010.
The funder had no role in study design, data collection, analysis, or
interpretation; manuscript preparation; or the decision to submit the work
for publication.

\section*{Declaration of competing interests}
The authors declare no competing financial or personal interests relevant to
this work.

\section*{Data and code availability}
Experimental code, data-generation scripts and supporting files are
maintained in the
\href{https://github.com/CeceliaChen96/Geometric-Bayesian-Framework-with-Force-measurement-Data-on-Pose-Estimation}{project GitHub repository}.
The repository is currently private. The code and supporting data are not
publicly available because aspects of the modelling technology are subject
to intellectual-property restrictions of Nanyang Technological University. 
Access requests from other researchers
should be directed to the corresponding author and are subject to the
relevant intellectual-property restrictions. We provide the
experimental settings and results in
Supplementary Sections~\SMref{supp:sec:geometry-contact}--\SMref{supp:s4:physical-validation}
and~\SMref{app:opt-evidence}. Specifically, the geometry and contact model are in
Supplementary Sections~\SMref{supp:s1-geometry}--\SMref{supp:s1-potential}, 
measurement-number selection and nested-budget results are reported in
Supplementary Sections~\SMref{supp:s2-selection}--\SMref{supp:s2-nested-results}, 
the synthetic data and probability estimates are in
Supplementary Sections~\SMref{supp:s3-cost}--\SMref{supp:s3-sampling-diagnostics}, 
the real-world validation is in
Supplementary Sections~\SMref{supp:s4:apparatus}--\SMref{supp:s4:results},
and paired comparisons with and without freezing are in
Supplementary Sections~\SMref{app:opt-newton}--\SMref{app:opt-distribution-errors}.

Throughout the paper, we use four different groups of data. 
The first comprises synthetic {generalized-force}
observations for the prior-mismatch and tangent-space Gaussian
comparisons (Sections~\ref{subsec:original_false_confidence}
and~\ref{subsec:laplace_historical} and the geometry and contact model in
Supplementary Sections~\SMref{supp:s1-geometry}--\SMref{supp:s1-potential}). 
The second is a fixed bank of 24 synthetic datasets, each
with ten {generalized-force observations}, shared by the Coupled MFG and particle methods.
Comparison settings, probability estimates, reference checks
and sampling diagnostics are given in Supplementary
Sections~\SMref{supp:s3-cost}--\SMref{supp:s3-sampling-diagnostics}.
Both synthetic groups were generated by solving the quasi-static contact
model at known object poses and adding independent Gaussian {generalized-force}
noise. 
The third group comprises force and torque measurements recorded using
a Kinova Gen3 robot and an ATI Mini40 sensor during probing of known
3D-printed objects. Reference poses were obtained from CAD geometry and
manual fixture measurements. Supplementary
Sections~\SMref{supp:s4:apparatus}--\SMref{supp:s4:results} provide
the apparatus, probing trajectory, additional results and processing limits.
Finally, Section~\ref{subsec:shape_pose_clouds} presents two additional
synthetic pose-cloud demonstrations. Experimental details are provided
in Supplementary Section~\SMref{supp:shape-clouds}, while the 
code, saved data and videos are provided in the GitHub repository. 
The pose-cloud package includes the
bunny field and the ellipsoid construction, together
with force measurements, paired pose samples, inference 
code, figures and videos, and numerical checks for both geometries.  We have provided experimental procedures and settings in as much detail as possible, 
so that interested people can replicate these experiments.

\section*{CRediT authorship contribution statement}
\textbf{Jianyu Chen:} Conceptualization, Methodology, Formal analysis, Software,
Investigation, Validation, Visualization, Writing -- original draft,
Writing -- review \& editing.

\textbf{Lin Yang:} Methodology, Formal analysis, Validation,
Writing -- review \& editing.

\textbf{Yibang Li:} Methodology, Investigation, Validation,
Writing -- review \& editing.

\textbf{Domenico Campolo:} Conceptualization, Supervision,
Writing -- review \& editing.

\textbf{Cyrus Mostajeran:} Conceptualization, Methodology, Supervision,
Project administration, Writing -- review \& editing.

\FloatBarrier
\appendix
\renewcommand{\thetheorem}{\Alph{section}.\arabic{theorem}}
\renewcommand{\theproposition}{\thetheorem}
\renewcommand{\thecorollary}{\thetheorem}
\renewcommand{\thedefinition}{\Alph{section}.\arabic{definition}}
\renewcommand{\theremark}{\Alph{section}.\arabic{remark}}
\renewcommand{\thesubsection}{\Alph{section}.\arabic{subsection}}
\begingroup
\section{\texorpdfstring{{Proofs for Section~\ref{sec:wrench_sensitivity_information}}}{Proofs for Section \getrefnumber{sec:wrench_sensitivity_information}}}
\label{app:proofs-sec3}
\label{app:proofs}
{This appendix proves the sensitivity and information results in
Section~\ref{sec:wrench_sensitivity_information}. Throughout the appendices,
derivatives use the right-perturbation convention in
Section~\ref{subsec:pose_coordinates}.}

\subsection{\texorpdfstring{{Proof of Lemma~\ref{lem:first_order_equilibrium}}}{Proof of Lemma \getrefnumber{lem:first_order_equilibrium}}}
\label{app:equilibrium_sensitivity}
\begin{proof}
The result follows by taking the first-order Taylor expansion of
\(\Phi(\xi_A,\xi_U,\xi_B)=0\) at the nominal equilibrium and solving for
\(\xi_A\).
\end{proof}

\subsection{\texorpdfstring{{Proof of Theorem~\ref{thm:reduced_wrench_fisher_information}}}{Proof of Theorem \getrefnumber{thm:reduced_wrench_fisher_information}}}
\label{app:reduced_information}
\begin{proof}
Since $r_k=y_k+\partial_{\xi_U}W^\star$, the chain rule and
\Cref{lem:first_order_equilibrium} give
\[
\begin{aligned}
 J_k&=\partial_{\xi_B}\partial_{\xi_U}W^\star
 =W_{UB}+W_{UA}\frac{\partial\xi_A}{\partial\xi_B}\\
 &=W_{UB}-W_{UA}W_{AA}^{-1}W_{AB}.
\end{aligned}
\]
For $W=W_{\rm ctrl}(X_A,X_U)+W_{\rm int}(X_A,X_B)$, $W_{UB}=0$.
Next, $r_k(\bar X_B\oplus\xi_B)=\bar r_k+J_k\xi_B+O(\|\xi_B\|^2)$.
The  linear residual gives
\[
\begin{aligned}
 \tfrac12\|\bar r_k+J_k\xi_B\|_{\Sigma_w^{-1}}^2
 &=\tfrac12\|\bar r_k\|_{\Sigma_w^{-1}}^2
 +\xi_B^\top J_k^\top\Sigma_w^{-1}\bar r_k\\
 &\quad+\tfrac12\xi_B^\top J_k^\top\Sigma_w^{-1}J_k\xi_B.
\end{aligned}
\]
Summing proves the stated Gauss--Newton coefficients. For stacked $J$ and
$\Sigma_K=I_K\otimes\Sigma_w\succ0$, the linearized Gaussian model has
information $I=J^\top\Sigma_K^{-1}J=H_{\rm data}$. Moreover,
$
 \xi_B^\top I\xi_B
 =\sum_k\|J_k\xi_B\|_{\Sigma_w^{-1}}^2.
$
Thus $I\succ0$ exactly when $J$ has full column rank; otherwise a nonzero
null direction satisfies $J_k\xi_B=0$ for every $k$. The data have no
quadratic Gauss--Newton curvature in that direction.
\end{proof}

\subsection{\texorpdfstring{{Proof of Theorem~\ref{thm:rotational_information_score}}}{Proof of Theorem \getrefnumber{thm:rotational_information_score}}}
\label{app:rotational_information}
\begin{proof}
For fixed $\phi$, stationarity of the quadratic form with respect to $v$
gives $I_{vv}v=-I_{v\phi}\phi$, with a unique solution because
$I_{vv}\succ0$. Substitution yields the Schur form
$\phi^\top I_{\rm rot}\phi$, which is nonnegative because $I\succeq0$.
Rayleigh--Ritz{~\cite[Sec.~4.2]{horn2013matrix}} gives the stated
unit-direction minimum, and homogeneity gives the bound for arbitrary $\phi$.
\end{proof}

\section{\texorpdfstring{{Proofs for Section~\ref{sec:local_bayesian_closure}}}{Proofs for Section \getrefnumber{sec:local_bayesian_closure}}}
\label{app:proofs-sec4}
\subsection{\texorpdfstring{{Proof of Proposition~\ref{prop:local_quadratic_posterior}}}{Proof of Proposition \getrefnumber{prop:local_quadratic_posterior}}}
\label{app:local_posterior}

\begin{proof}
Let $\xi=\xi_B$ and
$e_0(\xi)=E_0(\bar X_B\oplus\xi)$.
By~\Cref{ass:prior_regularity}, $e_0$ is $C^3$ near
$\xi=0$. Taylor's theorem therefore gives
\begin{equation*}
\begin{aligned}
e_0(\xi)
={}&e_0(0)+g_{\mathrm{prior}}^\top\xi\\
&+\frac12\xi^\top H_{\mathrm{prior}}^{\mathrm{ex}}\xi
+R_3(\xi),
\end{aligned}
\end{equation*}
where $|R_3(\xi)|\le C\|\xi\|^3$ for
sufficiently small $\xi$.
Subtracting the prior model
\eqref{eq:local_prior_model} gives
\begin{equation*}
\begin{aligned}
&e_0(\xi)-\widehat E_0(\bar X_B\oplus\xi)\\
&\quad=
\frac12\xi^\top
\left(H_{\mathrm{prior}}^{\mathrm{ex}}
-H_{\mathrm{prior}}\right)\xi
+R_3(\xi),
\end{aligned}
\end{equation*}
which establishes
\eqref{eq:prior_model_discrepancy}. Next, by \Cref{thm:reduced_wrench_fisher_information}\textup{(ii)}, the \mbox{Gauss--Newton}
data model is
\begin{equation*}
\begin{aligned}
\widehat J_{\mathrm{data}}
(\bar X_B\oplus\xi)
&=\frac12\sum_{k=1}^K
\|\bar r_k+J_k\xi\|_{\Sigma_w^{-1}}^2\\
&=J_{\mathrm{data}}(\bar X_B)
+g_{\mathrm{data}}^\top\xi
+\frac12\xi^\top H_{\mathrm{data}}\xi.
\end{aligned}
\end{equation*}
Adding the prior model gives
\begin{equation*}
\begin{aligned}
\widehat J_{\mathrm{MAP}}
(\bar X_B\oplus\xi)
={}&J_{\mathrm{MAP}}(\bar X_B)\\
&+(g_{\mathrm{data}}+g_{\mathrm{prior}})^\top\xi\\
&+\frac12\xi^\top
(H_{\mathrm{data}}+H_{\mathrm{prior}})\xi.
\end{aligned}
\end{equation*}
Hence,
$
g_{\mathrm{post}}
=g_{\mathrm{data}}+g_{\mathrm{prior}},
\quad
H_{\mathrm{post}}
=H_{\mathrm{data}}+H_{\mathrm{prior}},
$
as claimed. Finally, suppose $H_{\mathrm{post}}\succ0$.
Completing the square yields
\begin{equation*}
\begin{aligned}
\widehat J_{\mathrm{MAP}}
(\bar X_B\oplus\xi)
={}&C_0+
\frac12(\xi-\xi_{\mathrm{quad}}^\star)^\top
H_{\mathrm{post}}
(\xi-\xi_{\mathrm{quad}}^\star),
\end{aligned}
\end{equation*}
where
$
\xi_{\mathrm{quad}}^\star
=-H_{\mathrm{post}}^{-1}g_{\mathrm{post}},
C_0=J_{\mathrm{MAP}}(\bar X_B)
-\frac12g_{\mathrm{post}}^\top
H_{\mathrm{post}}^{-1}g_{\mathrm{post}}.
$
Positive definiteness guarantees that
$\xi_{\mathrm{quad}}^\star$ is the unique
minimizer of the quadratic model. Moreover, normalizing
$\exp[-\widehat J_{\mathrm{MAP}}]$
with respect to Lebesgue measure in the
Euclidean local coordinates gives
$
q_{\mathrm{quad}}
=\mathcal N\left(
\xi_{\mathrm{quad}}^\star,
H_{\mathrm{post}}^{-1}
\right).
$
This characterizes the Gaussian information
model associated with the prescribed
quadratic energy, without identifying it
with the exact nonlinear posterior.
\end{proof}

\subsection{\texorpdfstring{{Proof of Proposition~\ref{prop:prior_information}}}{Proof of Proposition \getrefnumber{prop:prior_information}}}
\label{app:prior_derivation}
\begin{proof}
\begingroup
\setlength{\abovedisplayskip}{5pt plus 1pt minus 1pt}
\setlength{\belowdisplayskip}{5pt plus 1pt minus 1pt}
\setlength{\abovedisplayshortskip}{3pt plus 1pt}
\setlength{\belowdisplayshortskip}{4pt plus 1pt minus 1pt}
\setlength{\jot}{2pt}
Write the fixed prior parameters as $(F,\mu,\Gamma,\Lambda)$, with
proper SVD $F=USV^\top$. The prior energy is
$-\tr(F^\top R_B)+\tfrac12\|p_B-\mu-\Gamma\nu_{R_B}\|_\Lambda^2$.

\paragraph{(i) Jacobian of $\nu_{R_B}$} \label{part1}
Use $J_\nu$ from~\eqref{eq:prior_statistic_jacobian}.
With $R_B(\phi)=\bar R_B\exp(\phi^\wedge)$, put
$Q(R_B)=U^\top R_BV$ and $\bar Q=Q(\bar R_B)$.
Since $V^\top\phi^\wedge V=(V^\top\phi)^\wedge$, expansion of the
exponential gives
$Q(R_B(\phi))=\bar Q+\bar Q(V^\top\phi)^\wedge+O(\|\phi\|^2)$.
Therefore,
\[
 \nu_{R_B(\phi)}-\bar\nu
 =\bigl(\bar Q(V^\top\phi)^\wedge S
       +S(V^\top\phi)^\wedge\bar Q^\top\bigr)^\vee
       +O(\|\phi\|^2),
\]
where $\bar\nu=(\bar QS-S\bar Q^\top)^\vee$.
The linear term is $J_\nu\phi$.

\paragraph{(ii) Conditional Gaussian term}
Put $e(\phi,v)=\bar p_B+v-\mu-\Gamma\nu_{R_B(\phi)}$,
$\bar e=e(0,0)$ and $E_G(\xi)=\tfrac12e(\phi,v)^\top\Lambda e(\phi,v)$.
{Part~\textup{(i)}} gives
\begin{equation}
 e(\phi,v)=\bar e+v-\Gamma J_\nu\phi+O(\|\phi\|^2).
\label{eq:B11_identity_new}
\end{equation}
Thus $J_e=\partial_\xi e(0)=[-\Gamma J_\nu\;\;I_3]$.
The chain rule~\cite{magnus2019matrix} gives the exact gradient
$\nabla_\xi E_G(0)=J_e^\top\Lambda\bar e$; using the linear residual yields
\begin{equation}
 \nabla_\xi^2\!\left(\tfrac12\|\bar e+J_e\xi\|_\Lambda^2\right)
 =J_e^\top\Lambda J_e.
\label{eq:B15_new}
\end{equation}
These are the conditional Gaussian contributions to the gradient and
Gauss--Newton curvature in~\Cref{prop:prior_information}.
The exact Hessian is
\[
 \nabla_\xi^2 E_G(0)=J_e^\top\Lambda J_e
   +\sum_{a=1}^3(\Lambda\bar e)_a\nabla_\xi^2 e_a(0).
\]
Here $\partial_v e=I_3$ and
$\partial_{\phi_i\phi_j}e(0)=-\Gamma\left.\partial_{\phi_i\phi_j}
\nu_{R_B(\phi)}\right|_{\phi=0}$.
Thus the correction is $-\diag(C_\nu,0_{3\times3})$, with $C_\nu$ defined
below~\eqref{eq:prior-curvature-correction}. Adding {the exact curvature of the negative log density of the matrix Fisher distribution} therefore gives that identity.

\paragraph{(iii) {Contribution from the matrix Fisher distribution}}
For $A=F^\top\bar R_B$, the identities
$\tr(A\phi^\wedge)=-(A-A^\top)^{\vee\top}\phi$ and
$(\phi^\wedge)^2=\phi\phi^\top-\|\phi\|^2I_3$ give
\[
\begin{aligned}
 &-\tr(A\exp(\phi^\wedge))+\tr(A)\\
 &\,=(A-A^\top)^{\vee\top}\phi
 +\tfrac12\phi^\top[\tr(A)I_3-\operatorname{sym}(A)]\phi
   +O(\|\phi\|^3).
\end{aligned}
\]
Thus {the exact gradient and curvature of the negative log density of the matrix Fisher distribution} are
$(A-A^\top)^\vee=-(\bar R_B^\top F-F^\top\bar R_B)^\vee$ and
$\tr(A)I_3-\operatorname{sym}(A)$, respectively.
Adding these to the conditional Gaussian contributions proves
\Cref{prop:prior_information}.
\endgroup
\end{proof}

\subsection{\texorpdfstring{{Proof of Theorem~\ref{thm:closed_form_reconstruction}}}{Proof of Theorem \getrefnumber{thm:closed_form_reconstruction}}}
\label{app:closed_form_update}

\begingroup
\begin{proof}
\begingroup
\setlength{\abovedisplayskip}{5pt plus 1pt minus 1pt}
\setlength{\belowdisplayskip}{5pt plus 1pt minus 1pt}
\setlength{\abovedisplayshortskip}{3pt plus 1pt}
\setlength{\belowdisplayshortskip}{4pt plus 1pt minus 1pt}
\setlength{\jot}{2pt}
We use the quadratic form $q$ in~\eqref{eq:cmfg_quadratic_input}
with information pair $(g,H)$ at
$\bar X_B$, with $\xi=(\phi^\top,v^\top)^\top$ and
$H_{\phi v}=H_{v\phi}^\top$.

\paragraph{(i) Eliminate the translational increment}
For each fixed $\phi$, the terms in $q$ that depend on $v$ are
$(g_v+H_{v\phi}\phi)^\top v+\tfrac12v^\top H_{vv}v$.
Since $H_{vv}\succ0$, their unique minimizer is
$v^\star(\phi)=-H_{vv}^{-1}(g_v+H_{v\phi}\phi)$.
Substituting this expression into $q$ gives
\[
\begin{aligned}
 q(\phi,v^\star(\phi))
 &=g_\phi^\top\phi+\tfrac12\phi^\top H_{\phi\phi}\phi\\
 &\quad-\tfrac12(g_v+H_{v\phi}\phi)^\top H_{vv}^{-1}
                  (g_v+H_{v\phi}\phi)\\
 &=-\tfrac12g_v^\top H_{vv}^{-1}g_v
   +a_{\mathrm{post}}^\top\phi
   +\tfrac12\phi^\top M_{\mathrm{post}}\phi.
\end{aligned}
\]
Thus $a_{\mathrm{post}}$ and $M_{\mathrm{post}}$ are the rotational
gradient and curvature after minimizing the quadratic model over $v$.
Completing the square, with $\Lambda_{\mathrm{post}}=H_{vv}$, gives
the identity
\begin{equation}
\begin{aligned}
q(\xi)
={}&a_{\mathrm{post}}^\top\phi
+\tfrac12\phi^\top M_{\mathrm{post}}\phi\\
&+\tfrac12\bigl\|v+\Lambda_{\mathrm{post}}^{-1}
                    (g_v+H_{v\phi}\phi)\bigr\|_{\Lambda_{\mathrm{post}}}^2
-\tfrac12g_v^\top\Lambda_{\mathrm{post}}^{-1}g_v.
\end{aligned}
\label{eq:app_thm43_profiled_final}
\end{equation}
No definiteness assumption on $M_{\mathrm{post}}$ is needed for this
decomposition.

\paragraph{(ii) Reconstruct the matrix Fisher parameter}
Let $A=\bar R_B^\top F_{\mathrm{post}}$ and
$f(\phi)=-\operatorname{tr}(F_{\mathrm{post}}^\top
\bar R_B\exp(\widehat\phi))$.
Using $\exp(\widehat\phi)=I_3+\widehat\phi+
\tfrac12\widehat\phi^2+O(\|\phi\|^3)$ and
$\widehat\phi^2=\phi\phi^\top-\|\phi\|^2I_3$ gives the rotational gradient
$-(A-A^\top)^\vee$ and curvature
$\operatorname{tr}(A)I_3-\operatorname{sym}(A)$.
Matching these with the reduced coefficients in part~(i) requires
\[
 -(A-A^\top)^\vee=a_{\mathrm{post}},
 \qquad
 \operatorname{tr}(A)I_3-\operatorname{sym}(A)=M_{\mathrm{post}}.
\]
Taking the trace of the second equation gives
$2\operatorname{tr}(A)=\operatorname{tr}(M_{\mathrm{post}})$,
so $\operatorname{sym}(A)=C_{\mathrm{post}}$.
The first equation gives
$\operatorname{skew}(A)=-\tfrac12\widehat{a_{\mathrm{post}}}$.
Together these determine
$A=C_{\mathrm{post}}-\tfrac12\widehat{a_{\mathrm{post}}}$ and hence
the stated $F_{\mathrm{post}}=\bar R_BA$.
In particular,
\begin{equation}
f(\phi)-f(0)
=a_{\mathrm{post}}^\top\phi
+\tfrac12\phi^\top M_{\mathrm{post}}\phi+O(\|\phi\|^3).
\label{eq:cl_fisher_expansion}
\end{equation}

\paragraph{(iii) Reconstruct the conditional Gaussian parameters}
Evaluate $\bar\nu_{\mathrm{post}}$ and $J_{\nu,\mathrm{post}}$
from this $F_{\mathrm{post}}$ using the prescribed proper SVD selection.
To obtain the matrix expression in~\eqref{eq:cmfg_statistic_update},
abbreviate $Q=\bar Q_{\mathrm{post}}$, $S=S_{\mathrm{post}}$ and
$V=V_{\mathrm{post}}$.
For $z=V^\top\phi$, differentiation of the rotational statistic gives
\[
J_{\nu,\mathrm{post}}\phi
=(Q\widehat zS+S\widehat zQ^\top)^\vee
=Q\bigl[\operatorname{tr}(SQ)I_3-SQ\bigr]z.
\]
The last equality follows by factoring the skew matrix as
$Q[\widehat zSQ+Q^\top S\widehat z]Q^\top$
and applying
$(\widehat z A+A^\top\widehat z)^\vee
=[\operatorname{tr}(A)I_3-A]z$ with $A=SQ$.
This proves~\eqref{eq:cmfg_statistic_update}.
For the CL residual in~\eqref{eq:cl_mfg_surrogate}, matching the
translation curvature, cross curvature and translation gradient requires
\[
\begin{aligned}
 \Lambda_{\mathrm{post}}&=H_{vv},\\
 -\Lambda_{\mathrm{post}}\Gamma_{\mathrm{post}}
       J_{\nu,\mathrm{post}}&=H_{v\phi},\\
 \Lambda_{\mathrm{post}}
 (\bar p_B-\mu_{\mathrm{post}}
          -\Gamma_{\mathrm{post}}\bar\nu_{\mathrm{post}})&=g_v.
\end{aligned}
\]
Since $H_{vv}$ and $J_{\nu,\mathrm{post}}$ are nonsingular,
the second equation gives
$\Gamma_{\mathrm{post}}=-H_{vv}^{-1}H_{v\phi}
J_{\nu,\mathrm{post}}^{-1}$; the third then gives
$\mu_{\mathrm{post}}=\bar p_B-H_{vv}^{-1}g_v
-\Gamma_{\mathrm{post}}\bar\nu_{\mathrm{post}}$.
Consequently,
\[
 e_{\mathrm{CL}}(\xi;\Theta_{\mathrm{post}})
 =v+H_{vv}^{-1}(g_v+H_{v\phi}\phi).
\]
Its Gaussian energy is exactly the square term
in~\eqref{eq:app_thm43_profiled_final}.

\paragraph{(iv) Verify the full local matching}
Write $E_G(\xi)=\tfrac12\|e_{\mathrm{CL}}
(\xi;\Theta_{\mathrm{post}})\|_{\Lambda_{\mathrm{post}}}^2$.
Part~(iii) matches its translation and cross blocks, while its rotational
contributions at zero are
\[
\begin{aligned}
 \nabla_\phi E_G(0)
 &=-J_{\nu,\mathrm{post}}^\top\Gamma_{\mathrm{post}}^\top g_v
   =H_{\phi v}H_{vv}^{-1}g_v,\\
 \nabla_\phi^2E_G(0)
 &=J_{\nu,\mathrm{post}}^\top\Gamma_{\mathrm{post}}^\top
   H_{vv}\Gamma_{\mathrm{post}}J_{\nu,\mathrm{post}}
   =H_{\phi v}H_{vv}^{-1}H_{v\phi}.
\end{aligned}
\]
Adding the matrix Fisher contributions from part~(ii) therefore gives
\[
\begin{aligned}
 \nabla_\phi\widehat E_{\mathrm{CL}}(0;\Theta_{\mathrm{post}})
 &=a_{\mathrm{post}}+H_{\phi v}H_{vv}^{-1}g_v=g_\phi,\\
 \nabla_\phi^2\widehat E_{\mathrm{CL}}(0;\Theta_{\mathrm{post}})
 &=M_{\mathrm{post}}+H_{\phi v}H_{vv}^{-1}H_{v\phi}
   =H_{\phi\phi}.
\end{aligned}
\]
All gradient and curvature blocks are thus matched.
Equations~\eqref{eq:app_thm43_profiled_final}
and~\eqref{eq:cl_fisher_expansion} also give
\begin{equation*}
\widehat E_{\mathrm{CL}}(\xi;\Theta_{\mathrm{post}})
-\widehat E_{\mathrm{CL}}(0;\Theta_{\mathrm{post}})
=q(\xi)+O(\|\xi\|^3),\qquad \xi\to0.
\end{equation*}
The conditional Gaussian term is quadratic and the matrix Fisher term
is smooth, so~\eqref{eq:local_matching} follows.
The parameters define a coupled MFG model because
$\Lambda_{\mathrm{post}}\succ0$; the local matching just proved is for
its CL surrogate.

\paragraph{(v) Uniqueness under the proper SVD selection convention}
Let $\widetilde\Theta$ be any other matching tuple, and set
$\widetilde e_0=\bar p_B-\widetilde\mu
-\widetilde\Gamma\widetilde{\bar\nu}$.
Its translation and cross blocks impose
\[
 \widetilde\Lambda=H_{vv},\qquad
 \widetilde e_0=H_{vv}^{-1}g_v,\qquad
 \widetilde\Gamma\widetilde J_\nu=-H_{vv}^{-1}H_{v\phi}.
\]
These identities determine the conditional Gaussian rotational
contributions exactly as in part~(iv).
The remaining matrix Fisher gradient and curvature must therefore be
$a_{\mathrm{post}}$ and $M_{\mathrm{post}}$.
The symmetric and skew-symmetric equations in part~(ii) then give
$\bar R_B^\top\widetilde F
=C_{\mathrm{post}}-\tfrac12\widehat{a_{\mathrm{post}}}$,
so $\widetilde F=F_{\mathrm{post}}$.
Using the same proper SVD selection convention now gives
$\widetilde{\bar\nu}=\bar\nu_{\mathrm{post}}$ and
$\widetilde J_\nu=J_{\nu,\mathrm{post}}$.
Its nonsingularity determines
$\widetilde\Gamma=\Gamma_{\mathrm{post}}$ from the cross block and
$\widetilde\mu=\mu_{\mathrm{post}}$ from the translation gradient.
This proves uniqueness.
\endgroup
\end{proof}
\endgroup

\begingroup
\subsection{Proof of Remark~\ref{rem:actual-mfg-curvature}}
\label{app:actual-mfg-curvature}
\begin{proof}
Write
$\Lambda=\Lambda_{\mathrm{post}}$,
$\Gamma=\Gamma_{\mathrm{post}}$,
$\bar\nu=\bar\nu_{\mathrm{post}}$, and
$J_\nu=J_{\nu,\mathrm{post}}$.
The parameter identities in Theorem~\ref{thm:closed_form_reconstruction}
give the CL translation residual
$
e_{\mathrm{CL}}(\xi)
=
\Lambda^{-1}g_v+v-\Gamma J_\nu\phi.
$
Set
$
r_\nu(\phi)
=
\nu_R(\bar R_B\exp(\widehat\phi);F_{\mathrm{post}})
-\bar\nu-J_\nu\phi,
$
and
$
h(\phi)=\Gamma r_\nu(\phi).
$
The actual translation residual is $e_{\mathrm{CL}}-h$, so
\[
\widetilde{\mathcal V}_u(\xi)
-
\widehat E_{\mathrm{CL}}(\xi;\Theta_{\mathrm{post}})
=
-e_{\mathrm{CL}}(\xi)^\top\Lambda h(\phi)
+\frac12h(\phi)^\top\Lambda h(\phi).
\]
By construction, $h(0)=0$ and $D_\phi h(0)=0$.
Consequently, this difference has zero gradient at $\xi=0$,
and its only possibly nonzero Hessian block is rotational.
For $i,\ell\in\{1,2,3\}$, that block equals
\[
\begin{aligned}
&-\bigl(\Lambda e_{\mathrm{CL}}(0)\bigr)^\top
  \partial_{\phi_i\phi_\ell}^2 h(0)\\
&\quad=
-g_v^\top\Gamma
  \partial_{\phi_i\phi_\ell}^2
  \nu_R(\bar R_B\exp(\widehat\phi);F_{\mathrm{post}})
  \big|_{\phi=0}.
\end{aligned}
\]
This is precisely $(C_u)_{i\ell}$.
The claimed identities follow from the gradient and Hessian
matching of the CL surrogate.
For the coordinate-Lebesgue density, the additional energy is
$
-\log j(\phi)
=\log(8\pi^2)+\|\phi\|^2 /12 
+O(\|\phi\|^4).
$
Its gradient vanishes at zero and its rotational Hessian is $I_3/6$.
Thus the same MFG law has coordinate-Lebesgue energy curvature
\[
\nabla_\xi^2
\bigl[\widetilde{\mathcal V}_u(\xi)-\log j(\phi)\bigr]_{\xi=0}
=H+\operatorname{diag}(C_u+I_3/6,0_{3\times3}).
\]
This completes the proof of Remark~\ref{rem:actual-mfg-curvature}.
\end{proof}

\subsection{Proof of Corollary~\ref{cor:statistic-jacobian-condition}}
\label{app:statistic-jacobian-condition}
\begin{proof}
Write 
$
A=\bar R_B^\top F_{\mathrm{post}}
=C_{\mathrm{post}}-\frac12\widehat a_{\mathrm{post}},
$
and abbreviate the proper SVD factors by $U, S, V$.
Since $\bar Q=U^\top\bar R_BV$ and $F_{\mathrm{post}}=USV^\top$,
we have $
S\bar Q=V^\top A^\top V,$ and hence
$
\operatorname{tr}(S\bar Q)=\operatorname{tr}(A).
$
Substituting these identities into the expression for
$J_{\nu,\mathrm{post}}$ yields
\begin{equation}
\label{eq:statistic-jacobian-identity}
\begin{aligned}
J_{\nu,\mathrm{post}}
&=
\bar Q\bigl[
 \operatorname{tr}(S\bar Q)I_3-S\bar Q
\bigr]V^\top\\
&=
U^\top\bar R_B
\bigl[
 \operatorname{tr}(A)I_3-A^\top
\bigr]\\
&=
U^\top\bar R_B
\bigl(M_{\mathrm{post}}-\tfrac12\widehat a_{\mathrm{post}}\bigr).
\end{aligned}
\end{equation}
The last equality follows from
$
C_{\mathrm{post}}
=
\frac12\operatorname{tr}(M_{\mathrm{post}})I_3
-M_{\mathrm{post}},
$
which gives
$
\operatorname{tr}(A)I_3-A^\top
=
M_{\mathrm{post}}-\frac12\widehat a_{\mathrm{post}}.
$
This proves the Jacobian identity~\eqref{eq:statistic-jacobian-identity}. Set
$B=M_{\mathrm{post}}-\frac12\widehat a_{\mathrm{post}}$.
The matrix $U^\top\bar R_B$ is orthogonal, so
$J_{\nu,\mathrm{post}}$ and $B$ have the same singular values.
For every $x\ne0$, skew-symmetry and the Cauchy--Schwarz
inequality~\cite[Sec.~0.6]{horn2013matrix} imply 
\[
\|Bx\|_2\|x\|_2
\ge
x^\top Bx
=
x^\top M_{\mathrm{post}}x
\ge
\lambda_{\min}(M_{\mathrm{post}})\|x\|_2^2.
\]
Taking the infimum over unit vectors proves the singular-value
bound and hence the inverse-norm bound.
Finally, if $H\succ0$, its Schur complement
$
M_{\mathrm{post}}
=
H_{\phi\phi}-H_{\phi v}H_{vv}^{-1}H_{v\phi}
$
is positive definite.
\end{proof}
\endgroup

\endgroup

\section{\texorpdfstring{{Proofs and auxiliary procedures for Section~\ref{sec:safeguarded_refinement}}}{Proofs and auxiliary procedures for Section \getrefnumber{sec:safeguarded_refinement}}}
\label{app:proofs-sec5}
\begingroup
\subsection{Optional freezing: matching and error bounds}
\label{app:frozen}
\label{app:opt-freezing}
The fixed-statistic option is {an approximation to the Coupled MFG update}.
This appendix defines the two outputs at a common local input, characterizes
their CL matching and bounds their distributional difference.
\endgroup

\subsubsection{\texorpdfstring{{Statement and proof of Corollary~\ref{cor:frozen_statistic_reconstruction}}}{Statement and proof of Corollary \getrefnumber{cor:frozen_statistic_reconstruction}}}
\label{app:proof-frozen-statistic}

\begingroup
\begin{corollary}[Fixed-statistic implementation]
\label{cor:frozen_statistic_reconstruction}
Use the prior statistics defined in~\Cref{subsec:prior_information_form}:
$$
\begin{aligned}
\bar\nu&=\nu_R(\bar R_B;F_0)\in\mathbb R^3,\\
J_\nu&=\left.
\frac{\partial\nu_R(\bar R_B\exp(\widehat\phi);F_0)}
{\partial\phi}\right|_{\phi=0}\in\mathbb R^{3\times3}.
\end{aligned}
$$
Keep using the same $F_{\mathrm{post}}$ and $\Lambda_{\mathrm{post}}=H_{vv}\succ0$
from~\Cref{thm:closed_form_reconstruction}, and assume $J_\nu$
is nonsingular. With $(\bar\nu,J_\nu)$ fixed, define
\begin{equation}
    \label{eq:frozen_statistic_update}
\begin{aligned}
\Gamma_{\mathrm{post}}^{\mathrm{frz}}
&=-H_{vv}^{-1}H_{v\phi}J_\nu^{-1},\\
\mu_{\mathrm{post}}^{\mathrm{frz}}
&=\bar p_B-H_{vv}^{-1}g_v
-\Gamma_{\mathrm{post}}^{\mathrm{frz}}\bar\nu,
\end{aligned}
\end{equation}
and form
$\Theta_{\mathrm{post}}^{\mathrm{frz}}
=(F_{\mathrm{post}},\mu_{\mathrm{post}}^{\mathrm{frz}},
\Lambda_{\mathrm{post}},\Gamma_{\mathrm{post}}^{\mathrm{frz}})$.
With $\Theta=\Theta_{\mathrm{post}}^{\mathrm{frz}}$, replace
$(\bar\nu_F,J_{\nu,F})$ by $(\bar\nu,J_\nu)$
in~\eqref{eq:cl_mfg_surrogate}.
The resulting energy has gradient $g$ and Hessian $H$ at $\xi=0$.
The returned MFG CL surrogate satisfies~\eqref{eq:local_matching}
if and only if
$
\Gamma_{\mathrm{post}}^{\mathrm{frz}}
(\bar\nu_{\mathrm{post}}-\bar\nu)=0,$ and $
\Gamma_{\mathrm{post}}^{\mathrm{frz}}
(J_{\nu,\mathrm{post}}-J_\nu)=0,
$
where $(\bar\nu_{\mathrm{post}},J_{\nu,\mathrm{post}})$ are evaluated
using $F_{\mathrm{post}}$ as in~\Cref{thm:closed_form_reconstruction}.
\end{corollary}
\endgroup

\begin{proof}
\begingroup
\setlength{\abovedisplayskip}{5pt plus 1pt minus 1pt}
\setlength{\belowdisplayskip}{5pt plus 1pt minus 1pt}
\setlength{\abovedisplayshortskip}{3pt plus 1pt}
\setlength{\belowdisplayshortskip}{4pt plus 1pt minus 1pt}
\setlength{\jot}{2pt}
Write $\Lambda=H_{vv}\succ0$. Substituting
\eqref{eq:frozen_statistic_update} into the fixed-statistic surrogate gives
the conditional translation residual
\[
\bar p_B+v-\mu_{\mathrm{post}}^{\mathrm{frz}}
-\Gamma_{\mathrm{post}}^{\mathrm{frz}}(\bar\nu+J_\nu\phi)
=v+\Lambda^{-1}(g_v+H_{v\phi}\phi).
\]
Together with {the unchanged negative log density of the matrix Fisher distribution},
\eqref{eq:app_thm43_profiled_final} and~\eqref{eq:cl_fisher_expansion}
therefore establish its local matching. The returned MFG uses the statistics induced by $F_{\mathrm{post}}$.
Its CL residual is
\[
\begin{aligned}
e_{\mathrm{CL}}(\xi;\Theta_{\mathrm{post}}^{\mathrm{frz}})
={}&v+\Lambda^{-1}(g_v+H_{v\phi}\phi)\\
&-\Gamma_{\mathrm{post}}^{\mathrm{frz}}
\bigl[(\bar\nu_{\mathrm{post}}-\bar\nu)
+(J_{\nu,\mathrm{post}}-J_\nu)\phi\bigr].
\end{aligned}
\]
Its translation gradient and cross block at zero therefore satisfy
\[
\begin{aligned}
\nabla_v\widehat E_{\mathrm{CL}}(0)-g_v
&=-\Lambda\Gamma_{\mathrm{post}}^{\mathrm{frz}}
(\bar\nu_{\mathrm{post}}-\bar\nu),\\
\nabla_{v\phi}^2\widehat E_{\mathrm{CL}}(0)-H_{v\phi}
&=-\Lambda\Gamma_{\mathrm{post}}^{\mathrm{frz}}
(J_{\nu,\mathrm{post}}-J_\nu).
\end{aligned}
\]
Since $\Lambda$ is nonsingular, full local matching requires both
conditions in the corollary. Conversely, these conditions make the two
residuals identical for every $\xi$. {Their negative log densities for the matrix Fisher distribution also coincide}, so the returned MFG CL surrogate has the established local
gradient $g$ and Hessian $H$.
\endgroup
\end{proof}

\begingroup
\subsubsection{\texorpdfstring{{Sufficient conditions for optional freezing}}{Sufficient conditions for optional freezing}}
\label{app:proof-freezing-certificate}
Fix the same nominal pose and local input $(g,H)$ for
Theorem~\ref{thm:closed_form_reconstruction} and
Corollary~\ref{cor:frozen_statistic_reconstruction}, under their hypotheses:
$H_{vv}\succ0$ and both statistic Jacobians are nonsingular.
Use the same proper SVD convention and write
$F=F_{\mathrm{post}}$, $\Lambda=H_{vv}$ and $\nu(R)=\nu_R(R;F)$.
Set $(\mu_u,\Gamma_u)=(\mu_{\mathrm{post}},\Gamma_{\mathrm{post}})$ and
$(\mu_f,\Gamma_f)=(\mu_{\mathrm{post}}^{\mathrm{frz}},
\Gamma_{\mathrm{post}}^{\mathrm{frz}})$
{for the Coupled MFG and frozen outputs, respectively.}
{Using the matrix Fisher notation of
Section~\ref{subsec:coupled_mfg_prior}, write
$\mathcal{MF}(F)(dR)=p(R;F)\,dR$, where $dR$ is normalized
Haar measure on the compact Lie group $\mathrm{SO}(3)$. This measure is
invariant under left and right group translations~\cite[Sec.~II-A]{lee2018bayesian}.}
The two returned laws are
\begin{equation}
 \begin{aligned}
 Q_b(dR\,dp)&={\mathcal{MF}(F)(dR)}\\
 &\quad\times\mathcal N(dp;\mu_b+\Gamma_b\nu(R),\Lambda^{-1}),
 \quad b\in\{u,f\}.
 \end{aligned}
 \label{eq:opt-two-laws}
\end{equation}
Both densities use the posterior statistic $\nu(R)$.
Set $\delta\mu=\mu_f-\mu_u$, $\delta\Gamma=\Gamma_f-\Gamma_u$ and
$
 \Delta m(R):=\delta\mu+\delta\Gamma\nu(R),
$
which is the difference of their conditional translation means. 
If $H_{v\phi}=0$, both couplings vanish and the two laws coincide.
\begingroup
All vector norms below are Euclidean, and $\delta\Gamma_{:i}$ denotes
column $i$ of $\delta\Gamma$. In the fixed proper SVD $F=USV^T$,
$U,V\in\mathrm{SO}(3)$ and $S=\operatorname{diag}(s_1,s_2,s_3)$ with
$s_1\ge s_2\ge|s_3|$.
Put $Q_R=U^TRV$. For $R\sim\mathcal{MF}(F)$, define the matrix mean
and statistic covariance by
$
 D=\mathbb E Q_R, d_i=\mathbb E[(Q_R)_{ii}],
 \Sigma_\nu=\operatorname{Cov}(\nu(R)).
$
Choose a matrix factor $W\in\mathbb R^{3\times3}$ with $W^TW=\Lambda$
and set $\bar c_i=\sum_{j\ne i}|s_j|$.
\endgroup
\par\noindent Define the following two bounds:
$
 B_K=\tfrac12\|W\delta\mu\|^2
       +\tfrac12\sum_i\bar c_i\|W\delta\Gamma_{:i}\|^2,$ and
$B_p=\left(\|\delta\mu\|^2
       +\sum_i\bar c_i\|\delta\Gamma_{:i}\|^2\right)^{1/2}.
$
\begingroup
We use the Kullback--Leibler (KL) divergence with natural
logarithms~\cite[Sec.~8.5]{cover2006elements} and the total variation (TV) distance in
Definition~\ref{def:sec63-posterior-errors}.
\endgroup

Let $\Phi$ denote the
standard normal cumulative distribution function and $\operatorname{erf}$
the Gaussian error function~\cite[Secs.~7.2(i) and~7.20(iii)]{temmeDLMFErrorFunctions}.
For $x\in\mathbb R$, they satisfy $\operatorname{erf}(x)=2\Phi(\sqrt2x)-1$.

\begin{proposition}[Sufficient error budgets for optional freezing]
\label{prop:opt-freezing-certificate}
\leavevmode\par\nobreak\noindent
For the same-input laws in \eqref{eq:opt-two-laws}, let
$K_f=\operatorname{KL}(Q_f\Vert Q_u)$, we have 
\begin{align}
 K_f&=\operatorname{KL}(Q_u\Vert Q_f)\notag\\
 &=\tfrac12\{\delta\mu^T\Lambda\delta\mu+
          \operatorname{tr}(\Lambda\delta\Gamma{\Sigma_\nu}\delta\Gamma^T)\}
 \le B_K.\label{eq:opt-freezing-kl}
\end{align}
Then let $e_{\rm cm}$ denote the
root-mean-square (RMS) difference between their conditional translation means. That is
\begin{align}
 {e_{\rm cm}^2}&:=\mathbb E_{{\mathcal{MF}(F)}}\|\Delta m(R)\|^2
       \notag\\
       &=\|\delta\mu\|^2+
          \operatorname{tr}(\delta\Gamma{\Sigma_\nu}\delta\Gamma^T)
       \le B_p^2,\label{eq:opt-freezing-rms}\\
 d_{\rm TV}(Q_f,Q_u)
 &=\mathbb E_{{\mathcal{MF}(F)}}\left[2\Phi(\|W\Delta m(R)\|/2)-1\right]
 \notag\\
 &\le\operatorname{erf}(\sqrt{B_K}/2).\label{eq:opt-freezing-tv}
\end{align}
With a choice of rotation mode, the joint-mode translation
difference is exactly $\|\delta\mu\|$.
{Let $(Q_b)_p$ denote the translation marginal of $Q_b$.
For the Euclidean 2-Wasserstein distance $W_2$ on $\mathbb R^3$
\cite[Definition~6.1]{villani2009optimal}, we have the bounds
\begin{equation}
 W_2((Q_f)_p,(Q_u)_p)\le e_{\rm cm}\le B_p.
 \label{eq:opt-freezing-w2}
\end{equation}}
{For prescribed nonnegative budgets
$\epsilon_{\rm mode},\epsilon_p,\epsilon_K,\epsilon_{\rm TV}$, if}
\begin{equation}
 \begin{gathered}
 \|\delta\mu\|\le\epsilon_{\rm mode},\quad
 B_p\le\epsilon_p,\quad B_K\le\epsilon_K,\\
 \operatorname{erf}(\sqrt{B_K}/2)\le\epsilon_{\rm TV},
 \end{gathered}
 \label{eq:opt-freezing-budgets}
\end{equation}
{then the joint-mode translation difference for a common rotation mode
is at most $\epsilon_{\rm mode}$, and
$e_{\rm cm}\le\epsilon_p$, $K_f\le\epsilon_K$ and
$d_{\rm TV}(Q_f,Q_u)\le\epsilon_{\rm TV}$.}
\end{proposition}

\begingroup
\begin{proof}
\textit{Statistic moments.}
Invariance of Haar measure under left and right group translations gives
$Q_R=U^TRV\sim\mathcal{MF}(S)$, with density
$p(Q_R;S)\propto\exp\{\operatorname{tr}(SQ_R)\}$ and
$\nu=(Q_RS-SQ_R^T)^\vee$.
Expectations below are under $R\sim\mathcal{MF}(F)$.
The matrix Fisher first-moment formula~\cite[Theorem~II.2(i)]{lee2018bayesian}
gives $D=\operatorname{diag}(d_1,d_2,d_3)$.
For the $i$th standard basis vector $e_i$ and smooth $\psi$, set
$\mathcal L_i\psi(Q)=\left.\frac{d}{dt}\psi(\exp(t\widehat e_i)Q)\right|_{t=0}$.
Differentiation gives
$
 \mathcal L_i\log p(Q_R;S)=-\nu_i,$ and $
 \mathcal L_i\nu=\{\operatorname{tr}(Q_RS)I-Q_RS\}e_i.
$
The left group translation $Q\mapsto\exp(t\widehat e_i)Q$ preserves
Haar measure, so $\int\mathcal L_i\psi(Q)\,dQ=0$.
Applying this identity to $\psi=p(\,\cdot\,;S)$ and
$\psi=\nu_jp(\,\cdot\,;S)$ yields
$\mathbb E\nu_i=0$ and $\mathbb E\nu_i\nu_j=\mathbb E\mathcal L_i\nu_j$.
Consequently,
$
 \Sigma_\nu=\operatorname{tr}(SD)I-DS
 =\operatorname{diag}(c_1,c_2,c_3),$ where $
 c_i=\sum_{j\ne i}s_jd_j=\mathbb E\nu_i^2.
$
Thus $0\le c_i\le\bar c_i$, since $|d_j|\le1$. \\
\textit{KL and RMS differences.}
Write $m_b(R)=\mu_b+\Gamma_b\nu(R)$ and let $n_b(p\mid R)$ denote
the density of $\mathcal N(m_b(R),\Lambda^{-1})$.
The KL chain rule~\cite[Theorem~2.5.3]{cover2006elements} and the
Gaussian KL formula~\cite[Eq.~(2)]{hershey2007approximating} give
$
 K_f=\operatorname{KL}(Q_u\Vert Q_f)
 =\tfrac12\mathbb E\|W\Delta m\|^2.
$
Using $\mathbb E\nu=0$ and $\Sigma_\nu=\operatorname{diag}(c_1,c_2,c_3)$
from the preceding calculation, expansion of $\Delta m=\delta\mu+\delta\Gamma\nu$
gives, for any $A\in\mathbb R^{3\times3}$,
$
 \mathbb E\|A\Delta m\|^2
 =\|A\delta\mu\|^2+\sum_i c_i\|A\delta\Gamma_{:i}\|^2.
$
Taking $A=W$ and $A=I$ yields
\eqref{eq:opt-freezing-kl} and~\eqref{eq:opt-freezing-rms}, respectively,
using $c_i\le\bar c_i$ for the upper bounds.\\
\textit{Total variation.}
Set $t=\|W\Delta m(R)\|$ and $h(t)=2\Phi(t/2)-1$.
The equal-covariance Gaussian TV formula~\cite[Theorem~1]{barsov1986gaussian}
and the common rotation marginal give
$
 d_{\rm TV}(Q_f,Q_u)
 =\tfrac12\mathbb E\!\int_{\mathbb R^3}|n_f(p\mid R)-n_u(p\mid R)|\,dp
 =\mathbb E h(t).
$
Since $h$ is increasing and concave on $[0,\infty)$ and
$\mathbb E t^2=2K_f$, Jensen's and Cauchy--Schwarz inequalities yield
\begin{align*}
 d_{\rm TV}(Q_f,Q_u)
 &\le h(\mathbb E t)\le h\!\left(\sqrt{\mathbb E t^2}\right)\\
 &=\operatorname{erf}(\sqrt{K_f}/2)
 \le\operatorname{erf}(\sqrt{B_K}/2).
\end{align*}
This proves~\eqref{eq:opt-freezing-tv}.\\
\textit{Modes and translation marginals.}
The conditional Gaussian density has an $R$-independent maximum at
$p=m_b(R)$, so the joint rotation modes are the modes of $\mathcal{MF}(F)$.
At each such mode, $\mathcal L_i\log p=0$ implies $\nu=0$;
hence the joint-mode translations are $\mu_b$, proving their stated
difference $\|\delta\mu\|$.
For a common $R\sim\mathcal{MF}(F)$ and independent
$Z\sim\mathcal N(0,\Lambda^{-1})$, the variables $Y_b=m_b(R)+Z$
couple the translation marginals and satisfy $Y_f-Y_u=\Delta m(R)$.
Their second moments are finite because $\nu$ is bounded on $\mathrm{SO}(3)$.
The coupling characterization of $W_2$ gives
$
 W_2^2((Q_f)_p,(Q_u)_p)
 \le\mathbb E\|Y_f-Y_u\|^2=e_{\rm cm}^2\le B_p^2.
$
This proves~\eqref{eq:opt-freezing-w2}.
The claimed budget guarantees follow from
\eqref{eq:opt-freezing-kl}--\eqref{eq:opt-freezing-tv} and the joint-mode
statement whenever~\eqref{eq:opt-freezing-budgets} holds.
\end{proof}
\endgroup
\endgroup

\subsection{Auxiliary procedures for the MFG estimator}
\label{app:auxiliary_algorithms}
{%
This section expands the numerical routines used by
Algorithms~\ref{alg:single_pass_mfg_update}
and~\ref{alg:safeguarded_multipass_mfg_refinement}.
The data batch $\mathcal D=(\mathcal U,Y,\Sigma_w)$ and initial prior
$\Theta^{(0)}$ remain fixed throughout the outer search.
Damping, tolerances, iteration limits and relaxation sets are supplied
numerical settings. Candidate screening and final posterior acceptance
are separate steps.
}
\begingroup
\setlength{\intextsep}{6pt plus 1pt minus 1pt}
\edef\savedMainAlgorithmCount{\number\value{algorithm}}
\setcounter{algorithm}{0}
\renewcommand{\thealgorithm}{\Alph{section}.\arabic{algorithm}}
\providecommand{\theHalgorithm}{\arabic{algorithm}}
\renewcommand{\theHalgorithm}{auxiliary.\Alph{section}.\arabic{algorithm}}

\subsubsection{{Equilibrium, generalized-force and sensitivity evaluation}}
\label{app:auxiliary_inner}
{%
Algorithm~\ref{alg:auxiliary_inner} introduces a damped Newton variant
for the inner evaluations of the single-pass routine.
}
{%
{This is the original routine used for the results in
Section~\ref{subsec:original_false_confidence}. The additional comparison in
Supplementary Section~\SMref{app:opt-newton} uses Algorithm~\SMref{alg:opt-newton}.}
The comparison in Section~\ref{sec:sec63-frozen-particle} uses its
separately configured common numerical forward solver.
{The warm-start priority is a supplied state, the preceding outer solution
for the same probe, the preceding probe in the current batch, and finally
the commanded pose.}
Each Newton step uses a direct linear solve, with
least squares on failure; it has no inner line search or step clipping.
When the damped matrix is nonsingular, the numerical sensitivity is
$
 J_{k,\lambda}
 =-W_{UA}(W_{AA}+\lambda I_6)^{-1}W_{AB}.
$
Otherwise, the least-squares fallback in the algorithm is used.
For $\lambda>0$, this is a regularized approximation to the undamped
Jacobian in \Cref{thm:reduced_wrench_fisher_information}.
The convergence can be triggered by a small step, so the final
equilibrium residual is recorded and checked separately.
}

\begin{algorithm}[!htbp]
\small
\caption{{Equilibrium, generalized-force and reduced sensitivity evaluation}}
\label{alg:auxiliary_inner}
\begin{algorithmic}[1]
\REQUIRE Pose $X_B$; control $X_{U,k}$; warm start; damping $\lambda$;
tolerances $\varepsilon_{\rm eq},\varepsilon_{\rm step}$; iteration limit $N_{\max}$.
\ENSURE $X_{A,k}^\star$, $\hat y_k$, $J_{k,\lambda}$ and numerical diagnostics.
\STATE Select $X_A$ by the stated warm-start priority; set the convergence flag to false.
\FOR{$j=0,\ldots,N_{\max}-1$}
    \STATE Evaluate $F_A\gets W_A$ and $H_{AA}\gets W_{AA}$ at $(X_A,X_{U,k},X_B)$.
    \IF{$\|F_A\|_2\leq\varepsilon_{\rm eq}$}
        \STATE Set the last-step norm to zero and the flag to true; \textbf{break}.
    \ENDIF
    \STATE Solve $(H_{AA}+\lambda I_6)\Delta\eta_A=-F_A$, using least squares on failure.
    \STATE Set $X_A\gets X_A\oplus\Delta\eta_A$ and record $\|\Delta\eta_A\|_2$.
    \IF{$\|\Delta\eta_A\|_2\leq\varepsilon_{\rm step}$}
        \STATE Set the flag to true; \textbf{break}.
    \ENDIF
\ENDFOR
\STATE Set $X_{A,k}^\star\gets X_A$ and reevaluate $F_A,W_{AA}$ at the returned state.
\STATE Update the flag if $\|F_A\|_2\leq\varepsilon_{\rm eq}$; record the final residual,
last-step norm and iteration count.
\STATE Evaluate $\hat y_k=-\partial_{\xi_U}W_{\rm ctrl}$ and $W_{UA},W_{AB}$ at the returned state.
\STATE Solve $(W_{AA}+\lambda I_6)Z=W_{AB}$, using least squares on failure;
set $J_{k,\lambda}\gets-W_{UA}Z$.
\RETURN $(X_{A,k}^\star,\hat y_k,J_{k,\lambda})$ and the diagnostics.
\end{algorithmic}
\end{algorithm}


\subsubsection{Residual evaluation and posterior-induced branches}
\label{app:auxiliary_interfaces}
{%
$\EvalRes(X_B,\mathcal D,\mathcal X_A^{(0)})$ evaluates the full batch
using the supplied warm starts. It returns the solved batch,
the merit $\rho(X_B)$ from~\eqref{eq:residual_merit}, and a batch record
containing the convergence flags, equilibrium residuals, step norms,
Hessian diagnostics and reduced Jacobians.
The full-batch version described here computes sensitivities for its
numerical checks. Their additional cost must be counted when it is used;
the forward-only residual-check counts in
Corollary~\ref{cor:sec63-mechanical-work} assume that these additional
Jacobian evaluations are not requested. For $\bar X_B=(\bar R_B,\bar p_B)$, let $F_{\rm post}=USV^\top$
be the proper SVD. {A joint mode of the reconstructed Coupled MFG is}
$$
 \textsc{RecoverPose}(\Theta_{\rm post})
 =X_{\rm full}=(UV^\top,\mu_{\rm post}).
$$
Writing $X_{\rm full}=(R_{\rm full},p_{\rm full})$, the other endpoints are
$
 X_{\rm trans}=(\bar R_B,p_{\rm full}),\allowbreak\quad
 X_{\rm rot}=(R_{\rm full},\bar p_B).
$
$\textsc{Branches}$ returns these three endpoints together with their
relaxation sets. The full and translation-only branches use the
nonempty set $\mathcal A\subset(0,1]$, sorted in increasing order with
duplicates removed. When $s_{\rm rot}<\varepsilon_{\rm info}$, only the
rotation-only set is reduced: the supplied factor $c_{\rm rot}$ maps each
$\alpha\in\mathcal A$ to $\max(10^{-8},c_{\rm rot}\alpha)$.
The values are filtered to $(0,1]$, sorted and deduplicated; an empty
result is replaced by $\{\min\mathcal A\}$.
Otherwise $\mathcal A_{\rm rot}=\mathcal A$. For an endpoint $X_{B,b}=\bar X_B\oplus\xi_b$, every trial uses
$$
 \textsc{Relax}(\bar X_B,X_{B,b},\alpha)
 =\bar X_B\oplus(\alpha\xi_b).
$$
All three branches solve the full equilibrium batch at each trial.
}

\subsubsection{Candidate pool and branch search}
\label{app:auxiliary_candidates}
$\textsc{Cand}$ in Algorithm~\ref{alg:safeguarded_multipass_mfg_refinement}
is the candidate-pool routine below. Each branch starts from the same
current equilibrium batch and runs independently.
Within a branch, the returned batch from one trial warm-starts the next
trial, even when that trial is rejected.
The returned candidates have passed only batch and residual screening;
the fixed-prior single-pass recomputation in
Algorithm~\ref{alg:safeguarded_multipass_mfg_refinement}
still determines final acceptance.

\begin{algorithm}[!htbp]
\small
\caption{Posterior-induced candidate pool: \textsc{Cand}}
\label{alg:auxiliary_candidate_pool}
\begin{algorithmic}[1]
\REQUIRE $\bar X_B$, single-pass output $\mathfrak S$, branch descriptors
$\mathcal B$, data $\mathcal D$, and $\varepsilon_{\rm acc}$.
\ENSURE Screened candidate pool $\mathcal C$.
\STATE Read $\mathcal X_A^\star$ and $\rho_t$ from $\mathfrak S$; set $\mathcal C\gets\emptyset$.
\FOR{each endpoint and relaxation set $(X_{B,b},\mathcal A_b)$ in $\mathcal B$}
    \STATE $\mathcal C_b\gets
    \textsc{SearchBranch}(\bar X_B,\mathcal X_A^\star,X_{B,b},
                         \mathcal D,\mathcal A_b,\rho_t,\varepsilon_{\rm acc})$.
    \STATE Set $\mathcal C\gets\mathcal C\cup\mathcal C_b$.
\ENDFOR
\RETURN $\mathcal C$.
\end{algorithmic}
\end{algorithm}

\begin{algorithm}[!htbp]
\small
\caption{Branch-wise candidate search: \textsc{SearchBranch}}
\label{alg:auxiliary_branch_search}
\begin{algorithmic}[1]
\REQUIRE $\bar X_B$, warm batch $\mathcal X_A^\star$, endpoint $X_{B,b}$,
$\mathcal D$, ordered set $\mathcal A_b$, $\rho_t$, and $\varepsilon_{\rm acc}$.
\ENSURE Screened branch candidates $\mathcal C_b$.
\STATE Set $\mathcal C_b\gets\emptyset$ and
$\mathcal X_{A,\rm prev}\gets\mathcal X_A^\star$.
\FOR{each $\alpha\in\mathcal A_b$ in the supplied order}
    \STATE $X_c\gets\textsc{Relax}(\bar X_B,X_{B,b},\alpha)$.
    \STATE $(\mathcal X_{A,c}^\star,\rho_c,\text{batch record})
    \gets\EvalRes(X_c,\mathcal D,\mathcal X_{A,\rm prev})$.
    \IF{$\rho_c\leq\rho_t-\varepsilon_{\rm acc}$ and the batch checks pass}
        \STATE Add $(X_c,\mathcal X_{A,c}^\star,\rho_c)$ to $\mathcal C_b$.
    \ENDIF
    \STATE $\mathcal X_{A,\rm prev}\gets\mathcal X_{A,c}^\star$.
\ENDFOR
\RETURN $\mathcal C_b$.
\end{algorithmic}
\end{algorithm}

\subsubsection{Residual-descent fallback}
\label{app:auxiliary_fallback}
$\textsc{Fallback}$ is called only when the posterior-induced candidate
pool is empty. It returns an empty pool if
$\|g_{\rm data}\|_2\leq\varepsilon_g$.
Otherwise it searches the damped data direction in
Algorithm~\ref{alg:auxiliary_fallback}.
Its screening condition is the strict decrease $\rho_c<\rho_t$;
the final single-pass recomputation must still satisfy
$\rho_{\rm try}\leq\rho_t-\varepsilon_{\rm acc}$.

\begin{algorithm}[H]
\small
\caption{Residual-descent fallback: \textsc{Fallback}}
\label{alg:auxiliary_fallback}
\begin{algorithmic}[1]
\REQUIRE $\bar X_B$, $\mathfrak S$, $\mathcal D$, ordered set $\mathcal A$,
gradient threshold $\varepsilon_g$, and damping $\lambda_{\rm fb}>0$.
\ENSURE Screened fallback pool $\mathcal C_{\rm fb}$.
\STATE Read $g_{\rm data},H_{\rm data},\rho_t,\mathcal X_A^\star$ from $\mathfrak S$;
set $\mathcal C_{\rm fb}\gets\emptyset$.
\IF{$\|g_{\rm data}\|_2\leq\varepsilon_g$}
    \RETURN $\mathcal C_{\rm fb}$.
\ENDIF
\STATE Set $H_{\rm reg}\gets\operatorname{sym}(H_{\rm data})+\lambda_{\rm fb}I_6$.
\STATE Solve $H_{\rm reg}d_{\rm fb}=-g_{\rm data}$;
on failure set $d_{\rm fb}\gets-H_{\rm reg}^{\dagger}g_{\rm data}$.
\STATE Set $\mathcal X_{A,\rm prev}\gets\mathcal X_A^\star$.
\FOR{each $\alpha\in\mathcal A$ in the supplied order}
    \STATE $X_c\gets\bar X_B\oplus(\alpha d_{\rm fb})$.
    \STATE $(\mathcal X_{A,c}^\star,\rho_c,\text{batch record})
    \gets\EvalRes(X_c,\mathcal D,\mathcal X_{A,\rm prev})$.
    \IF{$\rho_c<\rho_t$ and the batch checks pass}
        \STATE Add $(X_c,\mathcal X_{A,c}^\star,\rho_c)$ to $\mathcal C_{\rm fb}$.
    \ENDIF
    \STATE $\mathcal X_{A,\rm prev}\gets\mathcal X_{A,c}^\star$.
\ENDFOR
\RETURN $\mathcal C_{\rm fb}$.
\end{algorithmic}
\end{algorithm}

\subsubsection{Numerical checks and termination}
\label{app:auxiliary_checks}
{%
Batch checks require finite outputs and the prescribed convergence flags,
then test the final equilibrium residuals, step norms, Hessian eigenvalues
and conditioning, and reduced Jacobian norms against the supplied limits.
Posterior checks test finite parameters and local coefficients, positive
conditional precision, conditioning and parameter magnitudes.
They do not impose a Schur positive semidefinite projection.
The returned initial state therefore need not satisfy the later candidate
gate when no update is accepted.

For the numerical data matrix $I=H_{\rm data}$, the rotational score is
computed through
$
 s_{\rm rot}=\lambda_{\min}\!\left(
 \operatorname{sym}\!\left[
 I_{\phi\phi}-I_{\phi v}
 (\operatorname{sym}(I_{vv})+10^{-10}I_3)^{-1}I_{v\phi}
 \right]\right).
$
The inverse denotes a linear solve. If that solve fails, the implementation uses
$(\operatorname{sym}(I_{vv}))^{\dagger}$ in its place.
The resulting Schur matrix is symmetrized before taking its smallest
eigenvalue; no positive semidefinite projection is applied.
This is a numerical regularization of the diagnostic in
\Cref{thm:rotational_information_score}, whose undamped identity requires
$I_{vv}\succ0$. Positive posterior conditional precision does not imply
this data-only condition.

{After an accepted update, $\textsc{Stop}$ uses supplied
tolerances $\varepsilon_R,\varepsilon_p\geq0$ and
$\varepsilon_{\rm stop}>\varepsilon_{\rm acc}\geq0$.
It is true precisely when
\[
 \delta_R^{(t)}\leq\varepsilon_R,
 \qquad\delta_p^{(t)}\leq\varepsilon_p,
 \qquad\rho_t-\rho_{t+1}\leq\varepsilon_{\rm stop}
\]
all hold. Acceptance is unchanged, so stopping allows residual improvements
in $[\varepsilon_{\rm acc},\varepsilon_{\rm stop}]$, subject to both
pose-change tests.} Empty candidate pools, failure of every final recomputation,
and exhaustion of $T_{\max}$ are separate exits.
These tests terminate the numerical search; the residual guarantee is
the acceptance property in \Cref{lem:residual_monotonicity}.
{The returned $X_B^\star$ is the lowest-residual nominal pose
among the initial and accepted states.}
} The prior-mismatch and paired benchmarks described in
Section~\ref{subsec:safeguarded_outer_refinement} use $T_{\max}=20$, while
the 24-dataset study in Section~\ref{sec:sec63-comparison} uses
$T_{\max}=5$. Both use $\varepsilon_R=10^{-6}$ rad,
$\varepsilon_p=10^{-6}$ m and $\varepsilon_{\rm acc}=10^{-6}$.
The nested-budget study is specified
in Supplementary Section~\SMref{supp:s2-nested-results}. The Gaussian
baseline has the separate exits described in Section~\ref{subsec:laplace_historical}.

\setcounter{algorithm}{\savedMainAlgorithmCount}
\endgroup

\subsection{\texorpdfstring{{Proof of Lemma~\ref{lem:residual_monotonicity}}}{Proof of Lemma \getrefnumber{lem:residual_monotonicity}}}
\label{app:residual_monotonicity}
\begin{proof}
\begingroup
The final recomputation in
Algorithm~\ref{alg:safeguarded_multipass_mfg_refinement}
enforces~\eqref{eq:monotone_residual} for every accepted update,
including candidates generated by the fallback routine.
Thus $\varepsilon_{\rm acc}=0$ gives a nonincreasing accepted sequence.
For $\varepsilon_{\rm acc}>0$, summing the acceptance inequalities and
using the nonnegativity of the norm in~\eqref{eq:residual_merit} yields
\[
0\leq\rho(X_B^{(N_{\rm acc})})
\leq\rho(X_B^{(0)})-N_{\rm acc}\varepsilon_{\rm acc}.
\]
Consequently, the number of accepted updates is finite and satisfies the
stated bound.
\endgroup
\end{proof}

\section{\texorpdfstring{{Proofs for Section~\ref{sec:evaluation}}}{Proofs for Section \getrefnumber{sec:evaluation}}}
\label{app:proofs-sec6}

\label{app:sec63-analysis}

\subsection{\texorpdfstring{{Proof of Proposition~\ref{prop:sec63-compact-comparison}\textup{(i)}}}{Proof of Proposition \getrefnumber{prop:sec63-compact-comparison}(i)}}
\label{app:sec63-posterior-proof}
\label{app:sec63-setup}

\begingroup
\begin{proof}
Fix the observations and the reconstructed law $Q_u$, so that
$\mathbb E_{\rm MC}$ averages only over the independent exact draws
$X_1,\ldots,X_S$ in Definition~\ref{def:sec63-posterior-errors}.
Set $m=Q_u(f)$ and $Z_s=f(X_s)-m$. Then
$\mathbb E_{\rm MC}Z_s=0$ and
$\mathbb E_{\rm MC}Z_s^2=\operatorname{Var}_{Q_u}(f)$;
independence gives $\mathbb E_{\rm MC}(Z_rZ_s)=0$ for $r\ne s$.
In particular, $\mathbb E_{\rm MC}\widehat Q_S(f)=m$, and
$
 \widehat Q_S(f)-\pi(f)
 =m-\pi(f)+\frac1S\sum_{s=1}^S Z_s.
$
Squaring and taking expectation eliminates both the bias--sample cross
term and the cross terms between distinct samples, yielding
\[
\begin{aligned}
 \operatorname{RMSE}_f(\widehat Q_S)^2
 &=[m-\pi(f)]^2+\frac1{S^2}\sum_{s=1}^S\mathbb E_{\rm MC}Z_s^2\\
 &=[Q_u(f)-\pi(f)]^2+\frac{\operatorname{Var}_{Q_u}(f)}{S}.
\end{aligned}
\]
The variational definition of TV in
Definition~\ref{def:sec63-posterior-errors} gives
$|Q_u(f)-\pi(f)|\le d_{\rm TV}(Q_u,\pi)$.
Moreover, $0\le f\le1$ implies $f^2\le f$ and $0\le m\le1$, hence
$
 \operatorname{Var}_{Q_u}(f)=Q_u(f^2)-m^2
 \le m(1-m)\le 1/4.
$
Substitution proves both the identity and the upper bound
in~\eqref{eq:sec63-mfg-risk}.
\end{proof}

\subsection{\texorpdfstring{{Posterior bound and proof of Proposition~\ref{prop:opt-actual-law}}}{Posterior bound and proof of Proposition \getrefnumber{prop:opt-actual-law}}}
\label{app:opt-gap}
\label{app:sec63-local-gap}

Let $Q$ be any fixed valid returned MFG with actual negative log density $\mathcal V_Q$, and let
$
 \pi(dX)=Z_\pi^{-1}e^{-\mathcal E(X)}\,d\mu(X),
 0<Z_\pi<\infty,
$
{with respect to the same product of Haar measure on $\mathrm{SO}(3)$
and Lebesgue measure on $\mathbb R^3$.} 
Fix a measurable local region $D$ with positive probability under both laws, 
and $\bar X\in D$ at which both energies are finite. Assume that, for a finite $\delta_Q\ge0$,
\begin{equation}
 \sup_{X\in D}\left|
 [\mathcal V_Q(X)-\mathcal V_Q(\bar X)]
 -[\mathcal E(X)-\mathcal E(\bar X)]\right|\le\delta_Q.
 \label{eq:opt-direct-gap}
\end{equation}
Set $\tau_Q=Q(D^c)$ and $\tau_\pi=\pi(D^c)$.

\Needspace{5\baselineskip}
\begin{proposition}[Actual-law posterior bound]
\label{prop:opt-actual-law}
Under \eqref{eq:opt-direct-gap},
\begin{equation}
 d_{\rm TV}(Q,\pi)\le
 B_Q:=\min\{1,\tau_Q+\tau_\pi+\tanh(\delta_Q/2)\}.
 \label{eq:sec63-total-bound}
\end{equation}
For $Q=Q_u$, \eqref{eq:sec63-mfg-risk} is bounded by $B_Q^2+1/(4S)$.
The statement applies to {the distribution $Q_u$ produced by the Coupled MFG update} and to a valid regularized output when its \emph{actual} energy is used in the gap condition~\eqref{eq:opt-direct-gap}.
\end{proposition}
\begin{proof}
Set $P=\pi(\cdot\mid D)$ and
$h(X)=\exp\{[\mathcal E(X)-\mathcal E(\bar X)]
-[\mathcal V_Q(X)-\mathcal V_Q(\bar X)]\}$.
Then $Q(\cdot\mid D)=hP/z$, where $z=P(h)$.
For $\delta_Q>0$, let $a=e^{-\delta_Q}$ and $b=e^{\delta_Q}$.
By~\eqref{eq:opt-direct-gap}, $a\le h\le b$.
The Edmundson--Madansky inequality~\cite[Sec.~2]{egozcue2012smallest} applied to $t\mapsto|t-z|$ gives
\[
\begin{aligned}
 d_{\rm TV}(Q(\cdot\mid D),P)
 &=\frac{P(|h-z|)}{2z}
 \le\frac{(b-z)(z-a)}{z(b-a)}\\
 &\le\frac{\sqrt b-\sqrt a}{\sqrt b+\sqrt a}
 =\tanh(\delta_Q/2),
\end{aligned}
\]
where the second inequality follows from $z+ab/z\ge2\sqrt{ab}$.
If $\delta_Q=0$, then $h=1$ and $Q(\cdot\mid D)=P$.
Finally, $d_{\rm TV}(Q,Q(\cdot\mid D))=\tau_Q$ and
$d_{\rm TV}(\pi,P)=\tau_\pi$; the triangle inequality and
$d_{\rm TV}\le1$ prove~\eqref{eq:sec63-total-bound}.
\end{proof}

\endgroup

\begingroup
\begin{proposition}[A local energy-gap bound for the actual MFG update]
\label{prop:local-gap-actual-mfg}
Let $\mathcal E=J_{\mathrm{MAP}}=J_{\mathrm{data}}+E_0$, up to a
constant, and $H=H_{\mathrm{data}}+H_{\mathrm{prior}}$ be as in
Proposition~\ref{prop:local_quadratic_posterior}, with exact prior
Hessian $H_{\mathrm{prior}}^{\mathrm{ex}}$.
Suppose that the full update of
Theorem~\ref{thm:closed_form_reconstruction} at $\bar X_B$, using $H$ and the
exact gradient $g=\nabla_\xi\mathcal E(\bar X_B\oplus\xi)|_{\xi=0}$,
returns a valid MFG law $Q_u$ with actual energy $\mathcal V_u$ and
curvature correction $C_u$ from Remark~\ref{rem:actual-mfg-curvature}.
Use the posterior $\pi$ and product measure $\mu$ defined
above, with $0<Z_\pi<\infty$.
For the locally $C^2$ residuals
$r_k(\xi)=y_k-h_k(\bar X_B\oplus\xi)$, set $\bar r_k=r_k(0)$ and
\[
\begin{aligned}
H_{\mathrm{curv,data}}
&=\sum_{k=1}^{K}\sum_{\alpha=1}^{6}
 (\Sigma_w^{-1}\bar r_k)_\alpha D_\xi^2r_{k,\alpha}(0),\\
\Delta H
&=\operatorname{diag}(C_u,0_{3\times3})-H_{\mathrm{curv,data}}\\
 &\quad-(H_{\mathrm{prior}}^{\mathrm{ex}}-H_{\mathrm{prior}}).
\end{aligned}
\]
For fixed scales $s_\phi,s_p>0$ and $r>0$, define
\[
\begin{gathered}
T_{\mathrm{sc}}=\operatorname{diag}(s_\phi I_3,s_pI_3),\\
\Psi(z)=\bar X_B\oplus T_{\mathrm{sc}}z,
\qquad D_r=\Psi(\overline B_r(0)),
\end{gathered}
\]
where $\overline B_r(0)=\{z\in\mathbb R^6:\|z\|_2\le r\}$.
Assume that $\Psi$ is a smooth injective chart on a neighbourhood
of this ball and that $\mathcal G=(\mathcal V_u-\mathcal E)\circ\Psi$
is $C^3$ there, within a single smooth equilibrium branch. Set
\[
\Delta H_{\mathrm{sc}}=T_{\mathrm{sc}}^\top\Delta H T_{\mathrm{sc}},
\qquad
M_3\ge\sup_{\|z\|_2\le r}\|D^3\mathcal G(z)\|_{\mathrm{op}},
\]
with $M_3<\infty$.
Then the centred energy-gap condition~\eqref{eq:opt-direct-gap}
holds with $Q=Q_u$, $D=D_r$, $\bar X=\bar X_B$, and
\[
\delta_Q=\delta_u(r)
:=\frac12\|\Delta H_{\mathrm{sc}}\|_2r^2+\frac{M_3}{6}r^3.
\]
If $Q_u(D_r)>0$ and $\pi(D_r)>0$, the TV bound~\eqref{eq:sec63-total-bound}
applies with $\tau_Q=Q_u(D_r^c)$ and $\tau_\pi=\pi(D_r^c)$.
For $H_{\mathrm{prior}}=H_{\mathrm{prior}}^{\mathrm{GN}}$,
\eqref{eq:prior-curvature-correction} gives
\[
\Delta H=\operatorname{diag}(C_u+C_{\nu,\mathrm{prior}},0_{3\times3})
-H_{\mathrm{curv,data}},
\]
where $C_{\nu,\mathrm{prior}}$ denotes $C_\nu$ in
Remark~\ref{rem:prior-curvature}.
\end{proposition}

\begin{proof}
{Applying the least-squares Hessian identity~\cite[Eq.~(10.5)]{nocedal2006numerical}
to the residuals $\Sigma_w^{-1/2}r_k$ gives}
$
\nabla_\xi^2\mathcal E(\bar X_B\oplus\xi)|_{\xi=0}
=H_{\mathrm{data}}+H_{\mathrm{curv,data}}
+H_{\mathrm{prior}}^{\mathrm{ex}}.
$
Together with Remark~\ref{rem:actual-mfg-curvature} and the exact-gradient
assumption, this yields
$D\mathcal G(0)=0$ and $D^2\mathcal G(0)=\Delta H_{\mathrm{sc}}$.
For $\|z\|_2\le r$, the segment $\{tz:0\le t\le1\}$ stays in
$\overline B_r(0)$. Taylor's integral remainder along this segment gives
\[
\begin{aligned}
\mathcal G(z)-\mathcal G(0)
&=\frac12z^\top\Delta H_{\mathrm{sc}}z+R_3(z),\\
|R_3(z)|
&\le\frac{M_3\|z\|_2^3}{2}\int_0^1(1-t)^2\,dt
=\frac{M_3}{6}\|z\|_2^3.
\end{aligned}
\]
Taking the supremum gives $\delta_u(r)$, and
Proposition~\ref{prop:opt-actual-law} gives the TV bound.
The GN-prior formula follows from~\eqref{eq:prior-curvature-correction}.
\end{proof}

\endgroup

\subsection{\texorpdfstring{{Proof of Proposition~\ref{prop:sec63-compact-comparison}\textup{(ii)}}}{Proof of Proposition \getrefnumber{prop:sec63-compact-comparison}(ii)}}
\label{app:sec63-particles}

{%
Let $\ell_k(X)=\tfrac12\|y_k-h_k(X)\|_{\Sigma_w^{-1}}^2$ and use
the fixed finite target path
\[
 \pi_t(dX)\propto p_0(X)
  \exp\!\left[-\sum_{k=1}^K\alpha_{t,k}\ell_k(X)\right]d\mu(X),
\]
where $
\alpha_{0,k}=0,\alpha_{T,k}=1,
$
with nondecreasing exponents. RM-PF \cite{gilks2001resamplemove} introduces observations in order,
 tempered SMC \cite{delmoral2006smc} uses
$\alpha_{t,k}=\beta_t$. Both paths end at $\pi_T=\pi$.
Assume positive bounded incremental weights
$G_t=\exp[-\sum_k(\alpha_{t,k}-\alpha_{t-1,k})\ell_k]$
on a common support, exact independent initialization from $\pi_0$
with equal weights, and fixed kernels satisfying $\pi_tM_t=\pi_t$.
The path and kernels are independent of $N$ and of the current particle
population.

\begin{proof}[Particle bound]
We use the finite-stage $L^2$ argument of
\cite[Lemma~4 and Theorem~2]{CrisanDoucet2002Survey}.
For the resampling rule used here, fix $\eta\in(0,1]$ and condition on
the reweighted particles with normalized weights $w_i$.
The decision to resample when $\operatorname{ESS}(w)<\eta N$ is then fixed.
{Independent stratified resampling is unbiased
\cite[Sections~2.3 and~3.3]{DoucCappeMoulines2005Resampling}.}
After independent moves with kernel $M_t$, the contributions for
$0\le f\le1$ are conditionally independent, giving variance at most
$1/(4N)$. If resampling is skipped, the moved particles retain $w_i$,
and the conditional variance is at most
$1/4\sum_iw_i^2\le1/(4\eta N)$.
Both branches are conditionally unbiased for the expectation under the
reweighted empirical measure followed by $M_t$.
Thus their local $L^2$ error is at most $1/(2\sqrt{\eta N})$. Independent initialization has $L^2$ error at most $1/(2\sqrt N)$.
Each fixed kernel $M_t$ maps $[0,1]$-valued functions into themselves,
while $G_t$ is bounded and $\pi_{t-1}(G_t)>0$. Applying the normalization estimate in~\cite[Lemma~4]{CrisanDoucet2002Survey}
at each of the finitely many stages, together with the local error bound above,
gives~\eqref{eq:sec63-particle-risk} with a finite
$C_{\mathrm{path}}$ independent of $N$.
\end{proof}
}

\subsection{\texorpdfstring{{Proof of Corollary~\ref{cor:sec63-mechanical-work}}}{Proof of Corollary \getrefnumber{cor:sec63-mechanical-work}}}
\label{app:sec63-work}

\begingroup
Each observation block is evaluated in full, with one equilibrium
solve and, when requested, one additional reduced Jacobian.
The count $n_{\mathrm{rec}}$ includes initialization and every attempted
full-batch reconstruction, including final posterior-health checks. The quantity
$n_{\mathrm{res}}$ counts only separate forward-only batch checks.
For the particles, current-state likelihood terms are cached and
inherited on resampling. Paths and move counts are fixed independently
of $N$, and each move uses one derivative-free Metropolis--Hastings~\cite{hastings1970montecarlo}
proposal.

\begin{proof}
Each full-batch reconstruction contributes $(K,K)$ to
$(n_{\mathrm{eq}},n_J)$, and each separate residual check contributes
$(K,0)$. Therefore, for the proposed framework,
\[
\begin{aligned}
 (n_{\mathrm{eq}},n_J)
 &=\sum_{r=1}^{n_{\mathrm{rec}}}(K,K)
   +\sum_{r=1}^{n_{\mathrm{res}}}(K,0)\\
 &=\bigl(K(n_{\mathrm{rec}}+n_{\mathrm{res}}),Kn_{\mathrm{rec}}\bigr).
\end{aligned}
\]
For RM-PF, incorporating observation $k$ requires $N$ solves, and
its $NM_k$ proposals each require $k$ solves.
Hence
$
 n_{\mathrm{eq}}
 =\sum_{k=1}^{K}(N+NM_k k)
 =N(K+\sum_{k=1}^{K}kM_k).
$
For tempered SMC, initialization requires $NK$ solves, and each of the
$NM_t$ proposals at stage $t$ evaluates all $K$ observations. Thus
$
 n_{\mathrm{eq}}
 =NK+\sum_{t=1}^{T}NM_tK
 =NK(1+\sum_{t=1}^{T}M_t).
$
The particle moves are derivative-free, so $n_J=0$ for both methods.
Reweighting and resampling require no further solves.
This proves~\eqref{eq:sec63-mechanical-counts}, whose particle counts
are linear in $N$ for fixed paths and move counts.
By \Cref{def:coupled_mfg}, an exact MFG draw samples its rotation and
then its conditional Gaussian translation. Neither step evaluates the
mechanical model, so further draws leave the construction counts
unchanged, independently of $S$.
\end{proof}
\endgroup

\section*{Declaration of generative AI and AI-assisted technologies in the
manuscript preparation process}

During the preparation of this manuscript and its supplementary material, the authors used
ChatGPT 6 Pro and Codex (6.1 Astra) to assist with manuscript revision, clarify the
mathematical presentation, and perform consistency checks against experimental records. These
tools also assisted with the development of simulation and visualization code for the two
examples in Section~\ref{subsec:shape_pose_clouds}.  The authors reviewed and revised the AI-assisted text and code
and checked the resulting numerical outputs. The authors take full responsibility for the
accuracy, integrity, and content of the manuscript and its supplementary material.

\bibliographystyle{elsarticle-num}
\bibliography{references}
\end{document}